\documentclass[11pt, dvipsnames]{article}

\pdfoutput=1
\usepackage{enumitem}
\usepackage{titlesec}
\usepackage{authblk}
\usepackage[dvipsnames]{xcolor} 
\usepackage{ucs}
\usepackage{longtable}
\usepackage[colorlinks=true,
  urlcolor=ForestGreen,
  linkcolor=Blue,
  citecolor=Violet,
  pagebackref=true,
  pdftitle={Titel},
  pdfsubject={},
  pdfauthor={},
  pdfkeywords={keywords}
]{hyperref}
\usepackage{graphicx}
\usepackage{subfigure}
\usepackage[subfigure]{tocloft}
\usepackage[font=footnotesize]{caption}
\usepackage{tabularx}
\usepackage{nameref}
\usepackage{tikz-cd}

\usepackage{amsmath} 	  
\usepackage{amssymb}	  
\usepackage{amsthm}	    
\usepackage{mathtools}  
\usepackage[normalem]{ulem} 
\usepackage{graphicx,wrapfig,lipsum}
\usepackage{float}
\usepackage{floatflt}
\usepackage{natbib}
\usepackage[capitalize,nameinlink]{cleveref}

\usepackage[most]{tcolorbox}
\tcbuselibrary{breakable} 
\usepackage{wrapfig}
\usepackage[toc,page]{appendix}

\usepackage{hyperref}
\usepackage{cleveref}
\usepackage{algorithm}
\usepackage{algorithmicx}
\usepackage{algpseudocode}

\usepackage{array}
\usepackage{enumerate}

\usepackage{booktabs}
\newcommand{\R}{\mathbb{R}}

\theoremstyle{remark}

\title{curvature as a tool for evaluating dimensionality reduction and estimating intrinsic dimension}
\author[1,3]{Charlotte Beylier, \thanks{beylier@cbs.mpg.de}} 
\author[1,2]{Parvaneh Joharinad, \thanks{johari@mis.mpg.de}}
\author[1,2,4]{J\"urgen Jost, \thanks{jjost@mis.mpg.de}}
\author[3]{Nahid Torbati, \thanks{torbati@cbs.mpg.de}}
\affil[1]{Center for Scalable Data Analytics and Artificial Intelligence (ScaDS,AI) Dresden/Leipzig, Germany}
\affil[2]{Max Planck Institute for Mathematics in the Sciences, Leipzig, Germany}
\affil[3]{Max Planck Institute for Human Cognitive and Brain Sciences, Leipzig, Germany}
\affil[4]{Santa Fe Institute for the Sciences of Complexity, New Mexico, USA}

\begin{document}

\maketitle

\begin{abstract}
Utilizing recently developed abstract notions of sectional curvature, we introduce a method for constructing a curvature-based geometric profile of discrete metric spaces. The curvature concept that we use here captures the metric relations between triples of points and
other points. 
More significantly, based on this curvature profile, we introduce a quantitative measure to evaluate the effectiveness of data representations, such as those produced by dimensionality reduction techniques. Furthermore, Our experiments demonstrate that this curvature-based analysis can be employed to estimate the intrinsic dimensionality of datasets.
 We use this to explore the large-scale geometry of empirical networks and to evaluate the effectiveness of dimensionality reduction techniques. 
\end{abstract}

\section{Introduction}
Often, the geometric structure of datasets can be captured by graphs, and they play a central role in modern machine learning methods such as Convolutional Neural Networks (CNNs) \cite{lecun1998gradient} and Graph Neural Networks (GNNs) \cite{scarselli2008graph}. It is then natural, if not mandatory, to ask how well graphs preserve or reveal the underlying geometry. We therefore need principled tools that can infer geometric and topological features within these models. Graphs themselves being viewed as metric spaces, require tools evolving from geometry. However, many of such tools were initially developed for particular smooth structures, namely the Riemannian manifolds.\\
For example, if $S\subset\R^3$ is an oriented surface, the Gaussian curvature quantifies the deviation of $S$ at each point from flatness. Following Riemann's foundational work, it is well-established that Gaussian curvature can be computed using the intrinsic geometrical structure induced by the Euclidean geometry of $\R^3$. This established the foundation for the development of Riemannian geometry by extending the concept of Gaussian curvature to sectional curvature. The latter is defined as the measure of deviation from a flat geometry, providing a comprehensive set of invariants for the characterization of the geometry of a manifold \cite{Jost17a}. 
Sectional curvature, however, is derived from the second derivative of the metric tensor and thus cannot be directly applied to general metric spaces. Still, upper and lower bounds on sectional curvature can impose restrictions on the metric structure of manifolds, providing a qualitative measure of how general metric spaces (i.e., geodesic spaces) deviate from Riemannian manifolds of constant sectional curvature (i.e., model spaces).\\
 For such general metric spaces,  there exist  various synthetic curvature notions that replace the infinitesimal computations. Here, we use the notion introduced in \cite{Joharinad19}, which applies even to discrete metric spaces, for analyzing the large-scale geometry of graphs. Our  curvature profile is a distribution function evaluated on equilateral triples of nodes, organized according to their perimeter. This profile acts as a geometric signature, capturing curvature characteristics of the graphs as deviations from tree-like structure at different scales. \\
For a metric dataset $(X,d)$ ($d$ might record pairwise (dis)similarities),  we first transform the metric structure into a combinatorial model by constructing a neighborhood graph and evaluating its curvature profile as a geometric fingerprint. For quantifying how well low dimensional embedding methods preserve the intrinsic geometry of the data, we compute the 1-Wasserstein distance ($W_1$)  \cite{villani2008optimal} between the curvature profiles of the initial dataset and its corresponding embedding.  
In particular, varying the target dimension, we can identify the embedding whose curvature distribution best matches that of the original dataset, providing an effective estimate of the intrinsic dimension.\\
To demonstrate the effectiveness of the curvature profile in capturing and inferring underlying geometric structure, we first apply our method to metric datasets with known geometries, the circle, the plane, and a metric tree, and then to empirical data. Our full code is available at \footnote{\url{https://anonymous.4open.science/r/curvature-D6FE}}.

\section{Graphs and curvatures}
Curvature is a concept that originated in differential geometry and that plays a fundamental role in Riemannian geometry. It measures the local deviation of a space from being Euclidean. In fact, in 1854 Bernhard Riemann showed that the sectional curvatures of a Riemannian manifold provide a complete set of local invariants of the Riemannian metric \cite{Jost25a}. That is, if you know the curvatures, you know the intrinsic geometry of a Riemannian manifold. Here, \emph{intrinsic} refers to measurements that are taken inside the manifold and do not use any relations in an ambient space into which the manifold is embedded. And a contraction of the sectional curvatures, the Ricci curvatures became fundamental, for instance, in Einstein's theory of general relativity. \\
In the 20th century, mathematicians developed notions of synthetic curvature for more general metric spaces. When the metric space is Riemannian, they should of course reduce to the classical notions. The most useful such  generalized curvature notions were  introduced by Alexandrov \cite{Alexandrov57} and Busemann \cite{Busemann55}. They  provide inequalities that encode important local and global properties of the space. They have become powerful tools for probing the geometry of metric space. \\
Graphs can also be seen as metric spaces. In the simplest case, when the graph is unweighted, the distance between two vertices is the minimal number of edges needed to get from one to the other. Therefore, synthetic curvature notions also apply to graphs, and they have become useful tools for the analysis of empirical networks represented as graphs. \\
As indicated, there are two important notions of curvature, sectional and Ricci. While the latter is a contraction of the former, in Riemannian geometry they play somewhat opposite roles, see e.g. \cite{Jost17a}. From a positive lower bound on the Ricci curvature, one can derive many consequences for the local geometry, like diameter bounds, restrictions on the growth of the volume of balls as function of their radius, average divergence of geodesics, eigenvalue bounds and so on. In contrast, negative Ricci curvature does not imply any restriction at all. For sectional curvature, it is the opposite. Here, nonpositive or even better, negative sectional curvature implies many global restrictions, that is, restrictions on the large scale geometry. \\
Synthetic versions of either curvature have been applied in graph theory and network analysis. Since the Ricci curvature of a Riemannian manifold is evaluated on tangent vectors, the Ricci curvature of a graph is naturally computed for edges. Since the sectional curvature is evaluated on tangent planes and synthetic notions are usually formulated in terms of distance relations in geodesic triangles, the sectional curvature of a graph is naturally computed on triangles of vertices. Here, a triangle is a triple of vertices connected by shortest edge paths. These triangles can be large, in line with the observation that Riemannian sectional curvatures encode global properties.\\
There are several versions of synthetic Ricci curvature. While for Riemannian manifolds, they all coincide, as they should, for graphs they are very different. The most prominent notions were developed by Bakry-\'Emery, Forman, and Ollivier. The statistics of such Ricci curvatures on empirical networks has become a useful and versatile to detect significant properties and to distinguish different types of networks, see for instance, as samples of a very rich literature, \cite{Samal18}, \cite{Saucan18}, \cite{Mondal24}. But since in the present paper, we are rather interested in the global geometry, we shall work with sectional rather than Ricci curvature. \\
In the context of sectional curvatures, Gromov \cite{Gromov99} has developed the notion of $\delta$-hyperbolicity, as a global version of negative curvature. He applied this notion to the Cayley graph of groups and discovered that most groups are $\delta$-hyperbolic, that is, negatively curved. Similarly, empirical networks have been analyzed from this perspective, for instance in (\cite{Albert14, Boguna09, Jonckheere07,krioukov2010hyperbolic,Narayan11}, and again, it was argued that most networks are hyperbolic. \\
Our question here is different. We want to compare the global geometry of metric spaces. As argued, synthetic sectional curvatures should be a good tool. And we don't want just test a criterion for hyperbolicity, but rather compare curvature profiles, be they negative or positive. In fact, empirical networks, while perhaps predominantly negatively curved, may not show negative curvature for every triangle.  Indeed, most locally finite graphs do not exhibit negative curvature at small scales, and the non-positive curvature condition typically holds only at larger scales or asymptotically. Thus, in order to address this question, we need more than simply a criterion for hyperbolicity. And since the standard notions of synthetic sectional curvatures, like those of Busemann \cite{Busemann55} and Alexandrov \cite{Alexandrov57} are not meaningful for graphs, we have to utilize a more general version recently developed in \cite{Joharinad19,Joharinad23}. That will be described in the next section.

\section{Sectional curvature profile}
We shall now introduce the concept of curvature introduced in \cite{Joharinad19}, which is the foundation of our curvature profile, and describe how it encodes the geometry of a metric space. We shall also compare this condition with Gromov's $\delta$-hyperbolicity condition and explain how these two differ from one another.\\
We  consider a discrete dataset $X=\{x_1,\dots,x_N\}$ equipped with some distance function $d$ between data points. For instance, $X$ could be a subset of a feature space (e.g. $\R^n$), or  the set of nodes in a network, where edges encode pairwise correlations. In the first case, the distance  $d$  usually arises as the Euclidean distance. In the second case, it would be the integer valued graph distance, i.e. the minimal number of steps required to travel from one node to another via connected edges.\\
For each triple of points $x_1,x_2,x_3$ in our metric space $(X,d)$ with mutual distances $d(x_1,x_2), d(x_1,x_3), d(x_2,x_3)$, we convert those three distances into three other numbers $r_1,r_2,r_3$, the so called Gromov products, that are given as the 
 unique solutions of  
\[
r_i+r_j=d(x_i,x_j),\:1\leq i<j\leq3.
\]

\begin{figure}[H]
\centering{\begin{tikzpicture}
\draw[color=black] (6,0) -- (4,0) node[] {$\bullet$}; 
\draw[color=black] (4,0) -- (3,2) node[] {$\bullet$};
\draw[color=black] (3,2) -- (6,0) node[] {$\bullet$}; 
\draw (6,0) circle[radius=1.68cm];
\draw (4,0) circle[radius=3.15mm];
\draw (3,2) circle[radius=1.915cm];
\end{tikzpicture}
}
 \caption{}
 \label{EuclideanGromov}
\end{figure}
When the points $x_1,x_2,x_3 $ are the outer nodes of a tripod,  they posses a median $m$, the Gromov product measures the lengths of the edges of that tripod (i.e.  $r_i=d(x_i,m)$ for $i=1,2,3$), and 
\begin{equation}\label{hy1}
m\in B(x_1,r_1)\cap B(x_2,r_2) \cap B(x_3,r_3)\neq \emptyset
\end{equation}
In general, while the balls $B(x_i,r_i)$ supposedly intersect pairwise by construction, there might not be any triple intersection.  To obtain a nonempty triple intersection, we need to enlarge the radii. And how much we need to enlarge them yields the value of our curvature.  That is, the smallest multiplicative factor $\rho$ for which the intersection 
\begin{equation}
B(x_1,\rho r_1)\cap B(x_2,\rho r_2) \cap B(x_3,\rho r_3)
\end{equation}
 is nonempty, which is our quantitative curvature measure.\\
Thus, the factor $\rho$ indicates how much the balls must be scaled to obtain a common intersection point, and encodes curvature in terms of how far the space deviates from a tree-like structure. \\
Three equally spaced points $x_1,x_2,x_3$ on a circle represent the extreme case opposite to that of the tripod configuration. In this situation, the balls around them must be enlarged by a factor of $2$ to get a triple intersection.  \\
One can also compute $\rho$ as
\begin{equation}\label{rho}
\rho(x_1,x_2,x_3):=\inf_{x\in X}\max_{i=1,2,3}\frac{d(x_i,x)}{r_i}.
\end{equation}
Apart from the above mentioned  extremes, in complete metric spaces, the value of $\rho$ at every triple of points is a number between $1$ and $2$. If the infimum in (\ref{rho}) is attained by some point, then we may call that point, whose maximal weighted distance from the vertices has the smallest possible value, a {\em weighted circumcenter}.  In the Euclidean plane $\R^2$, and more generally in every complete $CAT(0)$ space, i.e. a space of nonpositive Alexandrov curvature, the infimum in \cref{rho} is attained by a unique point $m$ for any triple of points $(x_1,x_2,x_3 )$. Moreover,  $d(x_i,m)=\rho(x_1,x_2,x_3 ) r_i$ for $i=1,2,3$, c.f. Corollary 3.1 in \cite{Joharinad19}. In fact, for three arbitrary points $x_1,x_2, x_3\in \R^2$, the value of $\rho$ and the position of the weighted circumcenter $m$ are determined by solving the following system of equations
\begin{align*}
d^{2}(x_1,x_2)&=\rho^{2}(r_{1}^{2}+r_{2}^{2}-2 r_{1} r_{2}\cos(\alpha)),&\\
d^{2}(x_2,x_3)&=\rho^{2}(r_{2}^{2}+r_{3}^{2}-2 r_{2} r_{3}\cos(\beta)),&\\
d^{2}(x_1,x_3)&=\rho^{2}(r_{1}^{2}+r_{3}^{2}+2 r_{1} r_{3}\cos(\alpha+\beta)),&
\end{align*}
where $\alpha$ and $\beta$ are the angles at $m$ opposite to  $[x_1, x_2]$ and $[x_2, x_3]$, respectively. Notably,  $\rho$ achieves its maximum value, $\dfrac{2}{\sqrt{3}}$, at each equilateral triangle.\\
An important property of the function $\rho$  is its invariance under scaling the metric of the space. For instance, if we consider three vertices $(x_1,x_2,x_3 )$ located on a circle centered at the point $O$, regardless of the radius of the circle, the value of $\rho$ is given by 
\begin{equation}\label{rho-circle}
\rho(x_1,x_2, x_3)=\dfrac{2\pi}{\angle_{O}(x_i,x_j)}-1, 
\end{equation}
where $[x_i,x_j]$ is the edge with maximum length. One can also check that for three equidistant vertices on the circle $\rho$ reaches its maximum value $2$. \\
In practical applications, however, we are interested in determining the value of $\rho$ by considering the first instance at which the three scaled balls intersect. \\

Generalizing the phenomena that underpins this curvature concept to more than three points defines a hyperconvex space. The idea is that for any finite collection of balls that all overlap pairwise, there is at least one point where all of the balls meet.\\

The $\delta$-hyperbolicity condition, on the other hand, is set as follows. Instead of having to enlarge the radii by a multiplicative factor to generate a triple intersection, we only need to enlarge them by an additive constant $\delta$. In fact, the smallest $\delta$ for which
\begin{equation}\label{hy2}
B(x_1,r_1 +\delta)\cap B(x_2,r_2 +\delta) \cap B(x_3,r_3 +\delta) \neq \emptyset.
\end{equation}
is used to determine how negatively curved a space is.  More formally, in a $\delta$-hyperbolic space, the enlargement needed to obtain a triple intersection is bounded by the constant $\delta$, irrespective of the triangle's size. This characteristic implies that every finite metric space is $\delta$-hyperbolic for some $\delta$, as there will always be a constant that satisfies the condition, regardless of the space's geometry. However, this leads to a notable shortcoming: the $\delta$-hyperbolic condition does not inherently account for the scaling of the space. \\
To address this limitation, efforts have been made to refine the notion of $\delta$-hyperbolicity by introducing scaling factors. These factors are typically related to the perimeter of triangles or the diameter of the metric space. By incorporating these scaling measures, one can better distinguish spaces that are genuinely negatively curved from those that merely satisfy the $\delta$-hyperbolic condition. This refinement is crucial in applications where the distinction between negative curvature and mere boundedness has significant implications for the large-scale behavior of the space.
If the space is genuinely negatively curved, then asymptotically, the condition \cref{hy2} is not qualitatively different from \cref{hy1}.  Specifically, for large triangles, meaning when the values of $r_1,r_2,r_3$ are large, that fixed constant $\delta$ in $\delta$-hyperbolic spaces become rather insignificant. That is, at large scales, $\delta$-hyperbolic spaces become very different from Euclidean ones and behave more like tripod spaces. As with the three-point condition, which led to the hyperconvexity criterion, it is possible to define the generalized $\delta$-hyperbolic spaces by considering any family of closed balls.\\
In other words, the structural difference between $\delta$-hyperbolic spaces and hyperconvex spaces diminishes as the scale of the geometric structure increases. While $\delta$-hyperbolic spaces may resemble Euclidean spaces at small scales due to the additive bound $\delta$, their global behavior at larger scales starts to mimic the extreme cases of negative curvature, much like tripod spaces. \\
In any case, $\delta$-hyperbolicity is a good criterion for properties of a space that are hyperbolic at large scales. And while this criterion may be mostly satisfied for empirical networks, it may not always be satisfied. The curvature notion of \cite{Joharinad19,Joharinad23}, in contrast always applies. And when we find triangles that are not hyperbolic, we can still quantify their curvature. In that way, we can develop curvature profiles of metric spaces and compare them. \\

To conclude the theoretical discussion, we observe that $\rho$ does not merely depend on the metric space $(X,d)$, but also on the shape of the triangle under consideration. In particular, within the Euclidean plane, equilateral triangles achieve the maximum value of $\rho$, while degenerate triangles (those formed by three collinear points) yield the minimum value.  \\
For an adequate comparison, it is first necessary to classify the triangles. On the finite metric space $(X,d)$, one can introduce a measurement, i.e. $\lambda$, to group triangles in $X$ into classes $X_\lambda$. With $\lambda$ as the second parameter, we can compute the value of $\rho$ for triangles in each class $X_{\lambda}$, and trace how $\rho$ evolves as the perimeter (the first parameter) increases.  This allows us to compare the results with corresponding figures from model spaces, identifying which model space $X$ most closely resembles.\\
The most natural choice for the second parameter $\lambda$ is one that quantifies the deviation from equality in the triangle inequality.  For each triple $(x_1,x_2,x_3)$ in $X$, we define $\lambda$ as the maximum value of $\alpha$ that satisfies the following conditions
\begin{align}
\nonumber\alpha d(x_1,x_2)\leq d(x_1,x_3)+d(x_2,x_3),&\\
\nonumber \alpha d(x_1,x_3)\leq d(x_1,x_2)+d(x_2,x_3),&\\
 \alpha d(x_2,x_3)\leq d(x_1,x_2)+d(x_1,x_3).&
\end{align}
Since $\lambda$ is maximal for these  inequalities, at least one of them will hold with equality. If  $\lambda=1$, the three points are collinear, making the triangle degenerate. On the other hand, if $\lambda=2$, all inequalities become equalities, and the corresponding triangle is equilateral. In general $1\le \lambda \le 2$. 
We then put 
\begin{equation*}
X_{\lambda}:=\{ \text{triangles with measure equal to $\lambda$}\}.
\end{equation*}
For instance, $X_1$ is the set of all degenerate triangles and $X_2$ is the set of all equilateral triangles and every other triangle lies somewhere between these two extremes.\\
The class $X_1$ is not useful for our approach, as it consists of collinear points. In such a cases, the  value $\rho=1$ can always be assigned, since these points form a tripod with the middle one as the center regardless of the geometry of the underlying space. As a result, this class cannot effectively distinguish between different metrics. \\
In contrast, the class $X_2$, which corresponds to equilateral triangles,  is the most significant. As already noted,  $\rho$ attains its maximum value for triangles belonging to this class in Euclidean spaces and circles. We thus made the choice of only using equilateral triangles in our algorithm. 
\section{Computing curvature profile of networks and metric spaces}
\begin{figure}[H]
    \centering
    \graphicspath{{images/metric_datasets/intuitive_examples/}}
    \subfigure[]{\includegraphics[width=0.4\textwidth]{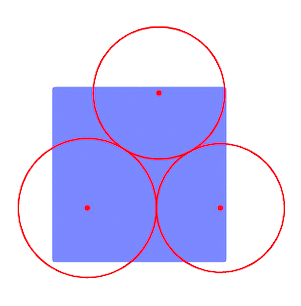}}
    \subfigure[]{\includegraphics[width=0.4\textwidth]{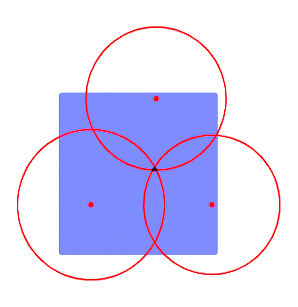}}
    \newline
    \subfigure[]{\includegraphics[width=0.4\textwidth]{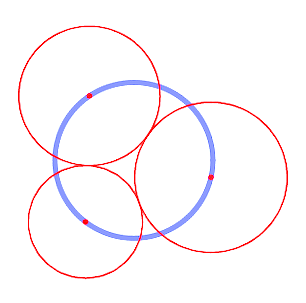}}
    \subfigure[]{\includegraphics[width=0.4\textwidth]{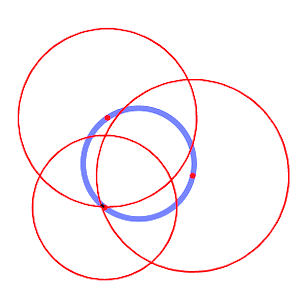}}
       \caption{Intersection pattern of three balls, which initial pairwise intersect on their boundaries, on the: Euclidean plane (first row) and circle (second row). }
    \label{fig:intersection}
\end{figure}

Using the curvature concept just defined to examine the geometry of finite metric spaces, in particular empirical networks, we first establish the algorithm for  unweighted graphs,  where the graph distance yields an integer-valued metric on the vertex set. In our pipeline, we employ a modified version of Dijkstra’s algorithm \cite{2020SciPy-NMeth},  enhancing computational efficiency and robustness, to compute the distance matrix $D = \bigl(d_{ij}\bigr)_{1 \le i,j \le N}$.  When two vertices belong to different connected components and no path exists between them, instead of putting their distance $= \infty$,  we simply use a sufficiently large finite value, specifically $100$ times the maximal finite distance. \\
 When, more generally, our dataset is a discrete metric space $(X,d)$, we replace it with a graph with vertex set $X$ that models the neighborhood properties of $X$. Two vertices are connected by an edge when their distance, given by the metric $d$  is sufficiently small. We can use either an $\epsilon$-neighborhood graph or a symmetric $k$-neighborhood graph. If the edges are weighted by the distances between their endpoints, the resulting graph distance provides an approximation of the intrinsic distance in $X$, c.f. \cite{Tenenbaum00}. \\
 The function $\rho$ is a three-point base function, returning values in the interval $[1,2]$.  We will examine how the factor $\rho$ defined above evolves from small to large such triangles, tracing sectional curvature from local neighborhoods to global and asymptotic scales. For simplicity and consistency, we restrict our computation to triangles of a fixed shape,   equilateral ones, recalling that on both the circle and the Euclidean plane, $\rho$ reaches its maximum value, $2$ and $\frac{2}{\sqrt{3}}$ resp., for equilateral triangles.  In hyperconvex spaces, the value of $\rho$ is always $1$ for any triangle. \\ 
We evaluate triangles of side lengths $2r\le diam(X)$ (the maximal pairwise distance in the graph). 
For efficiency and lower computation complexity, we apply the algorithm to a random sample of data points. For side length $2r$ we first filter all  points contained in at least one equilateral triangle of side $2r$, from which we then randomly sample a subset $S$ of size $m \times N$,  $N$ being the total number of points in  $X$. Afterwards, for each point in $S$ our algorithm starts the search for triangles and stops as soon as finds one.  So at each scale $r$, we will have maximum number of $|S|$ of triangles. We chose $m = 0.1$  as it balances result accuracy and computational cost.
  A hybrid approach that combines clustering with sampling,  for various number of clusters and sample sizes, did not produce significant changes in the resulting curvature profiles. \\ 
For computing $\rho$, instead of directly minimizing \cref{rho}, we adopt a strategy inspired by the persistent homology method in topological data analysis, as discussed in 
\cite{Edelsbrunner08, Zomorodian04}. This approach traces the intersection patterns of three balls centered at the vertices of each triangle as their radii are scaled. For each such  equilateral triangle, we place a ball at each vertex with an initial radius $r_{in} = r$ and incrementally increase its radius until at $r_ {out}$ a common intersection, termed the "triplet intersection", is detected.  The expansion factor $\rho$  is then given by 
 \begin{equation*}
 \rho = \frac{r_{out}}{r_{in}}.
 \end{equation*}
To demonstrate how geometry influences these intersection patterns, we provide an illustrative example, visualized in \ref{fig:intersection}. We first place the three points on a flat Euclidean plane, and then on a circle. Initially, balls are centered at each point with radii determined by Gromov products. As the radii scale up, we observe how the pairwise intersections of the balls evolve into a triple intersection. The geometric differences between the plane and the circle become apparent as the scaling progresses. The result, visualized in Figure \ref{fig:intersection},  highlights how the scaling behavior differs between flat and highly curved spaces. This process illustrates the shift from flat geometry to a space with strong positive curvature, offering insight into how the underlying geometry affects  $\rho$ in our approach.
This algorithm returns the plot of $\rho$ (vertical axis) against the parameter $r$ (horizontal axis), $r$ being half the side length of equilateral triangles in the graph. As we have multiple triangles at each scale, which may attain different values of $\rho$, we use a marker at the average or typical value of $\rho$  at that scale, its size encoding the number of equilateral triangles found at the corresponding scale $r$. Thus, larger markers represent more frequent occurrences. \\
The value of the function $\rho$ can be used to establish curvature inequalities between metric spaces. To phrase it more explicitly, consider two metric spaces  $(X,d_X)$ and $ (Y,d_Y)$. We say that the curvature of $X$ is no greater than that of $Y$, if  for any triples $x_1,x_2,x_3 \in X$ and $y_1,y_2,y_3 \in Y$ with $d_X(x_i,x_j)=d_Y(y_i,y_j)$, for all $i,j$, the following inequality holds
\begin{equation}\label{rhocomp}
\rho_X(x_1,x_2,x_3)\le \rho_Y(y_1,y_2,y_3),
\end{equation}
where value of $\rho$ in $x$ and $Y$ are computed using the respective distance functions $d_X$ and $d_Y$.  \\
As reference spaces, we consider Euclidean space, a cycle, and a hyperconvex space. Their corresponding $\rho$-values appear in the plots as three dashed horizontal lines: the yellow line at $\rho=1$ reflects tree-like behavior, the purple line at $\rho=\frac{2}{\sqrt{3}}$ corresponds to the Euclidean case, and the red dashed one at $\rho=2$ represents the configuration of three equidistant points on a circle. \\
For an empirical dataset, first constructing the neighborhood graph  effectively employs distance information to infer the curvature profile. We prefer such a graph-based representation to the initial metric-based one because distance functions in data often are not trustworthy beyond local neighborhoods, as they efficiently approximate similarities only within a close vicinity.  
To ensure that the constructed graph model accurately reflects the underlying data density, as the $k$-neighborhood graph model disregards how densely populated the neighborhood of each point is, we employ a density-aware graph construction method, called \verb|adaptive_graph_construction|. This approach is governed by two hyperparameters, $k_{\text{min}}$ and $k_{\text{max}}$, which define the minimum and maximum number of neighbors considered for each data point. For each point, local density is estimated as the inverse of the distance to its initially captured neighbors. We then compute the global minimum and maximum density values and normalize each point’s density to obtain a \textit{normalized density score}. Based on this score, we adaptively determine the number of neighbors $k$ for each point by interpolating between $k_{\text{min}}$ and $k_{\text{max}}$. For the graph construction, we used the \textit{NetworkX} package  \cite{networkx} and applied the Nearest Neighbors class from the \textit{scikit-learn} library \cite{scikit-learn}.\\
The pseudo-algorithm for computing the curvature profile $\rho$ is presented in \cref{alg:sectional-curvature}.

 \begin{algorithm}[H]
\floatname{algorithm}{Algorithm}
\caption{Discrete Sectional Curvature Computation}
\label{alg:sectional-curvature}
\begin{algorithmic}[1]
\Require Graph $G$ or point cloud data
\Ensure Sectional curvature values for sampled triangles

\If{input is not a graph}
    \State Construct graph $G$ using nearest-neighbor method
\EndIf

\State Compute distance matrix $D$ using combinatorial metric
\State $d_{max} \gets \max(D)$ \Comment{Maximum pairwise distance}

\If{clustering enabled}
    \State Partition vertices into $k$ clusters
    \State Sample $n$ points from each cluster
    \State $V_{subset} \gets$ sampled vertices
\Else
    \State $V_{subset} \gets V(G)$
\EndIf

\For{possible side lengths $l \in [1, d_{max}]$}
    \State $T_l \gets \emptyset$ \Comment{Set of triangles with side length $l$}
    \For{vertex triplets $(v_1, v_2, v_3)$ in $V_{subset}$}
        \If{forms equilateral triangle with side length $l$}
            \State $T_l \gets T_l \cup \{(v_1, v_2, v_3)\}$
        \EndIf
    \EndFor
    
    \For{triangles $t \in T_l$}
        \State $r_{init} \gets l/2$ \Comment{Initial ball radius}
        \State $r_{intersect} \gets$ \Call{FindTripleIntersection}{$t$}
        \State $\rho_t \gets r_{intersect}/r_{init}$ \Comment{Scale of expansion}
        \State Store curvature value for triangle $t$
    \EndFor
\EndFor

\Procedure{FindTripleIntersection}{triangle $t$}
    \State $r \gets l/2$ \Comment{Start with initial radius}
    \While{balls do not intersect}
        \State Increment $r$
        \State Update balls centered at triangle vertices
        \State Check for triple intersection
    \EndWhile
    \State \Return $r$
\EndProcedure

\end{algorithmic}
\end{algorithm}
\section{Results}
\subsection{Curvature Profile for Networks}
We first apply our method to obtain the curvature profile $\rho$ for several networks, including model classes such as  Erd\"os-R\'enyi (ER) \cite{Erdos59} and Watts-Strogatz(WS)  \cite{Watts98} networks (both with $1000$ nodes and an average degree $4$)
, as well as real-world networks such as the Yeast protein interaction ($1458$ nodes and $1948$ edges) \cite{alexandervc_yeast_ppi_2022} and the US Power Grid ($4941$ nodes and $6594$ edges) \cite{watts1998collective},  Facebook friendship,  Karate \cite{Rogel-Salazar2019,nr} and Football club networks \cite{kunegis2013konect,girvan2002community}. Our curvature profiles provide a distribution of triangles characterized by two parameters $(r,\rho)$, where $r$ denotes half the side length of equilateral triangles and $\rho$ represents the average (and non-averaged values) of associated expansion factors quantifying curvature. 
The plots of averaged and non-averaged $\rho$ and the box plots for model networks and empirical networks are presented in \cref{tab:modelnet} and \cref{tab:empnet}. \\
To investigate the effect of size and sparsity parameters on the curvature profile, we apply our method on ER and WS networks as we vary the size and average degree. For small networks, we choose $m=1$ in the choice of the sample size and for larger ones (constructed on $10,000$ nodes) we sample a subset of $1000$ triangles at each parameter $r$. The results presented in \cref{tab:modelnetpar} for averaged $\rho$ profile indicates that the geometry of ER-networks is stable as we scale the network, while WS-networks increasingly resemble tree-like geometries at larger scales.\\
Applying the curvature profile to real-world networks also reveals dependencies on the regions of the network from which triangles are sampled. In particular, the power grid network displays a more scattered distribution of $\rho$ values across the full range of the scale $r$. Nevertheless, the plot of averaged $\rho$, shown in the second row of \cref{tab:empnet}, indicates that the overall geometry of the power grid network is predominantly flat, accurately reflecting the underlying structural design of this network.

\begin{table}[H]
 \caption{Curvature profile for model networks Net-size-sparsity on each row.}
  \label{tab:modelnet}
\centering
\begin{tabular}{ c | c c c}
    \toprule
      \multicolumn{2}{c}{averaged}   & non-averaged &  boxplot \\		
      ER-1000-4    & \includegraphics[width=0.25\linewidth]{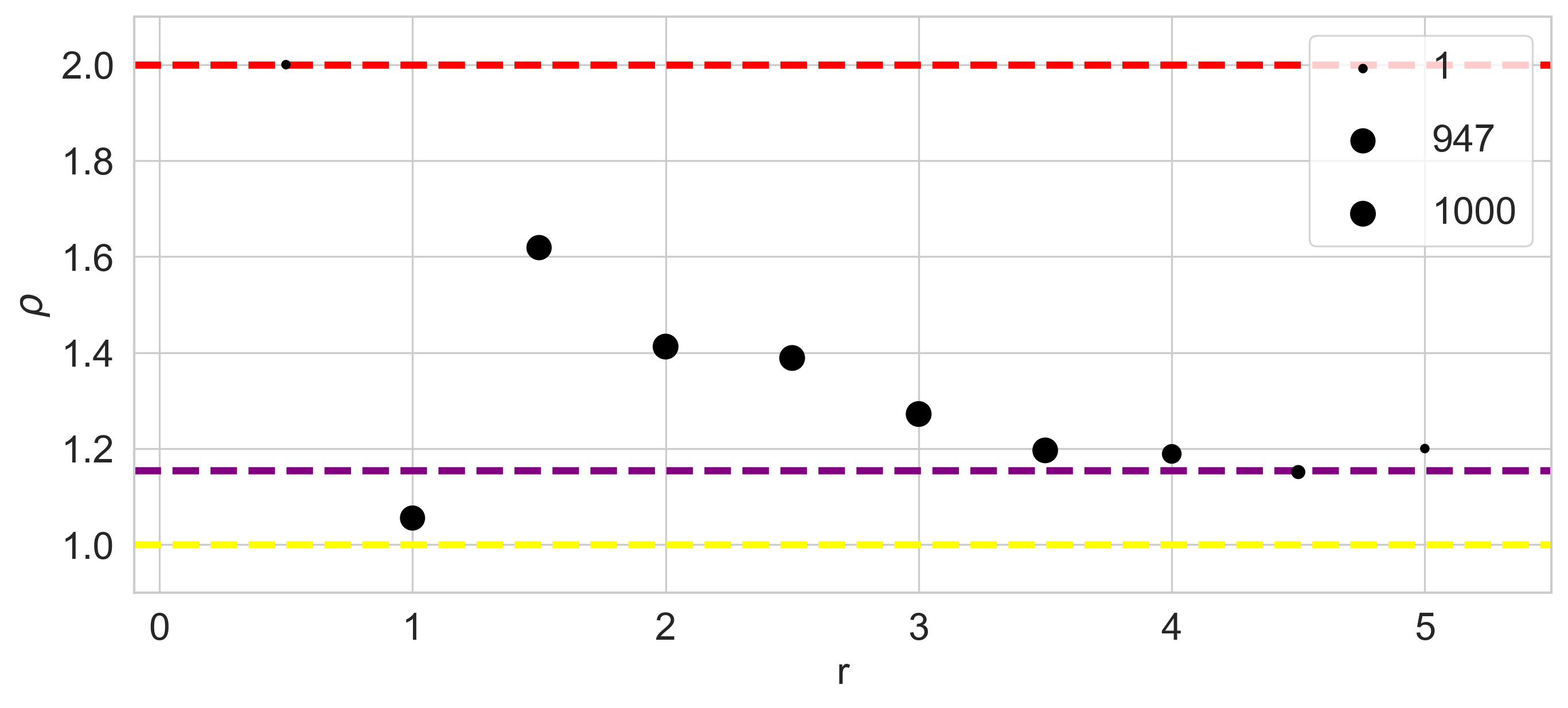}
    & \includegraphics[width=0.25\linewidth]{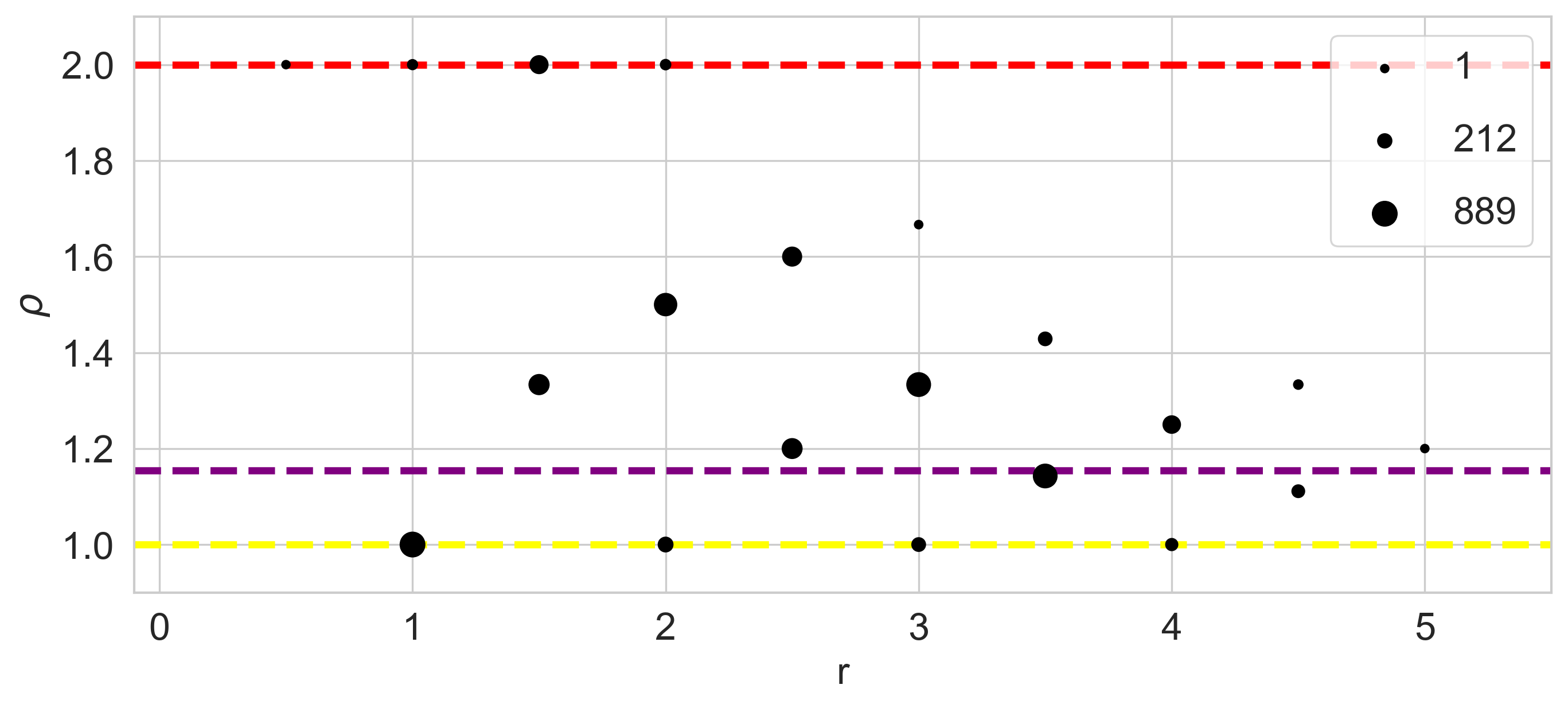}
      & \includegraphics[width=0.25\linewidth]{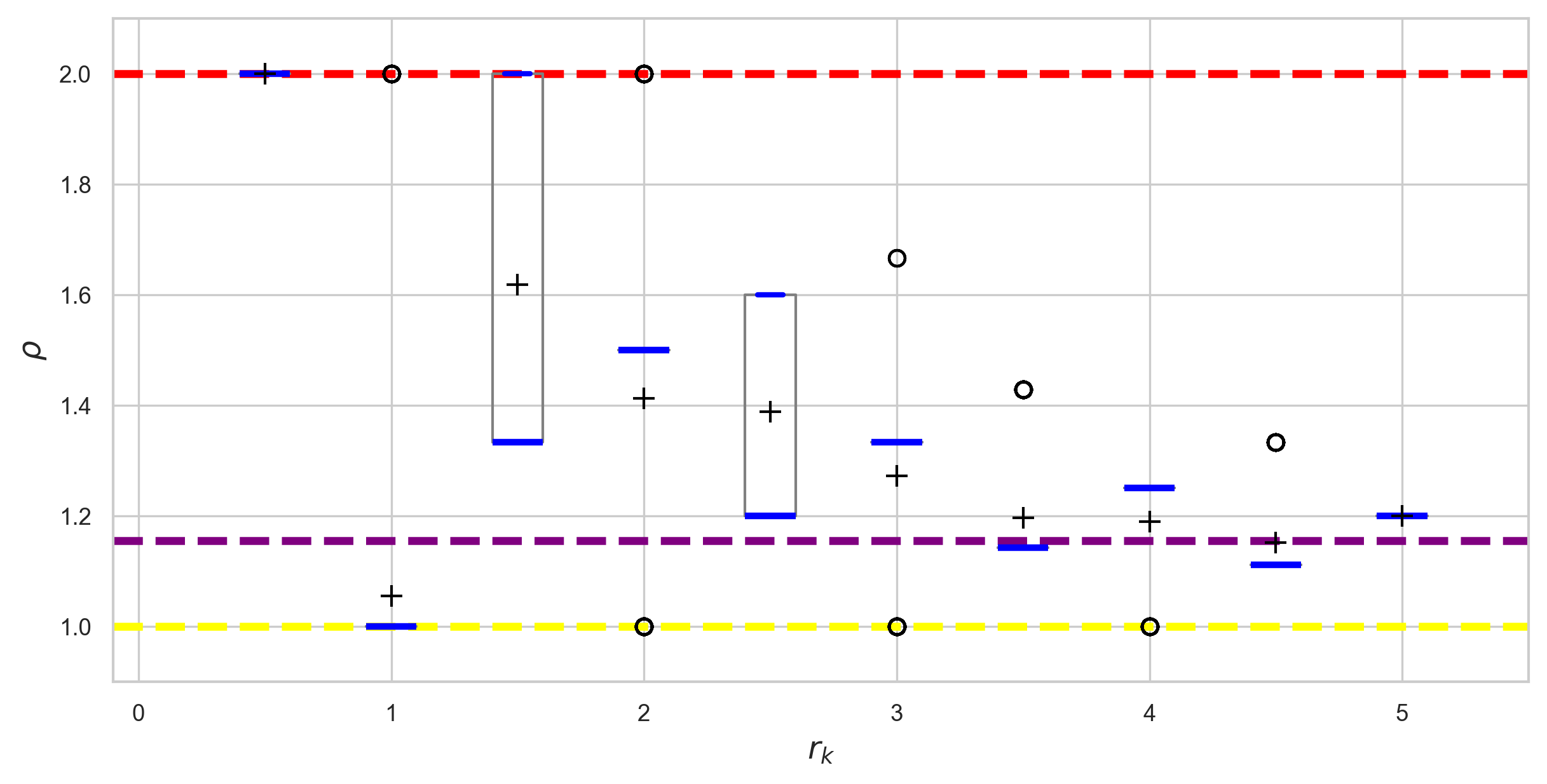}         
    \\
    \midrule
      WS-1000-4    & \includegraphics[width=0.25\linewidth]{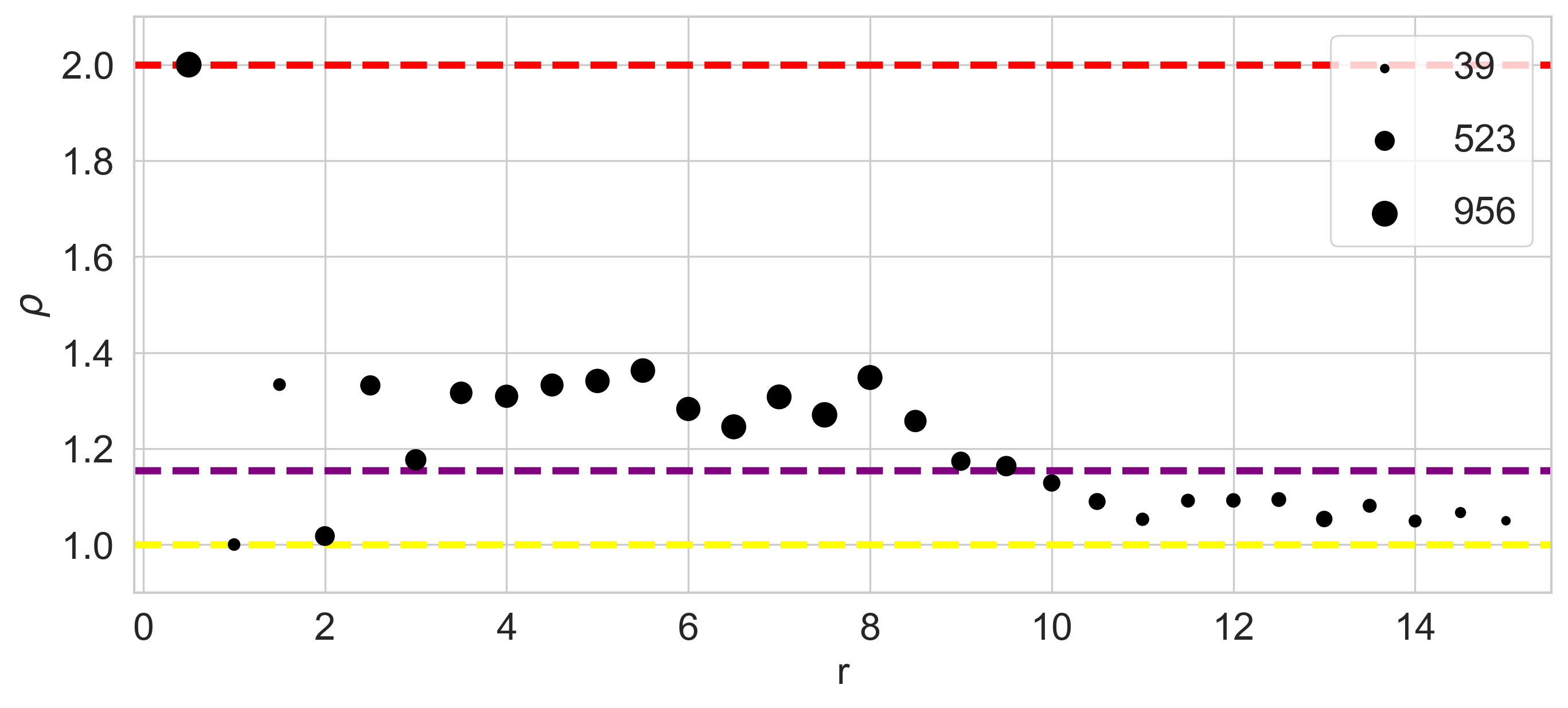}
    &  \includegraphics[width=0.25\linewidth]{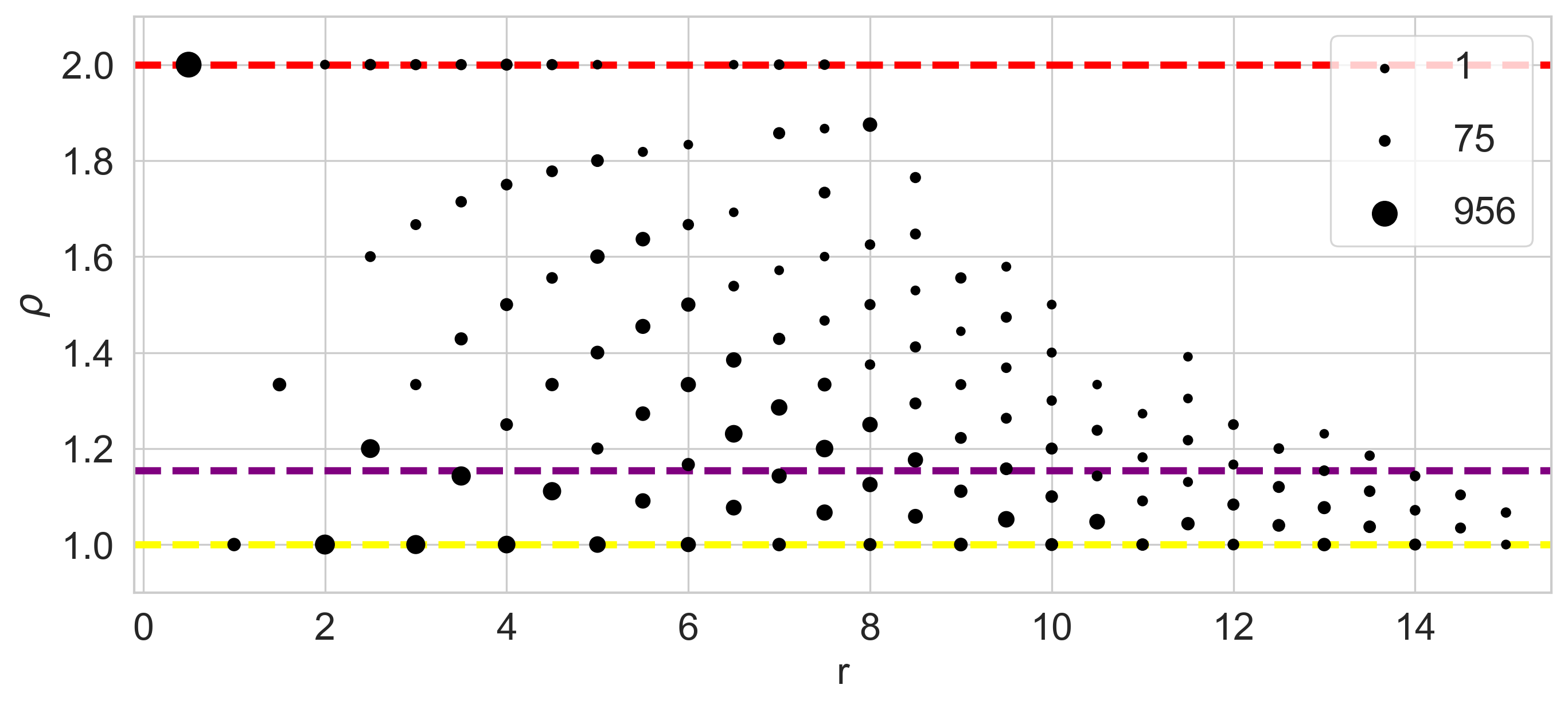}
      &  \includegraphics[width=0.25\linewidth]{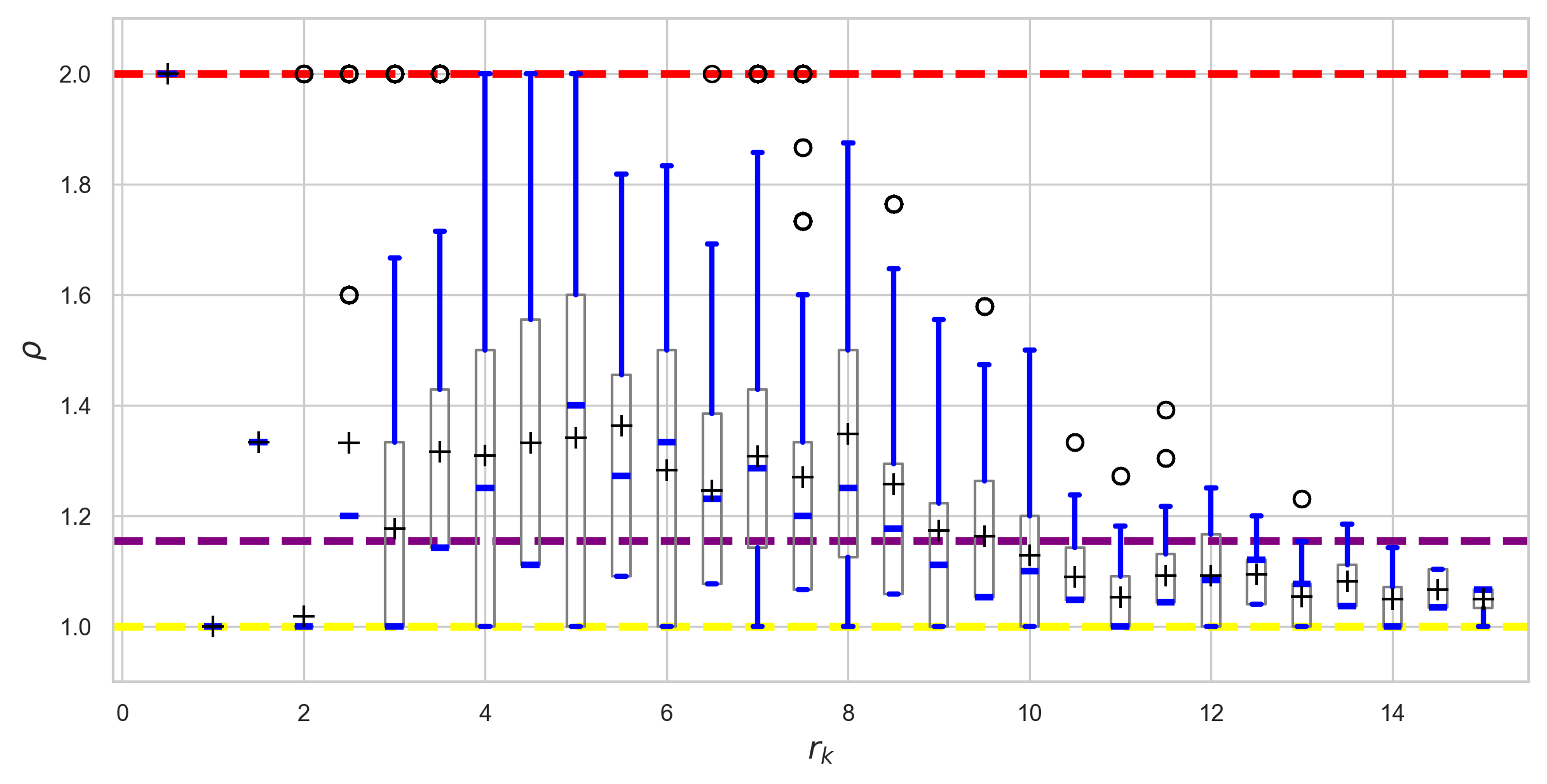} \\ 
     \bottomrule
     \end{tabular}
 \end{table}
\begin{table}[H]
 \caption{Curvature profile for empirical networks.}
  \label{tab:empnet}
\centering
\begin{tabular}{ c | c c c}
    \toprule
      \multicolumn{2}{c}{averaged}   & non-averaged &  boxplot \\		
       \midrule
       Yeast   & \includegraphics[width=0.25\linewidth]{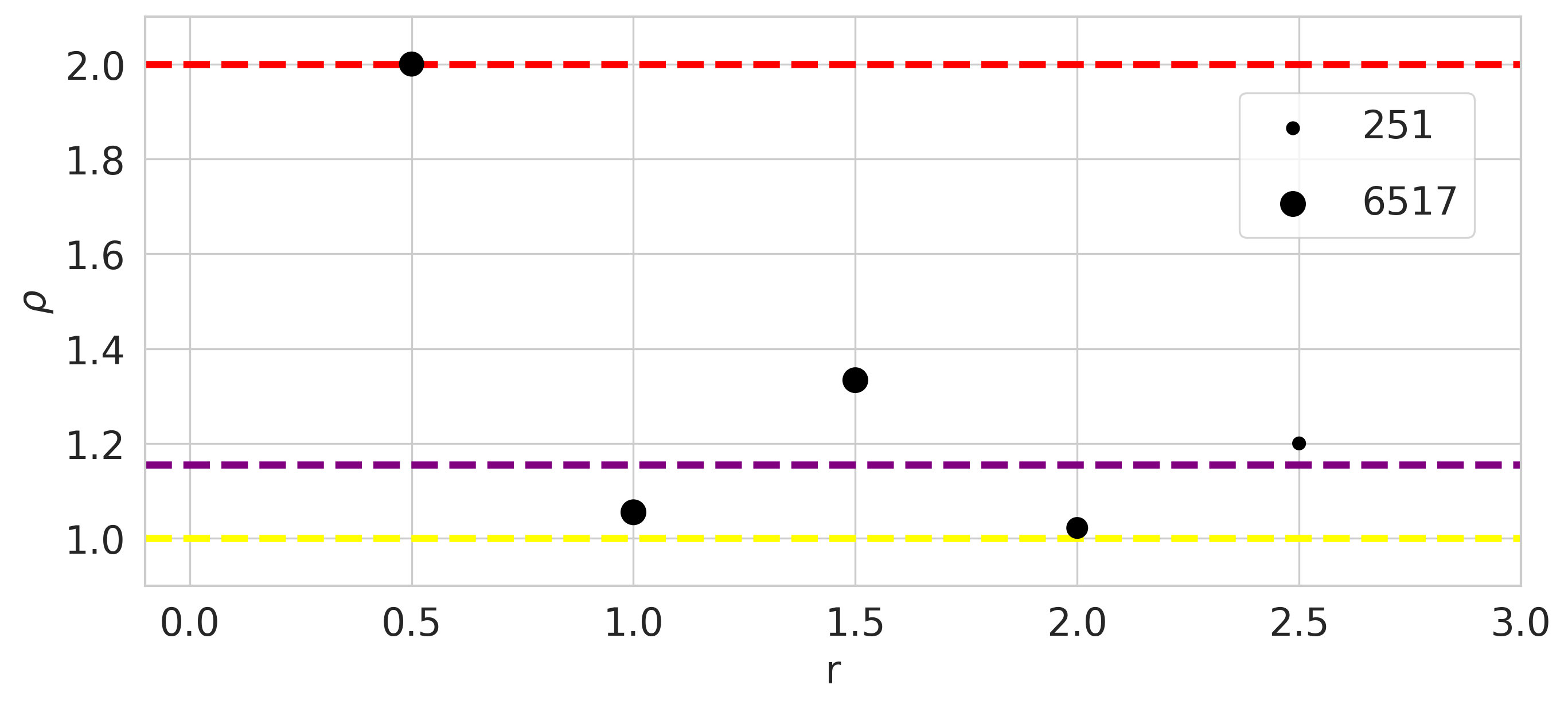}
      & \includegraphics[width=0.25\linewidth]{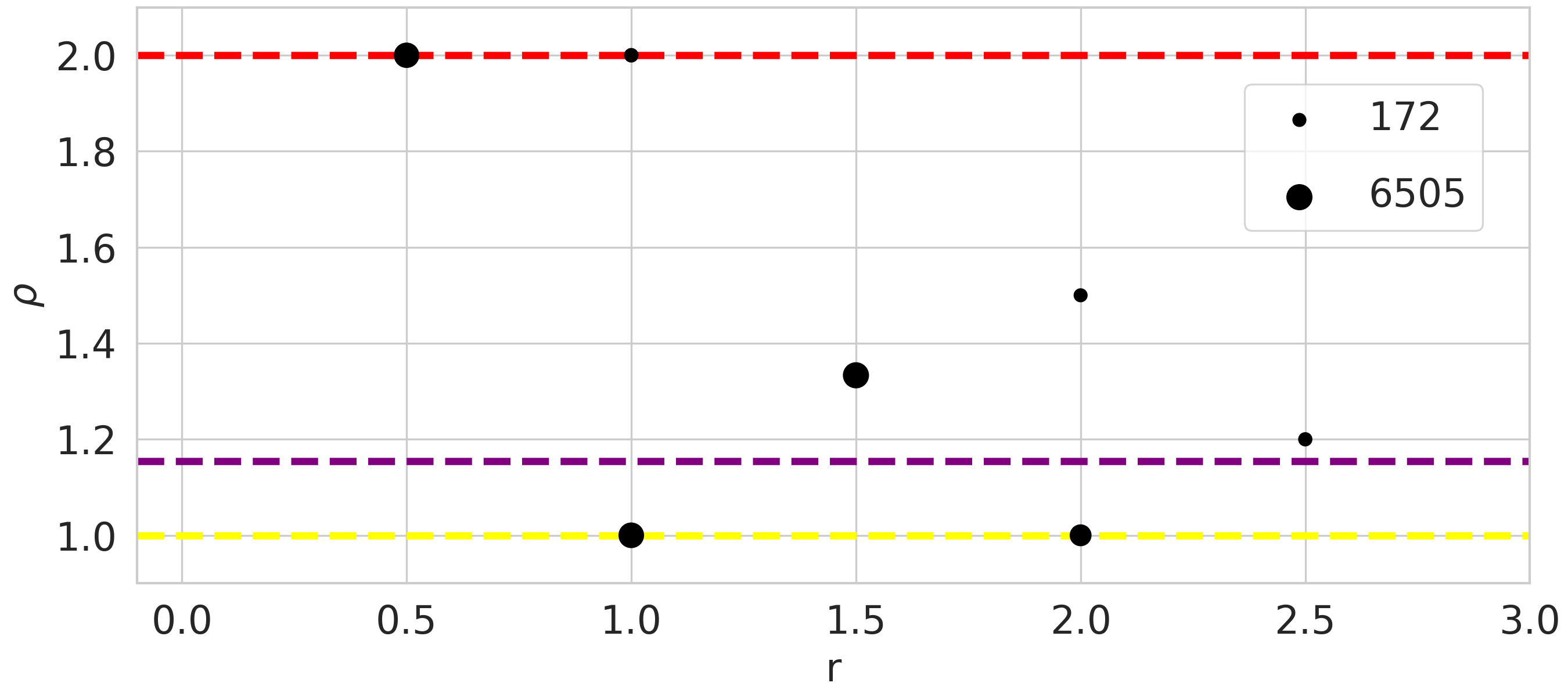}
      &  \includegraphics[width=0.25\linewidth]{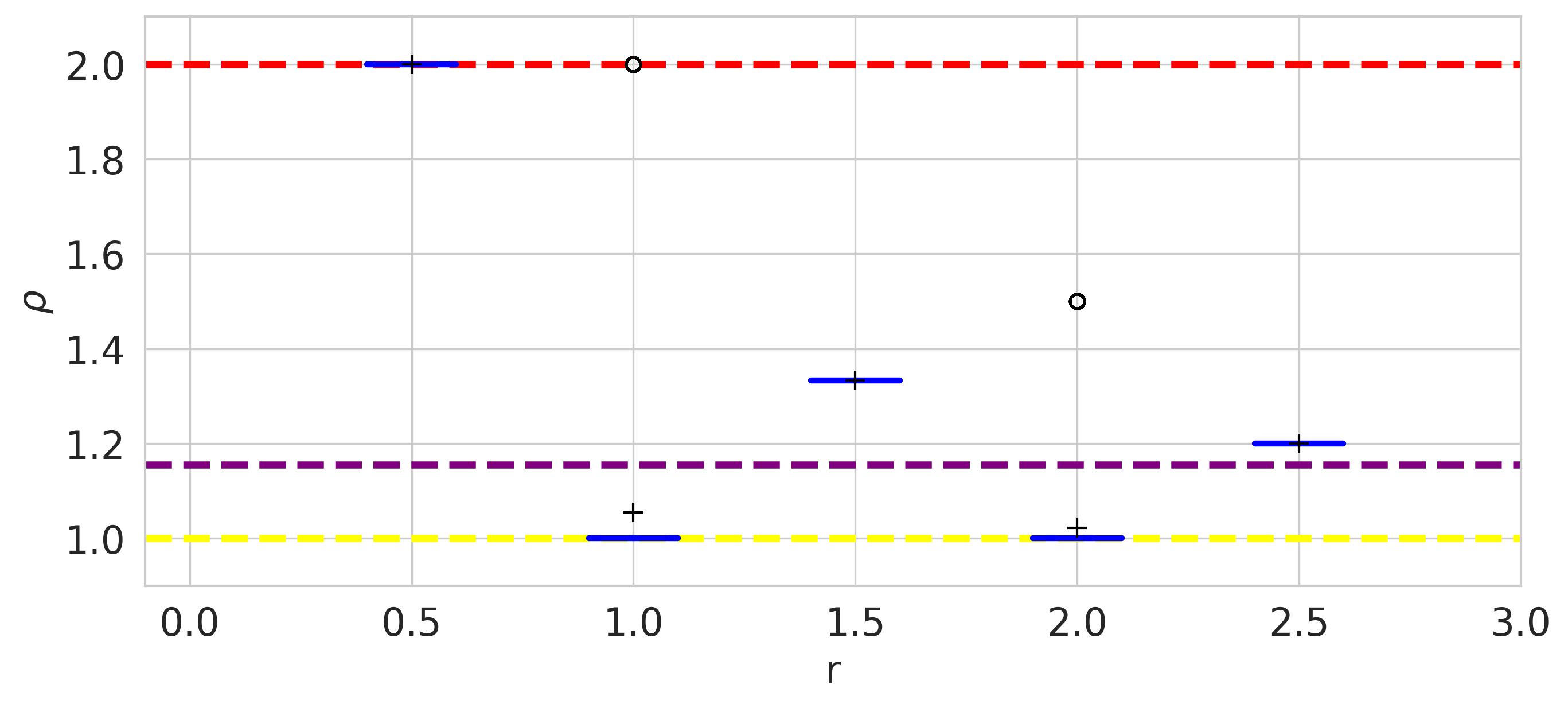} 
      \\ 
       \midrule 
       Power-grid   & \includegraphics[width=0.25\linewidth]{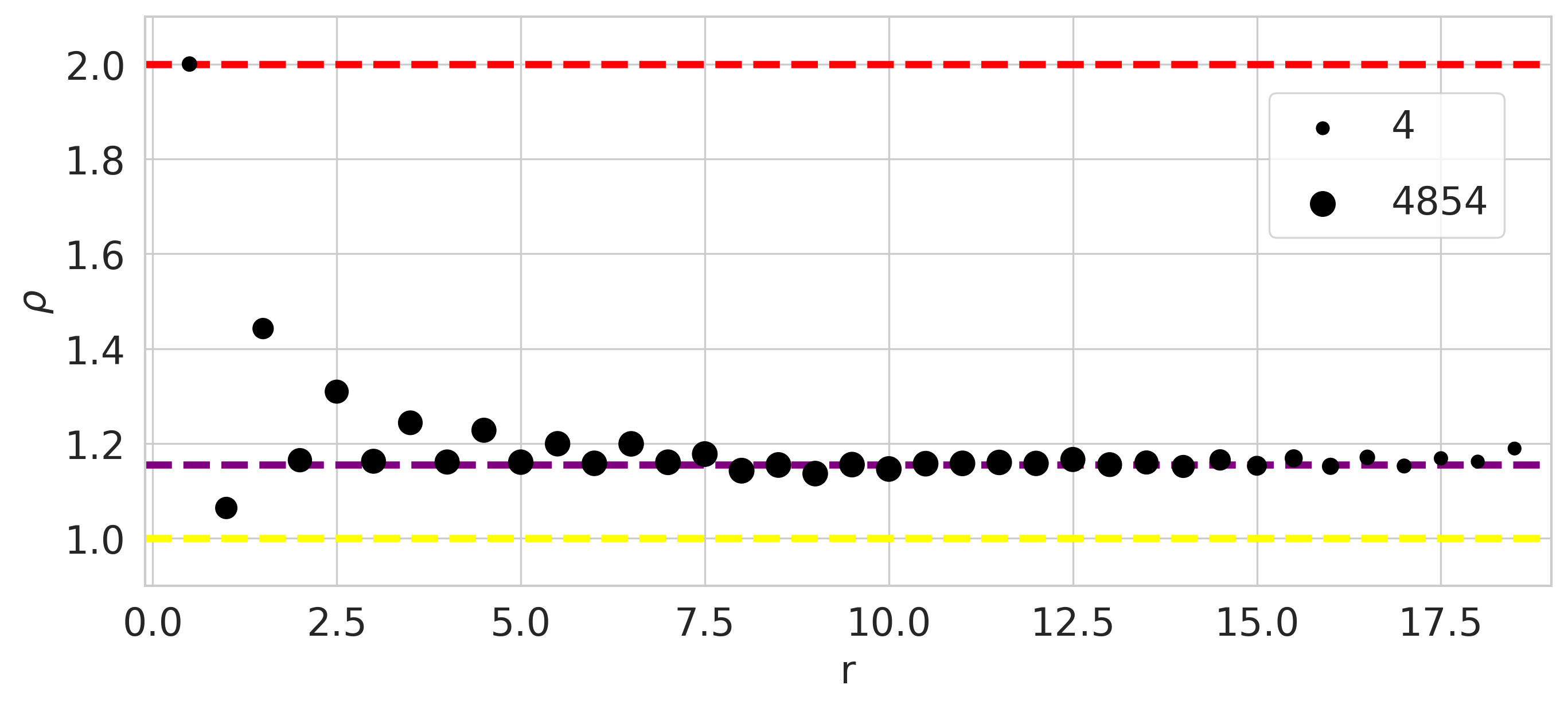}
       & \includegraphics[width=0.25\linewidth]{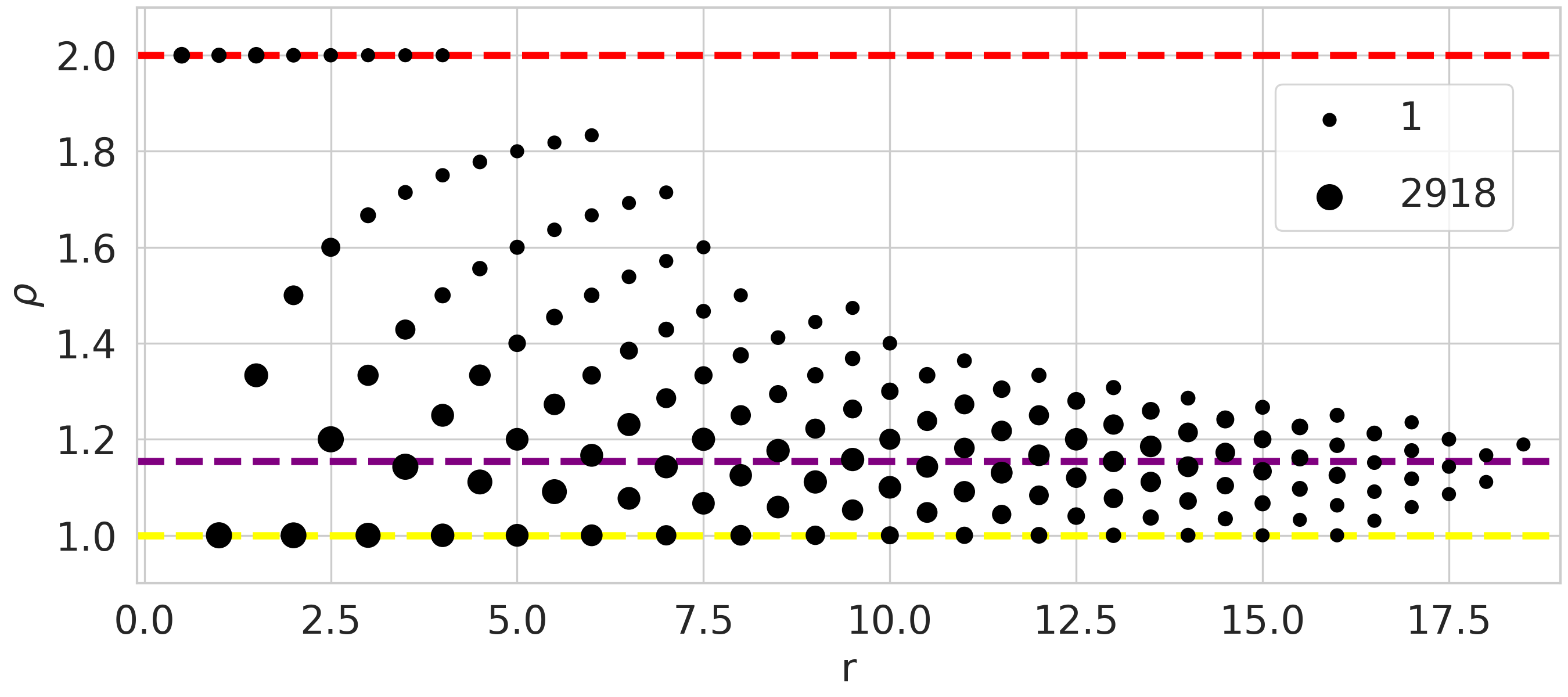}
      &  \includegraphics[width=0.25\linewidth]{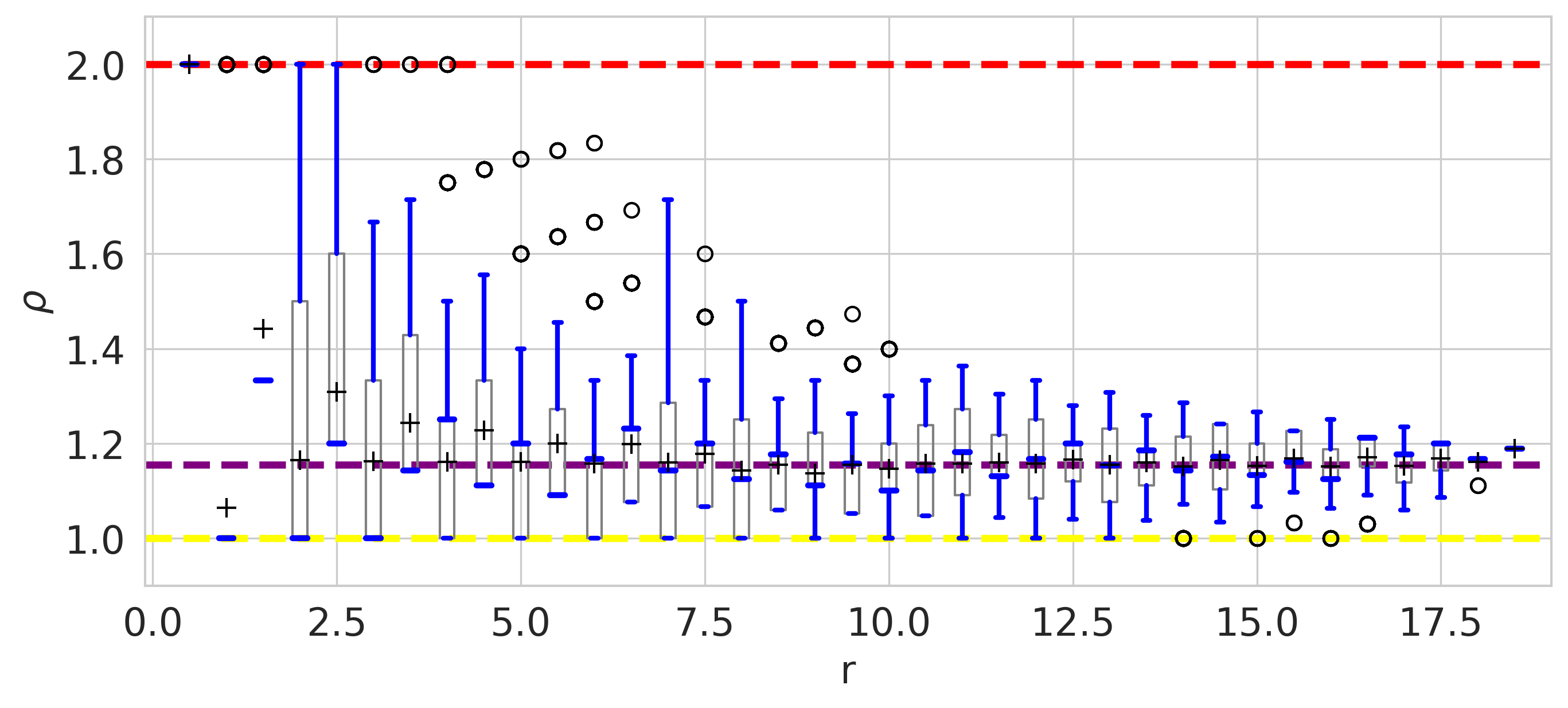} 
      \\ 
      \midrule 
       Facebook  & \includegraphics[width=0.25\linewidth]{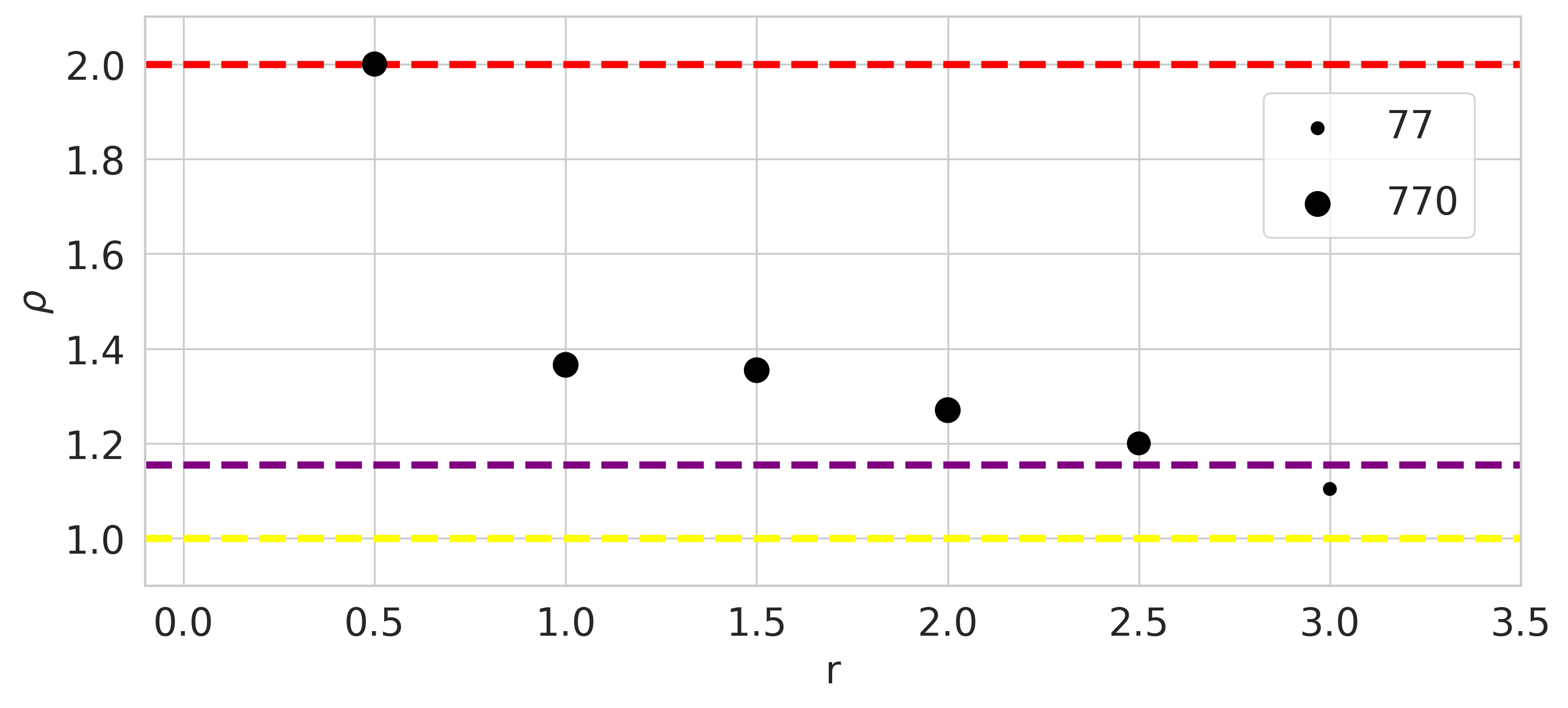}
       & \includegraphics[width=0.25\linewidth]{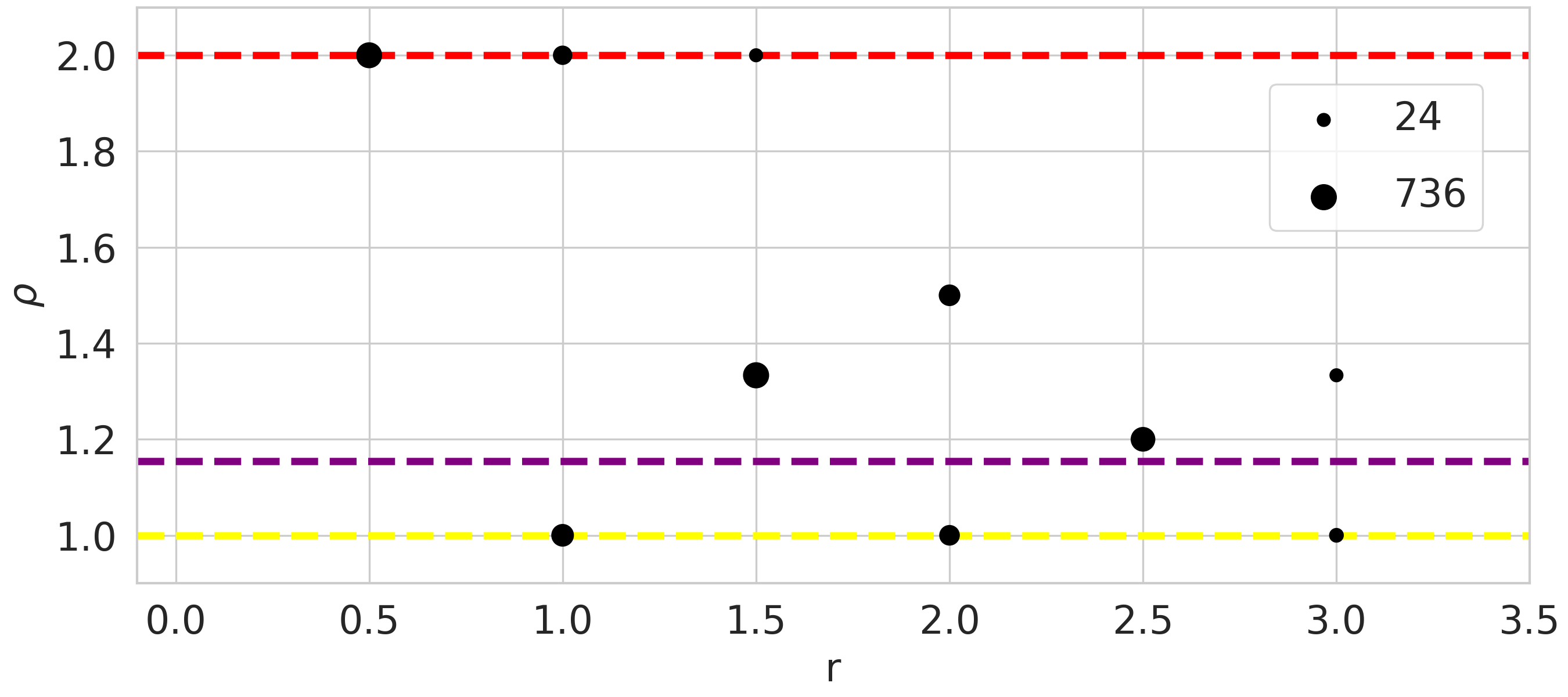}
      &  \includegraphics[width=0.25\linewidth]{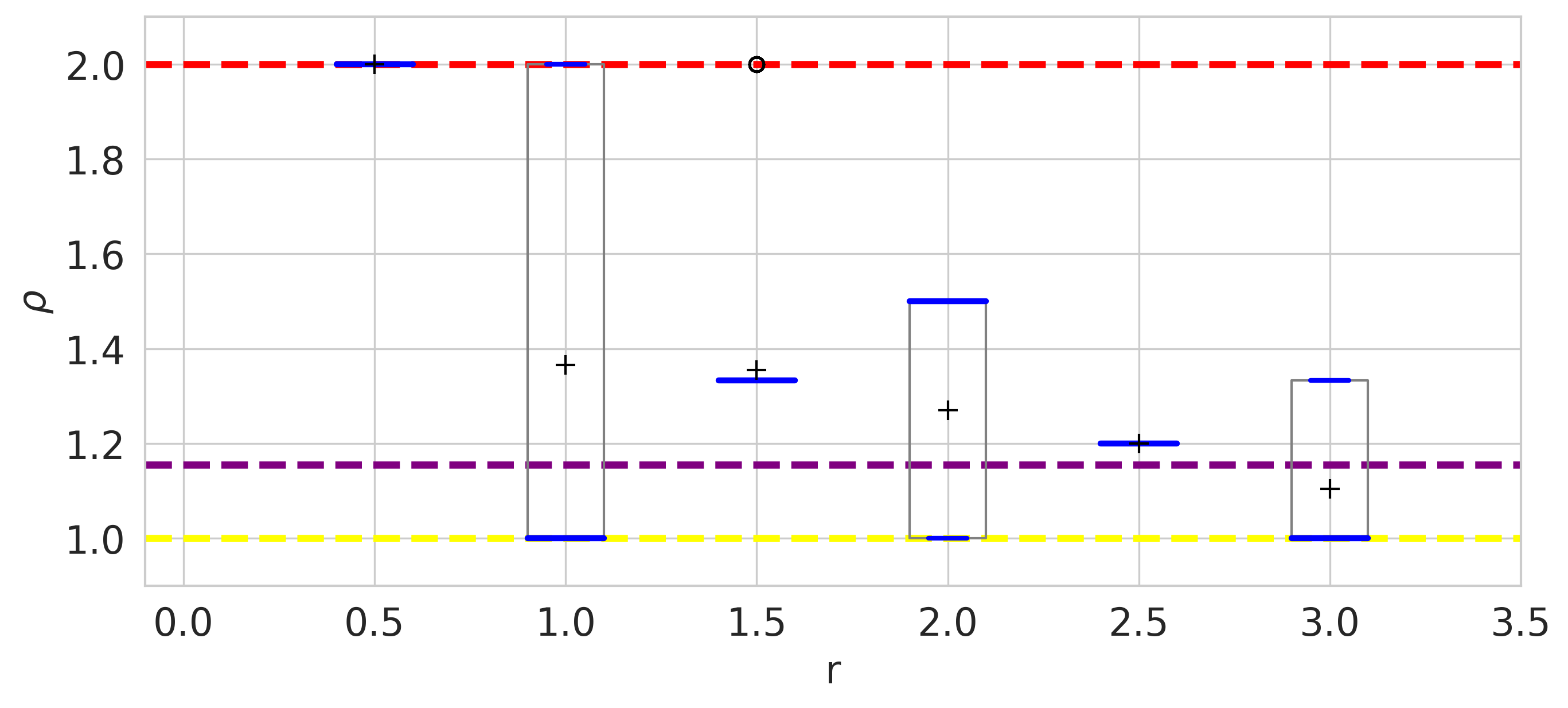} 
      \\ 
      \midrule 
       Football  & \includegraphics[width=0.25\linewidth]{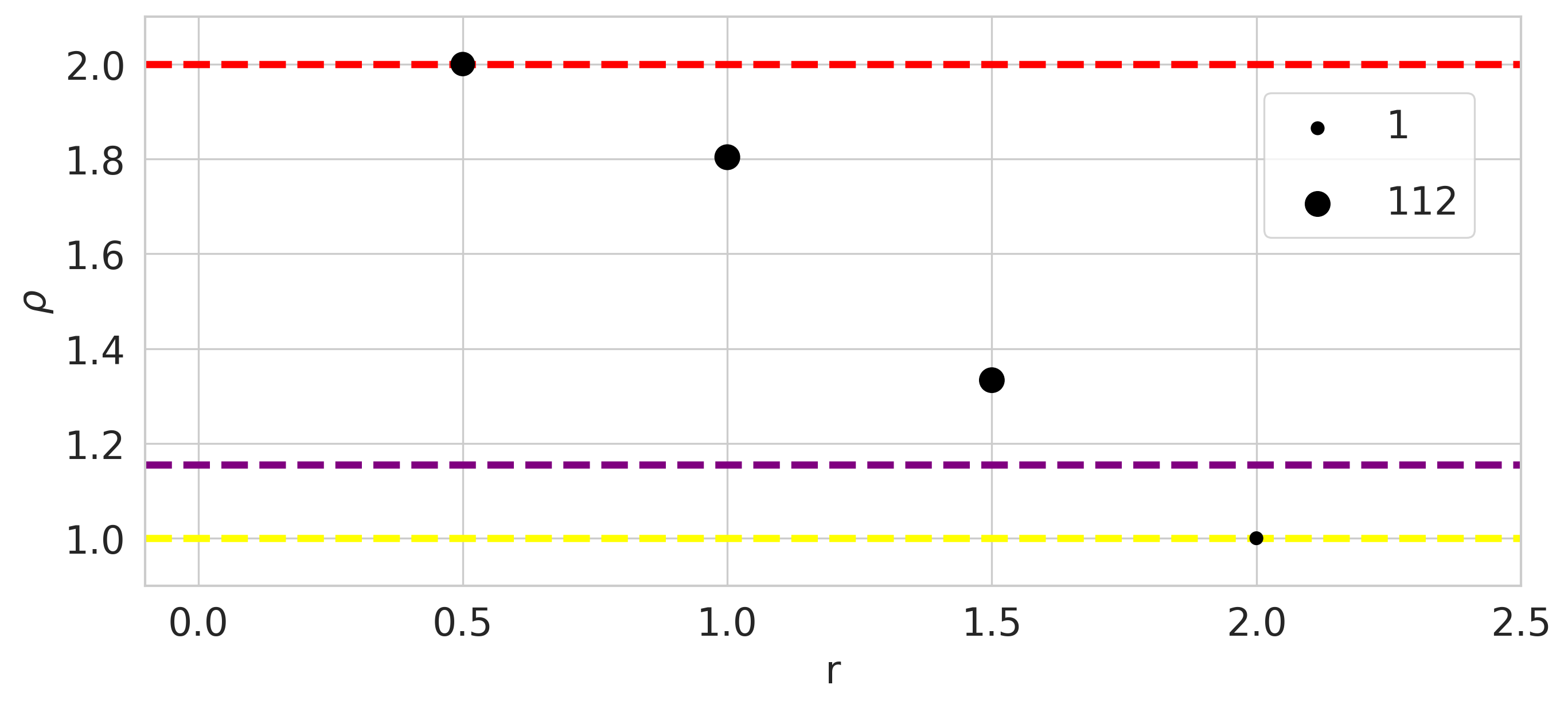}
       & \includegraphics[width=0.25\linewidth]{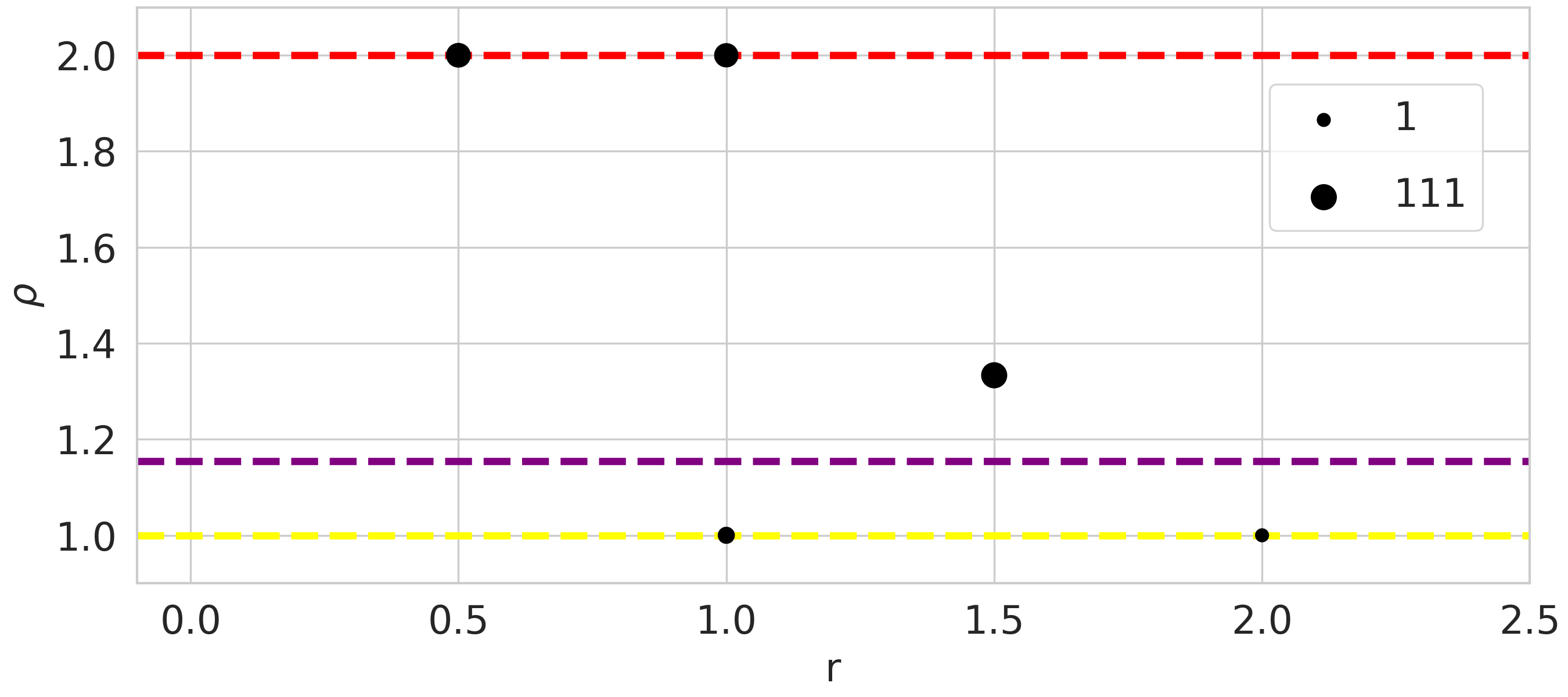}
      &  \includegraphics[width=0.25\linewidth]{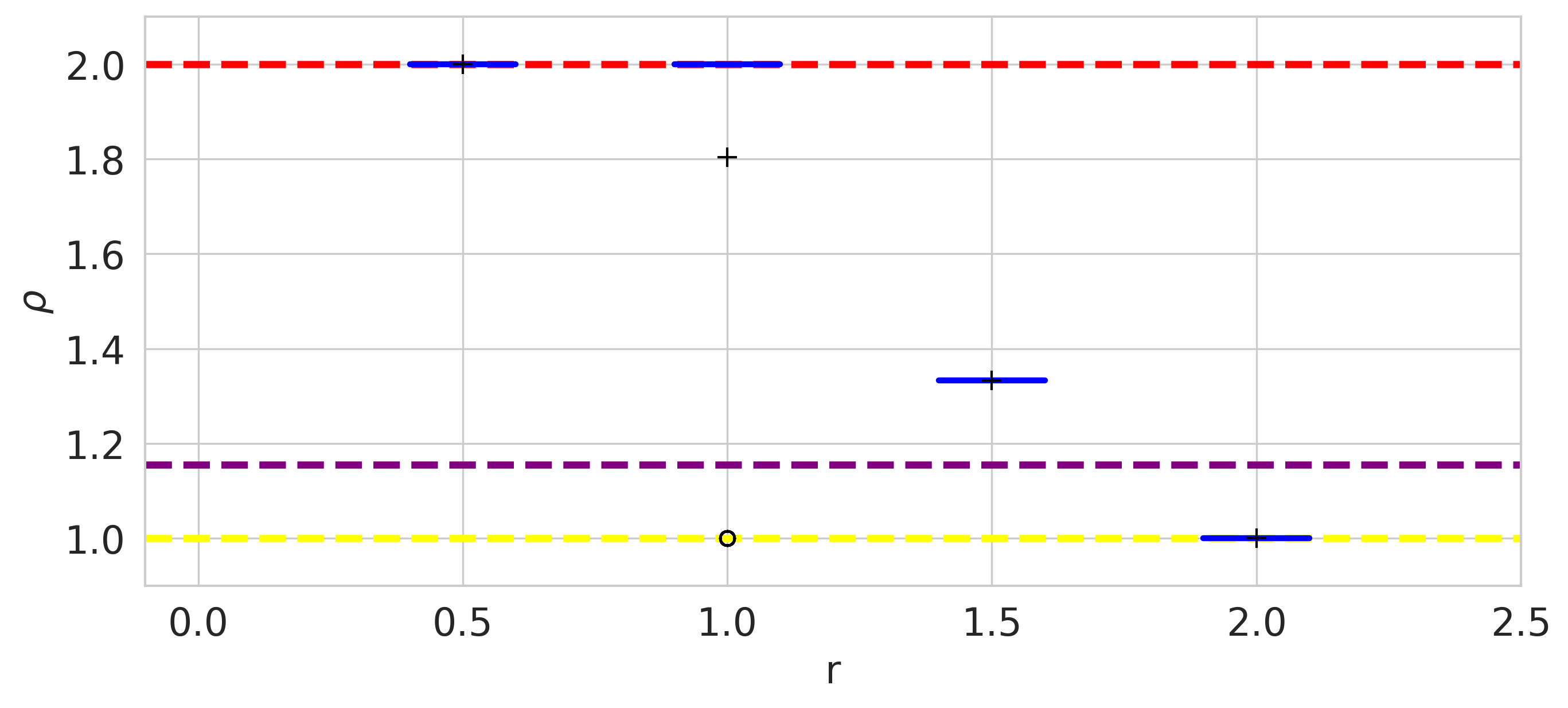} 
      \\ 
      \midrule 
       Karate  & \includegraphics[width=0.25\linewidth]{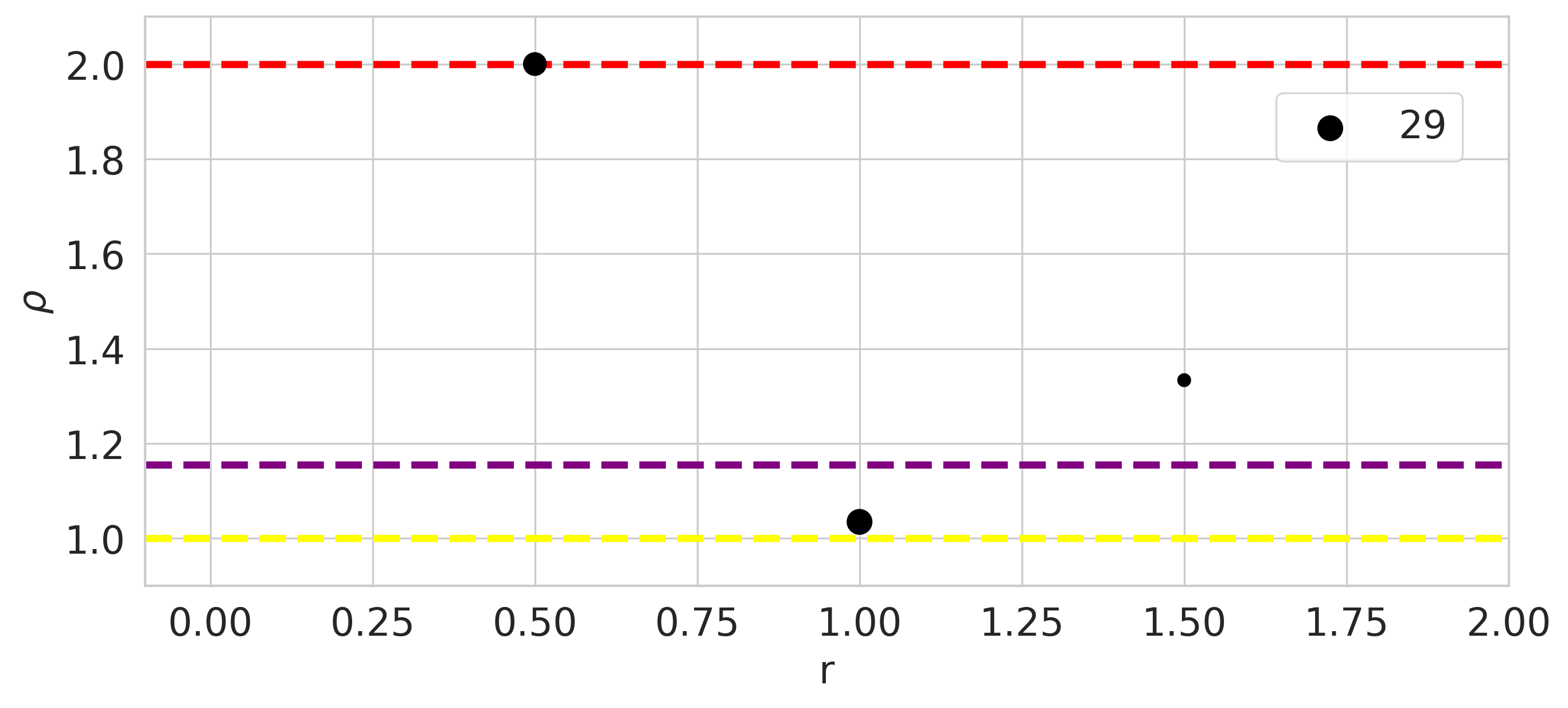}
       & \includegraphics[width=0.25\linewidth]{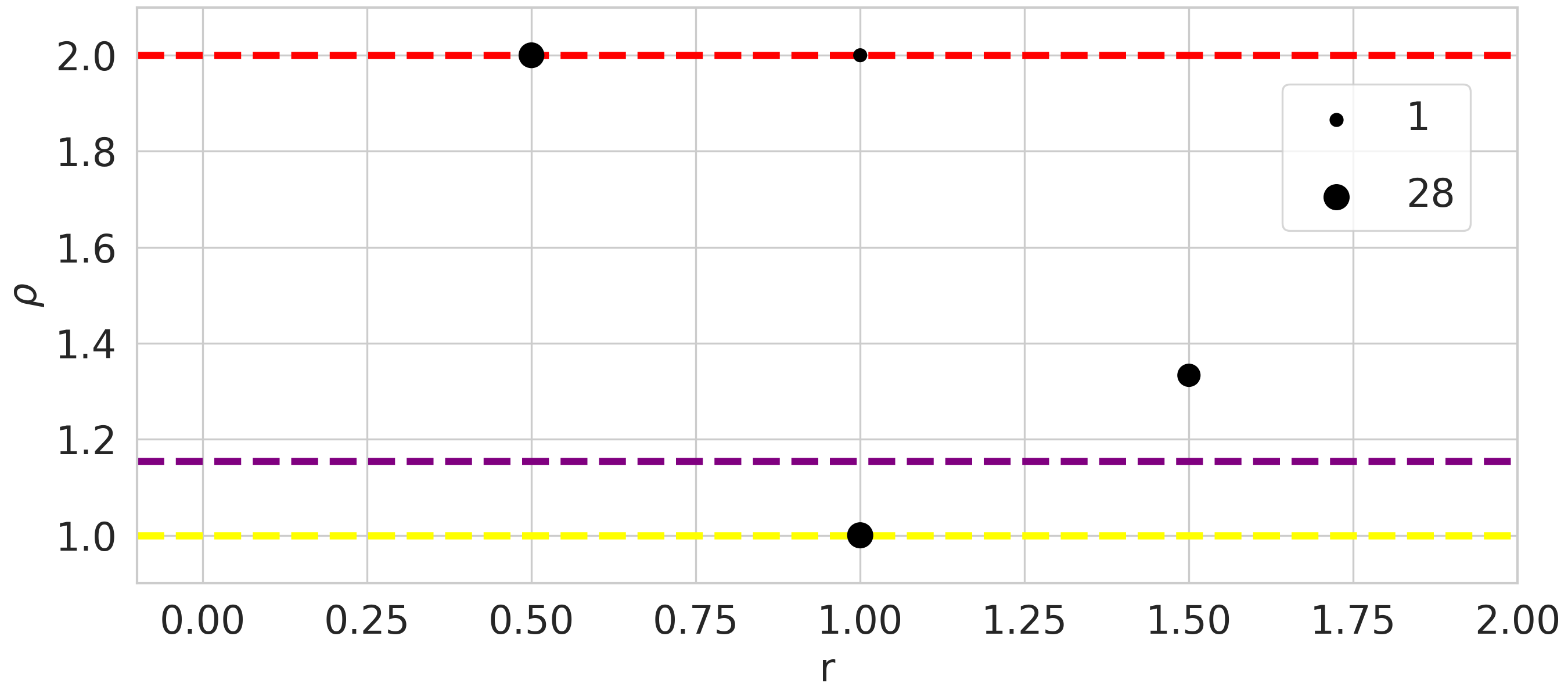}
      &  \includegraphics[width=0.25\linewidth]{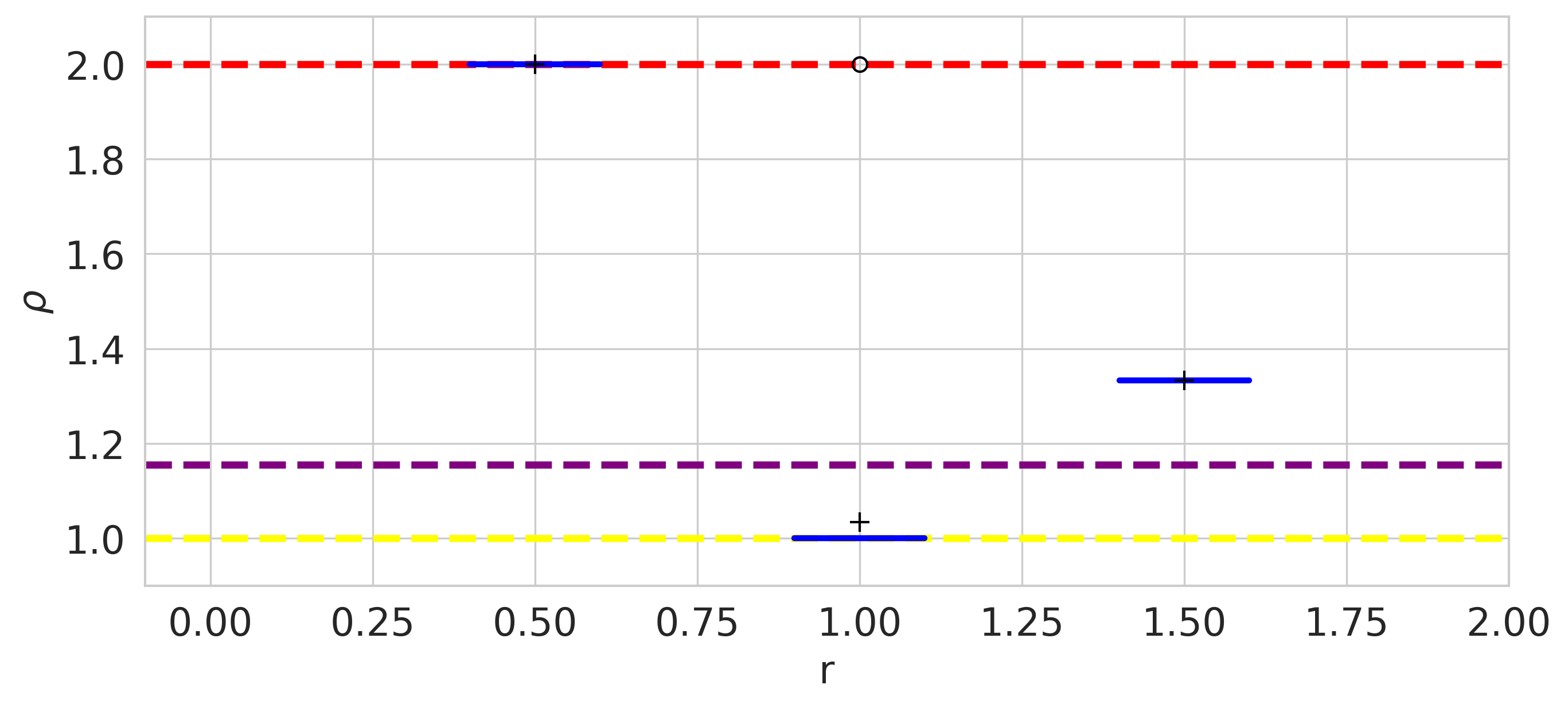} 
      \\ 
 \bottomrule
     \end{tabular}
 \end{table}
 \begin{table}[H]
 \caption{Curvature profile for model networks for various average degrees $ad$ and sizes.}
  \label{tab:modelnetpar}
\centering
\begin{tabular}{ c | c c c}
    \toprule
   \multicolumn{2}{c}{$ad=4$}   & $ad=6$ &  $ad=8$ \\	
       \midrule
       ER-100   & \includegraphics[width=0.25\linewidth]{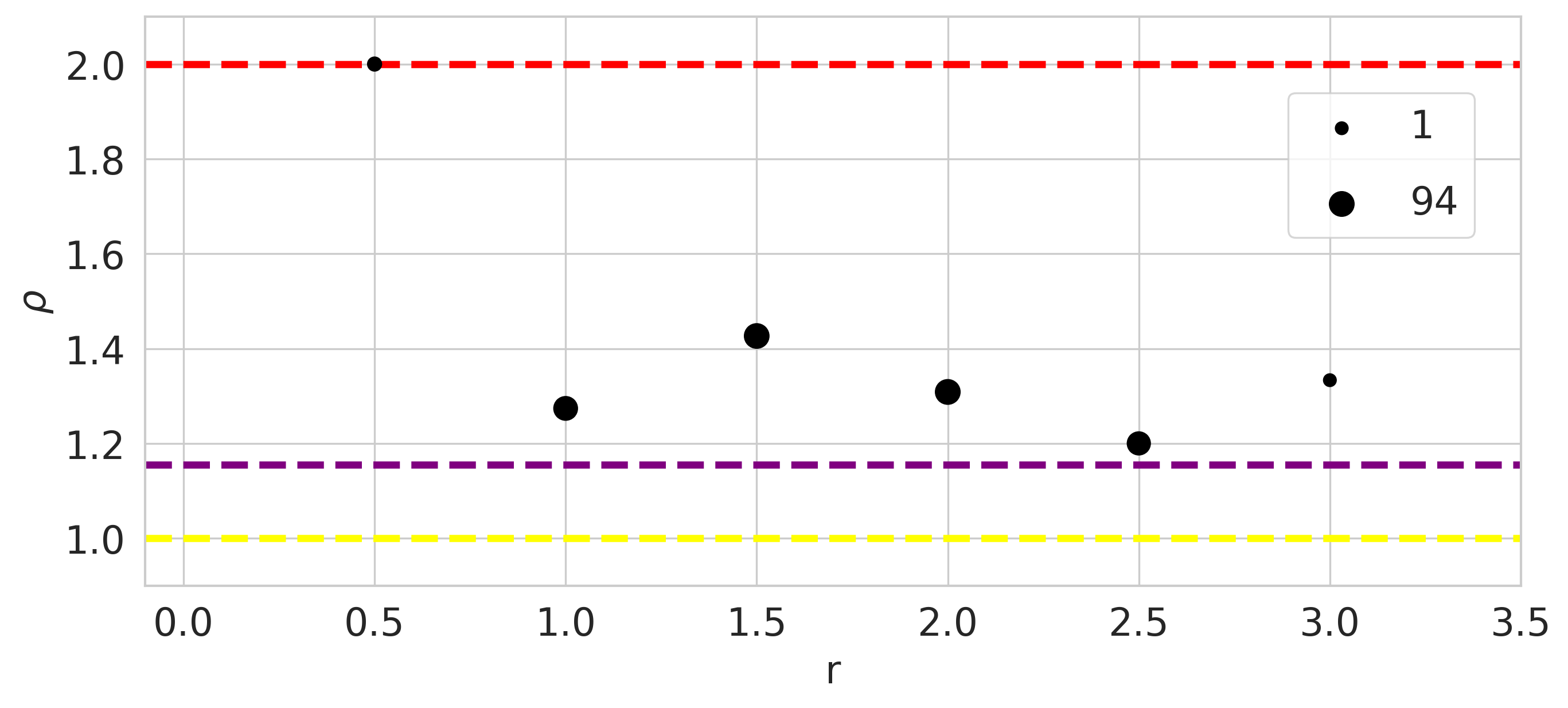} &\includegraphics[width=0.25\linewidth]{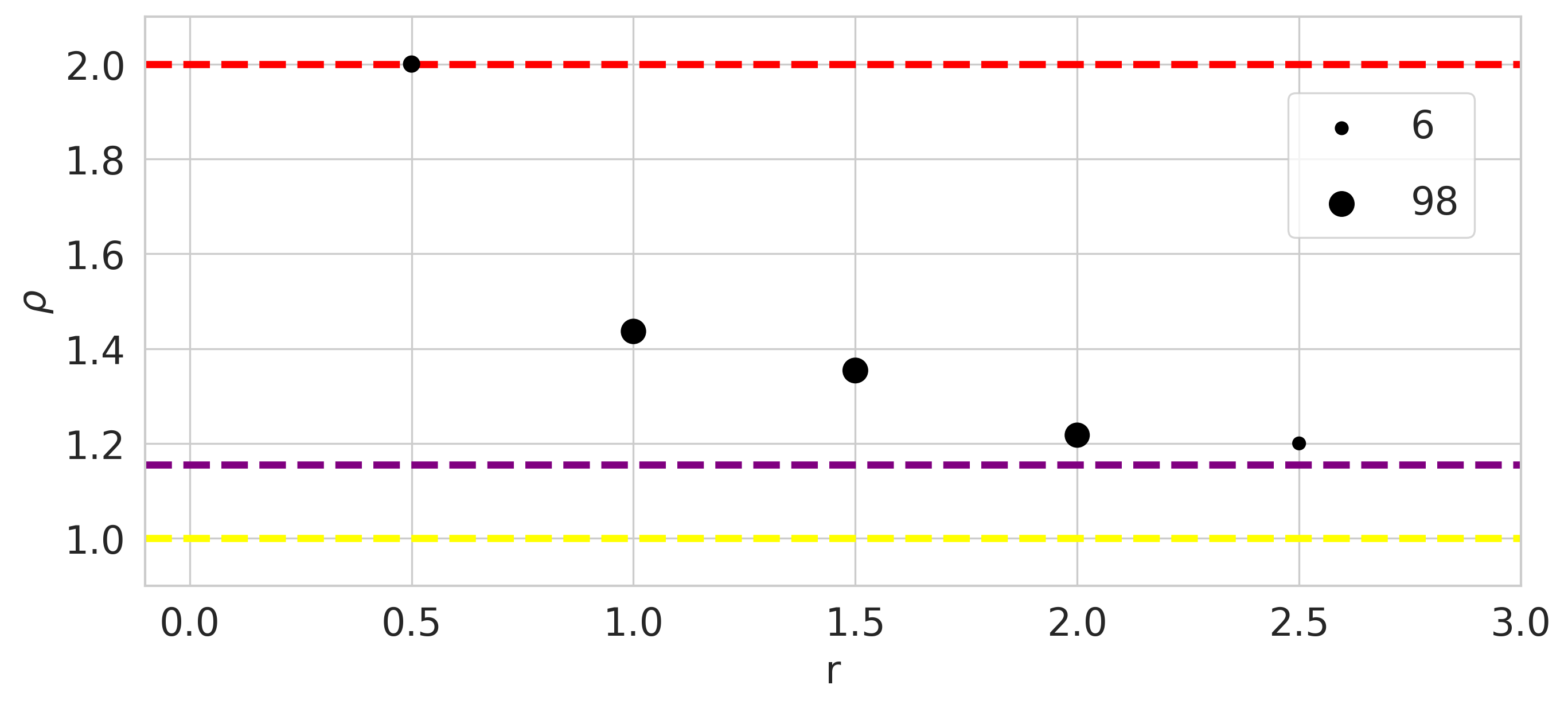}
         &\includegraphics[width=0.25\linewidth]{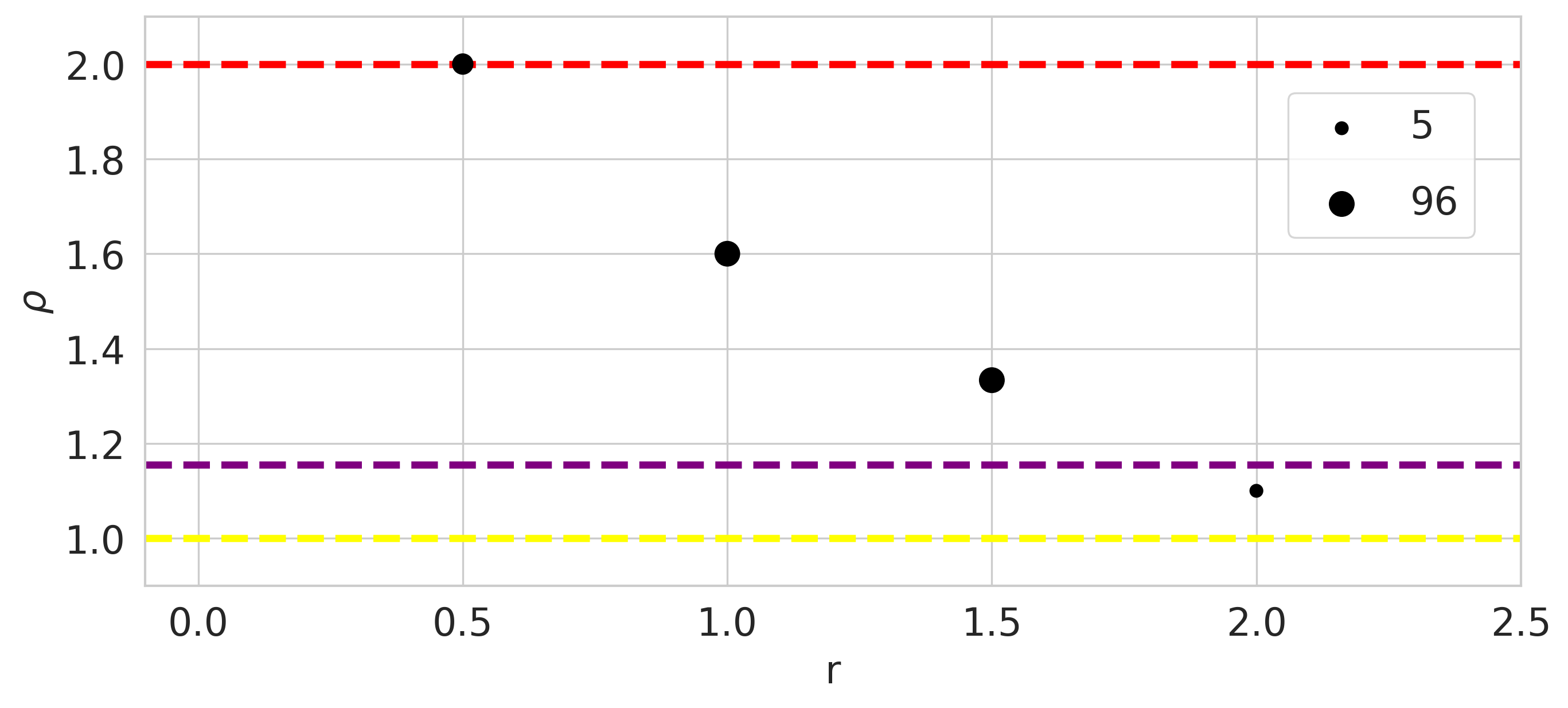}\\
       \midrule
      ER-1000    & \includegraphics[width=0.25\linewidth]{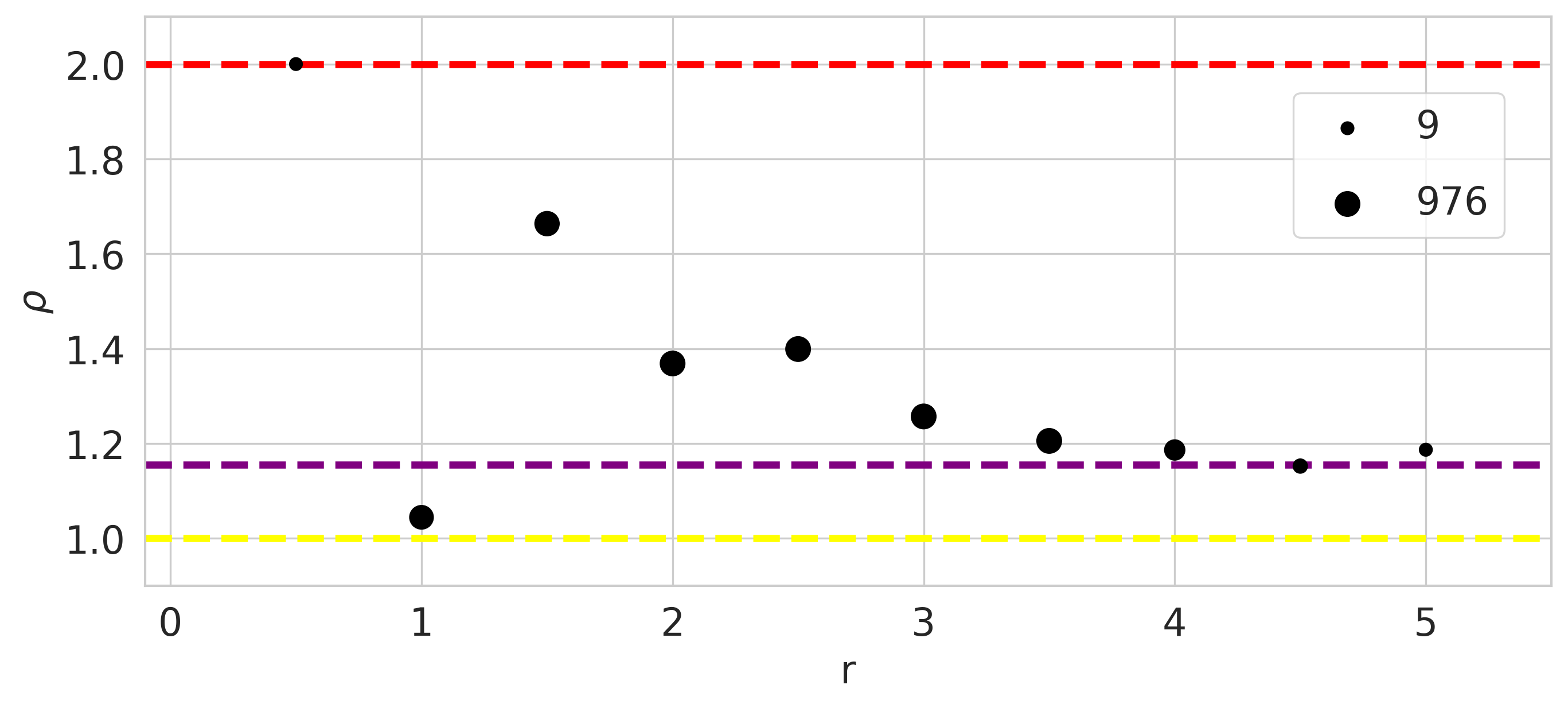} 
       &\includegraphics[width=0.25\linewidth]{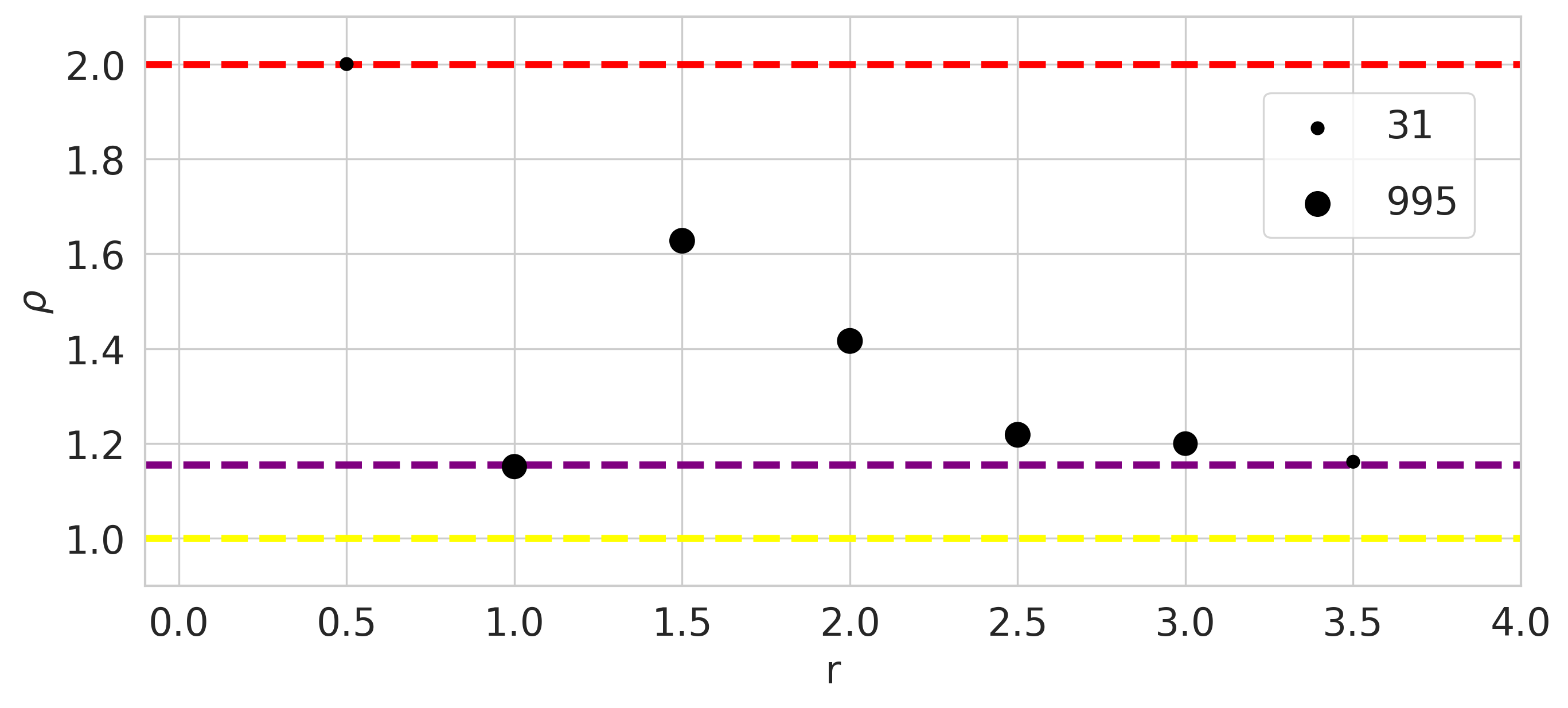}
       &  \includegraphics[width=0.25\linewidth]{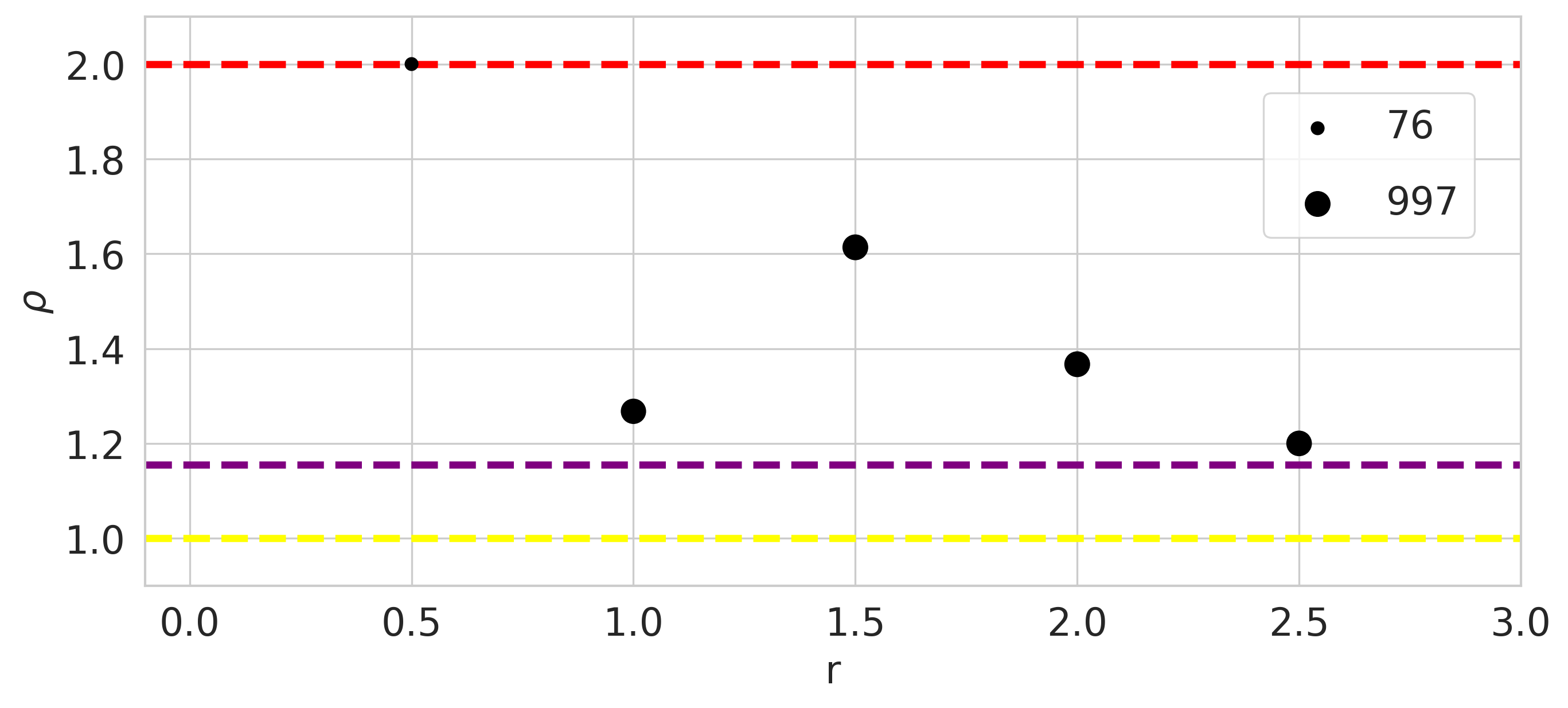}\\
       \midrule
       ER-10000    & \includegraphics[width=0.25\linewidth]{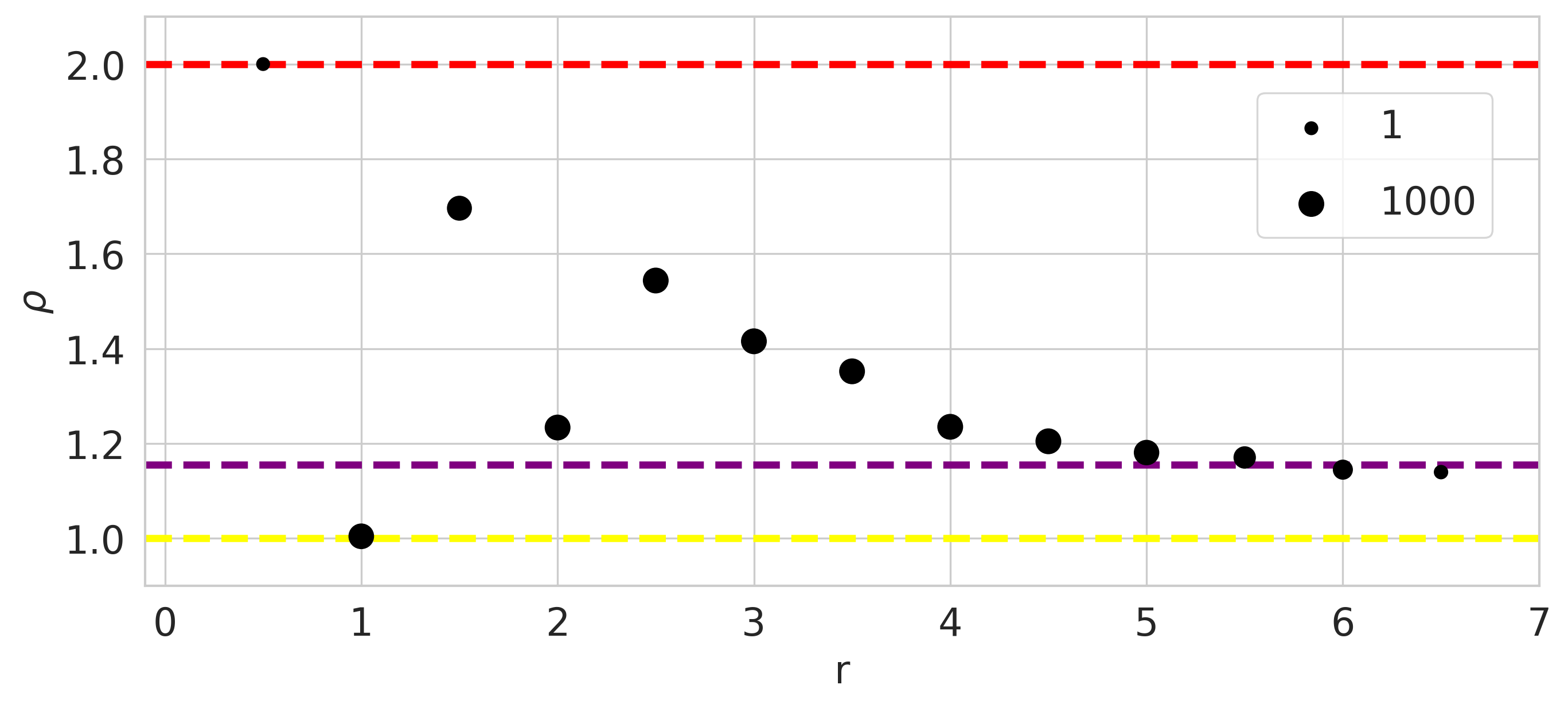} 
       &\includegraphics[width=0.25\linewidth]{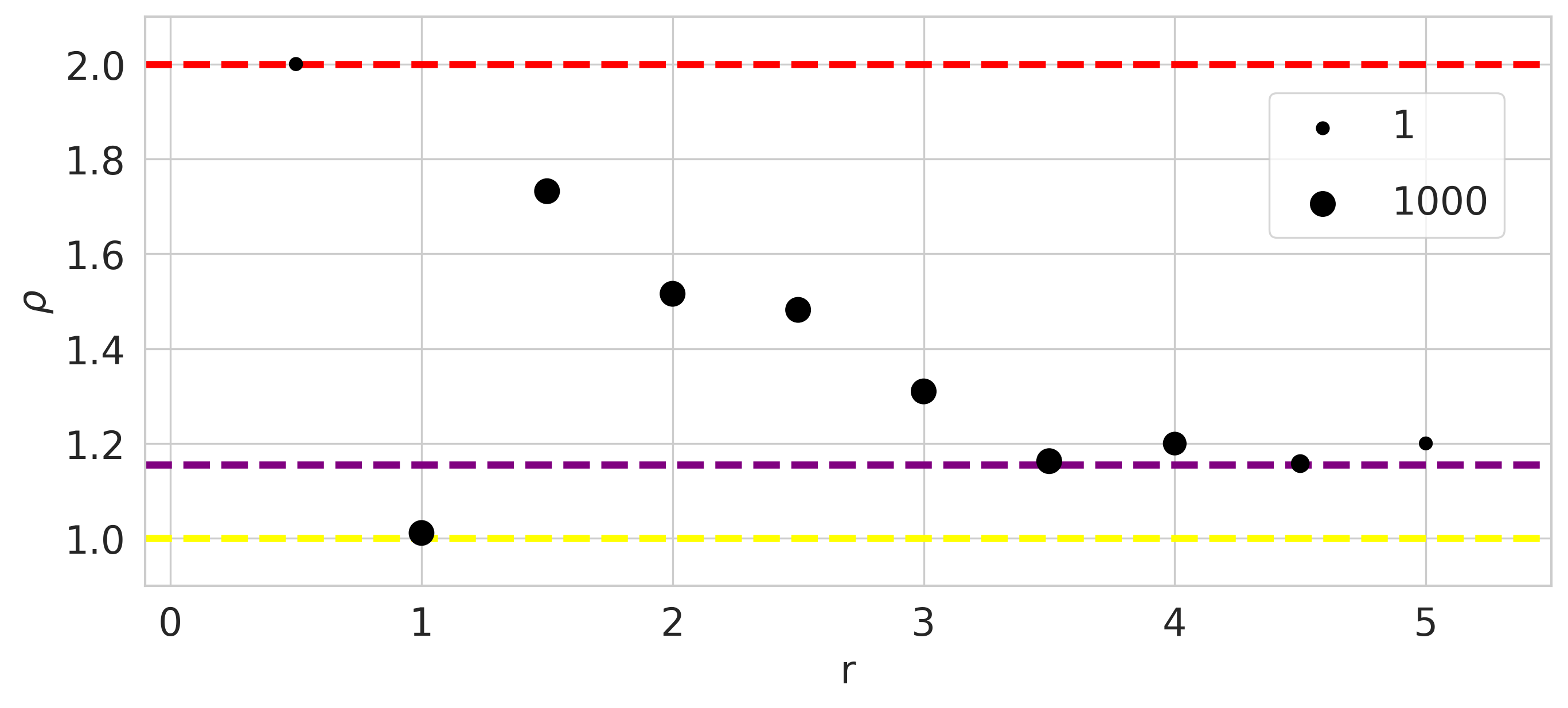}
       &  \includegraphics[width=0.25\linewidth]{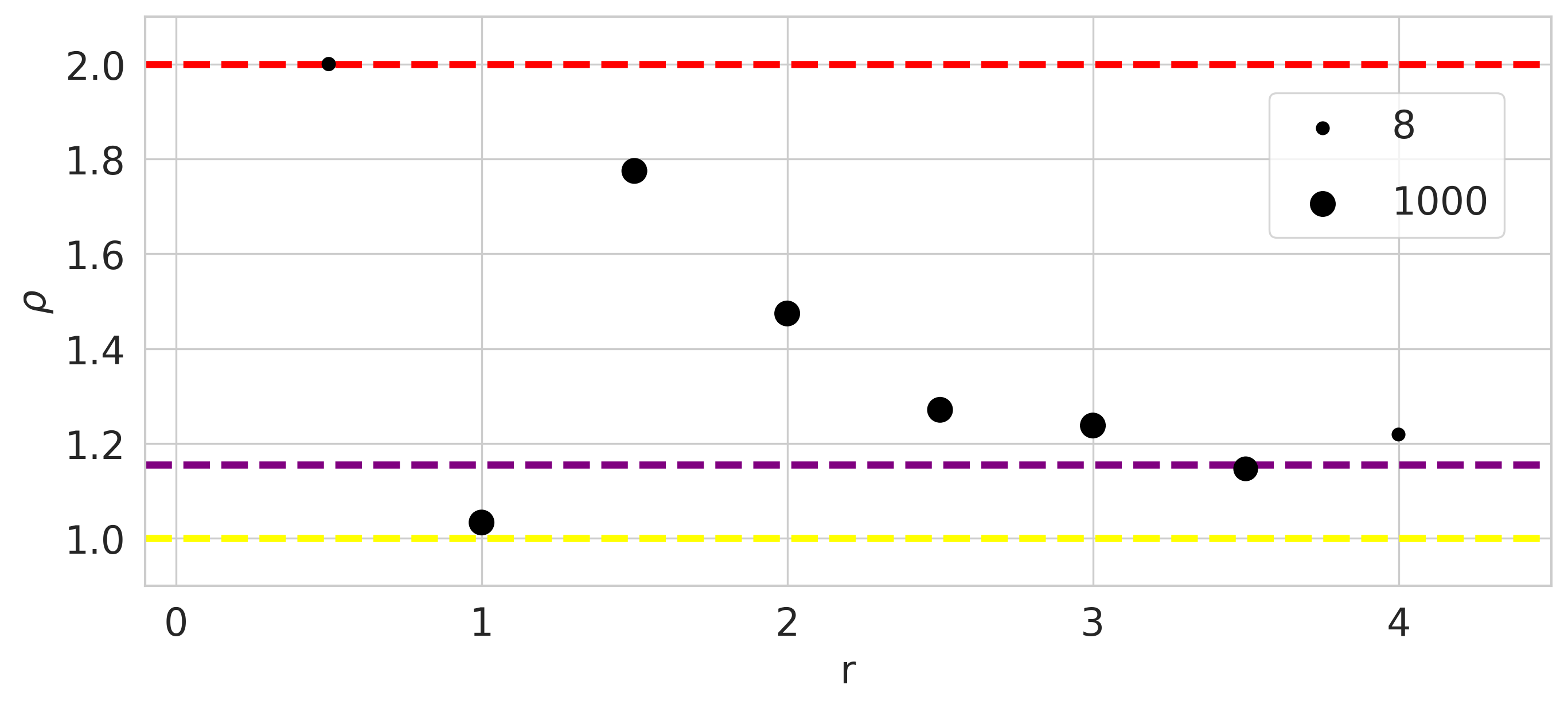}\\
       \midrule
        WS-100    & \includegraphics[width=0.25\linewidth]{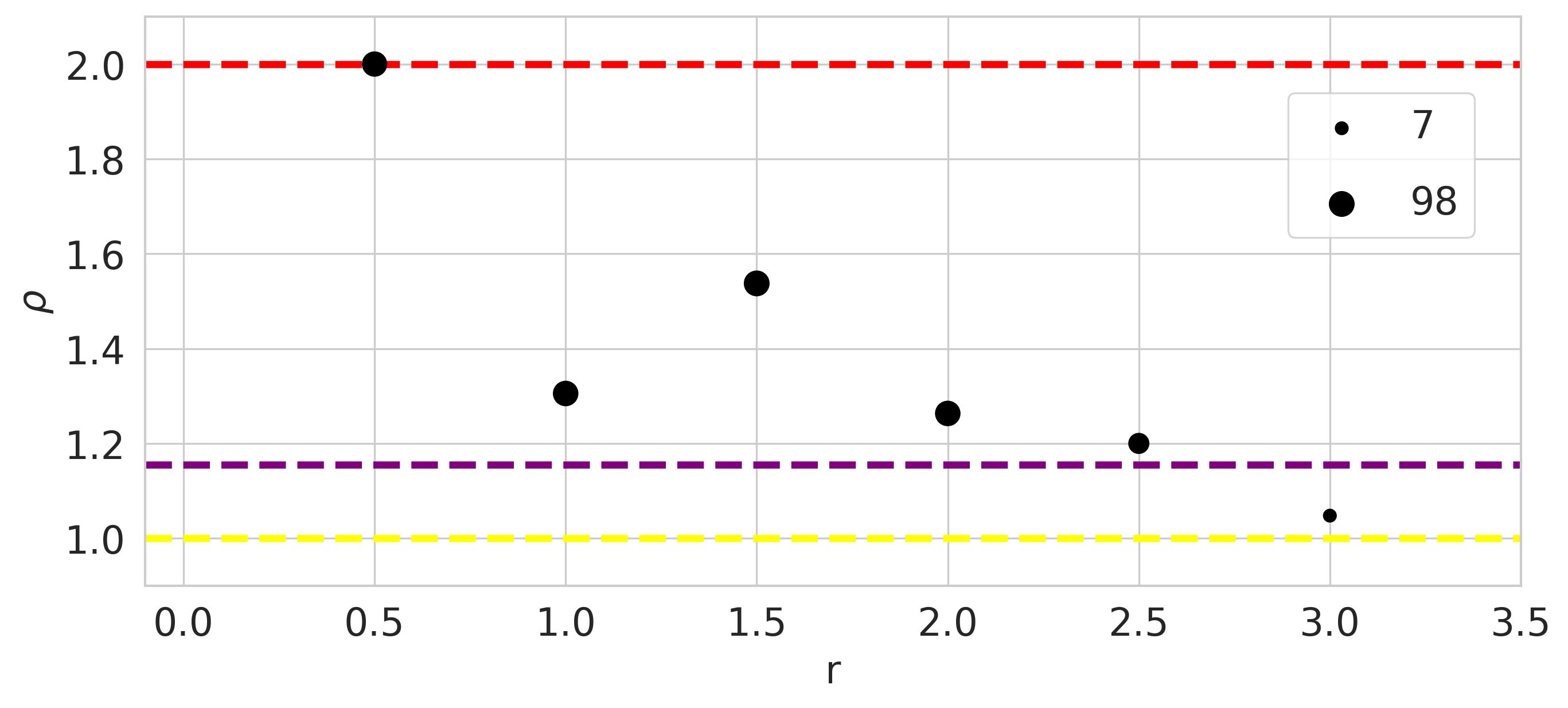}
    &  \includegraphics[width=0.25\linewidth]{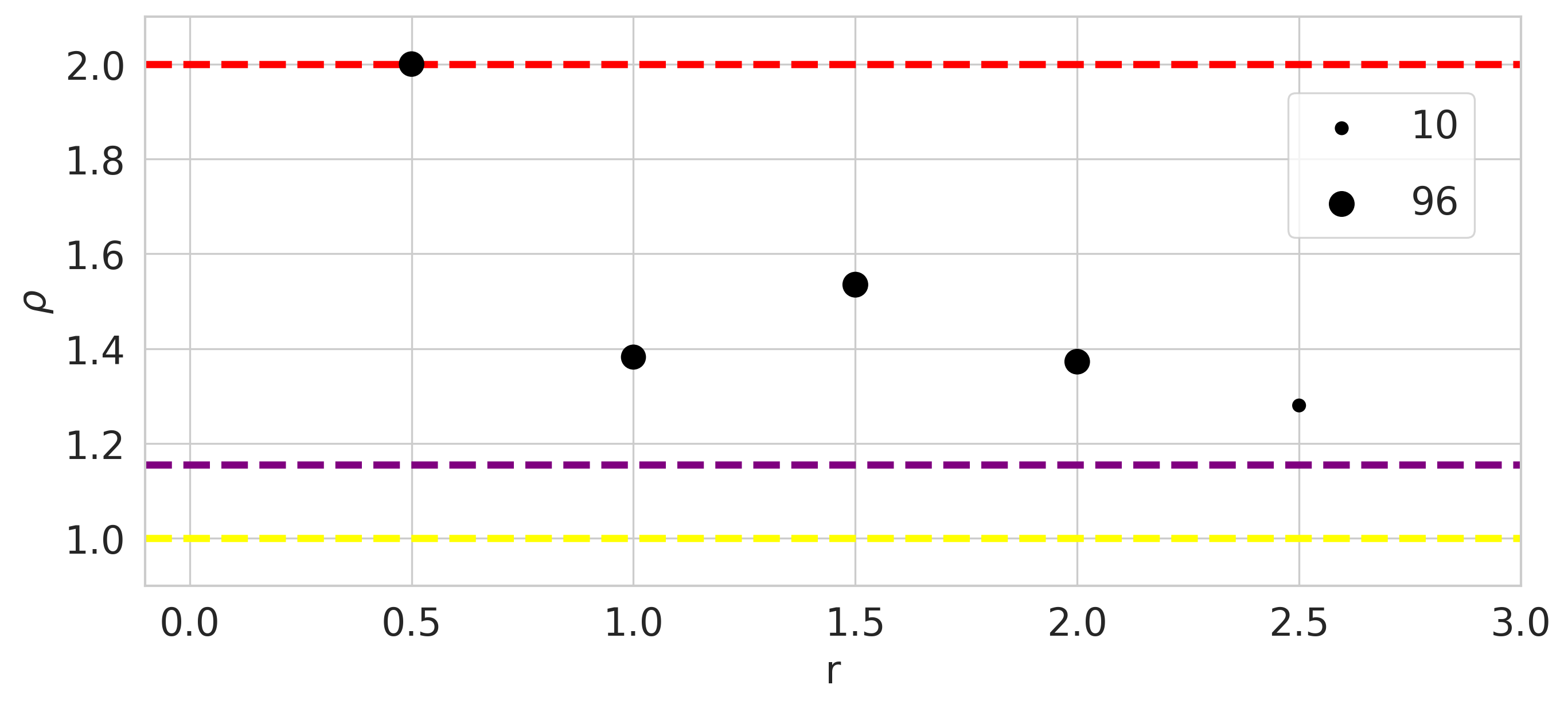}
      &  \includegraphics[width=0.25\linewidth]{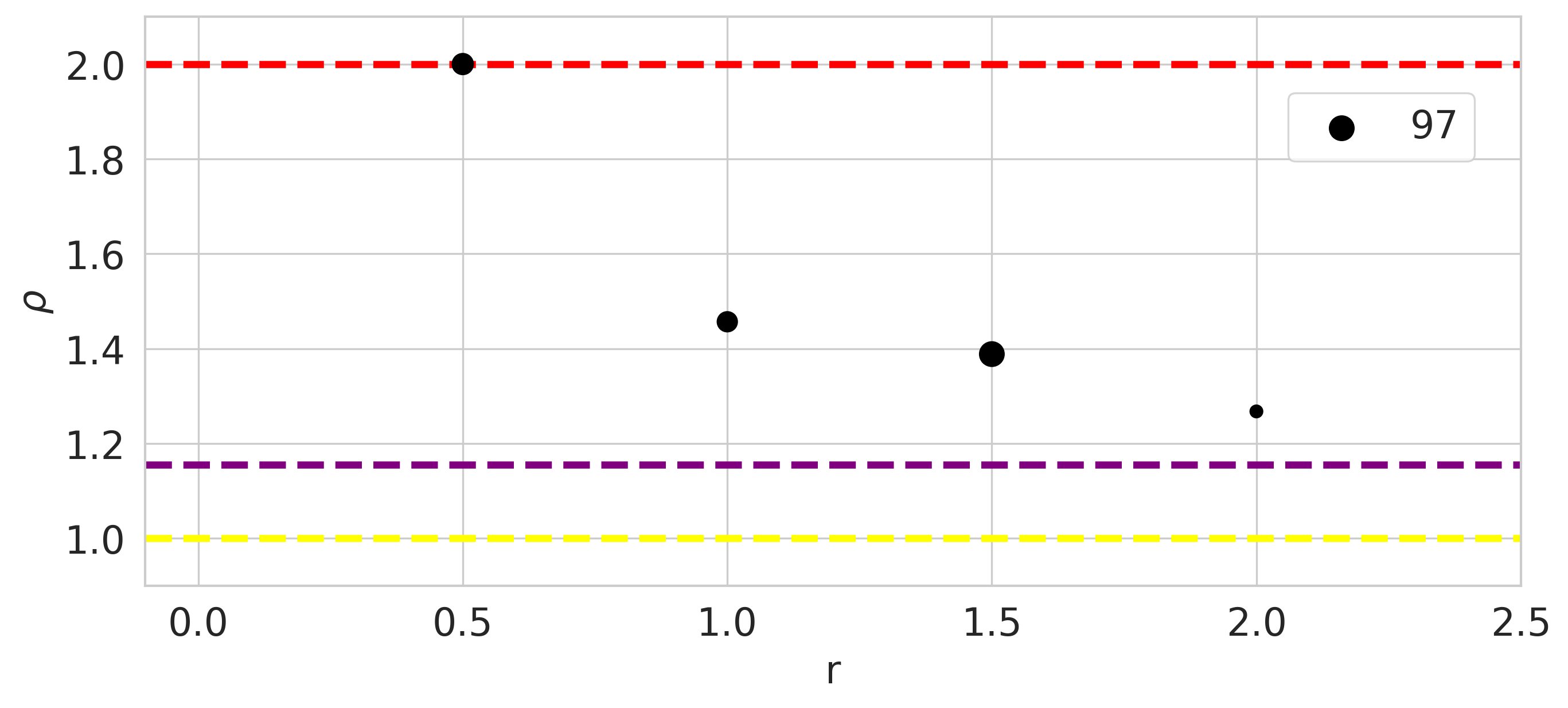} \\ 
       \midrule
  WS-1000    & \includegraphics[width=0.25\linewidth]{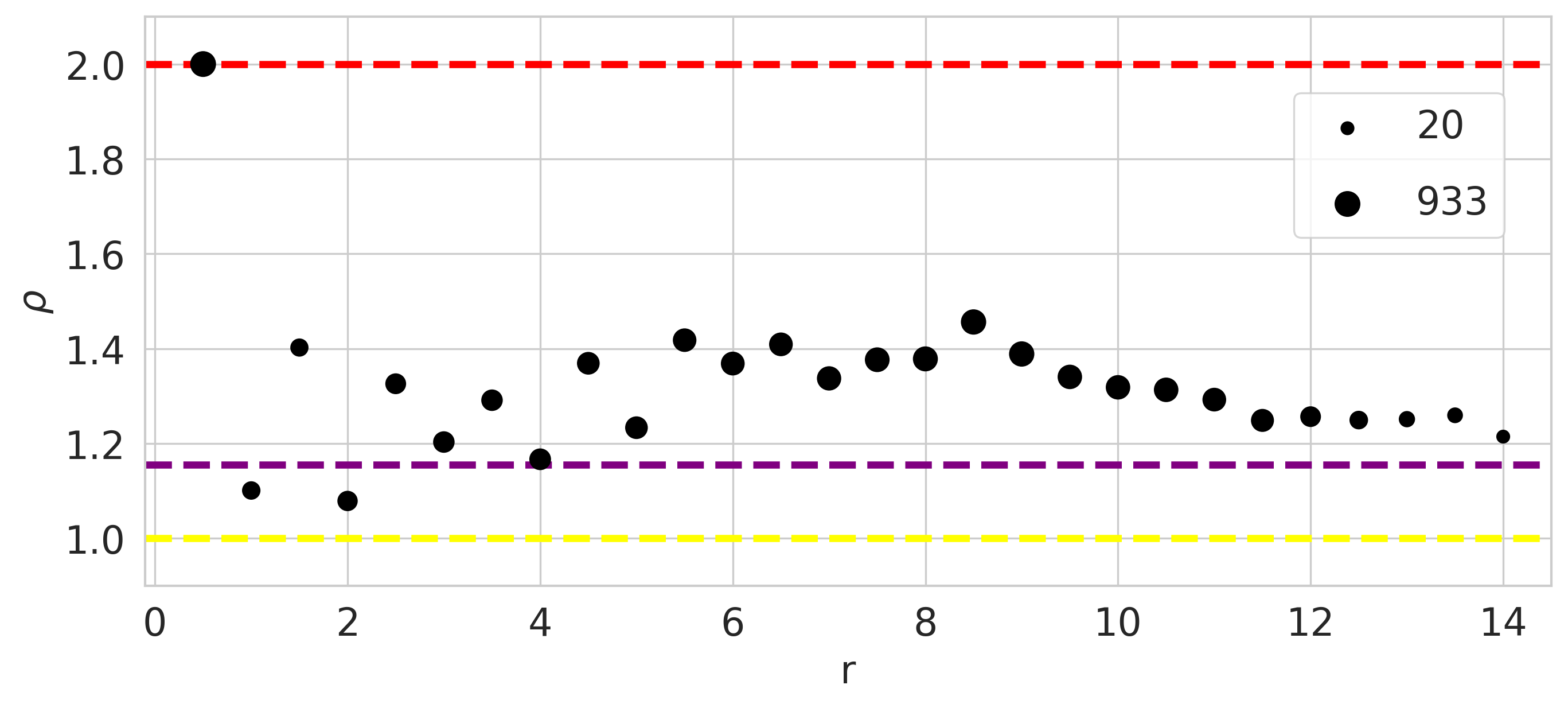}
    &  \includegraphics[width=0.25\linewidth]{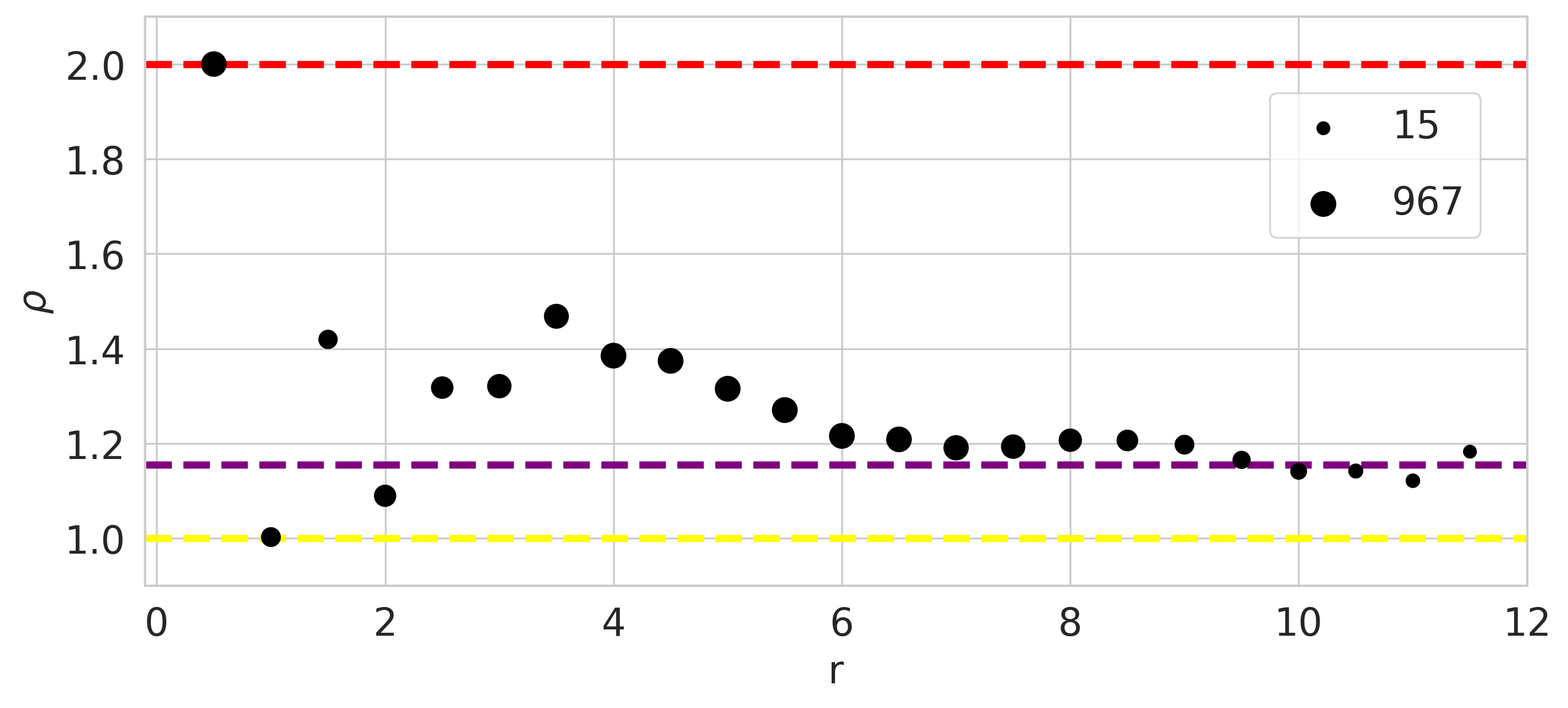}
      &  \includegraphics[width=0.25\linewidth]{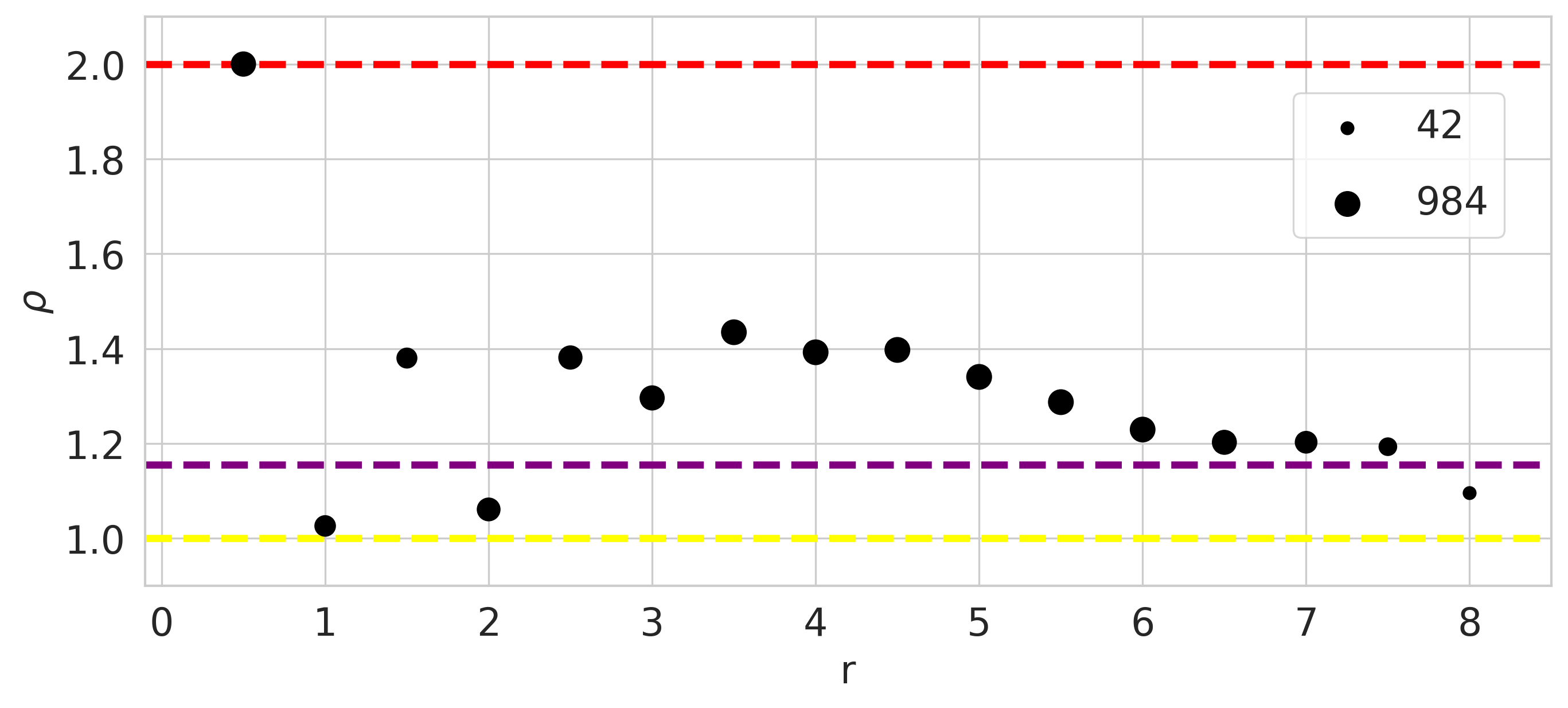} \\ 
       \midrule
  WS-10000    & \includegraphics[width=0.25\linewidth]{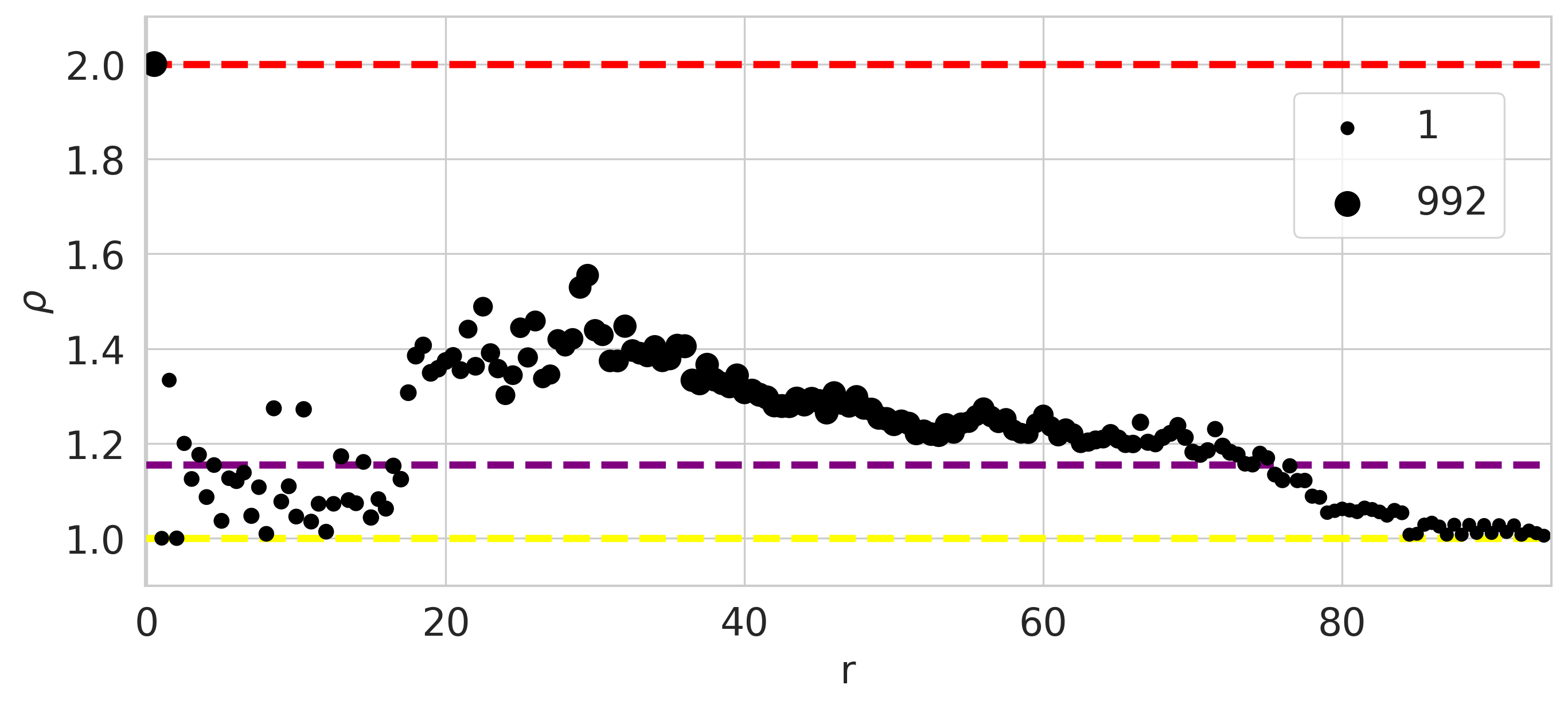}
    &  \includegraphics[width=0.25\linewidth]{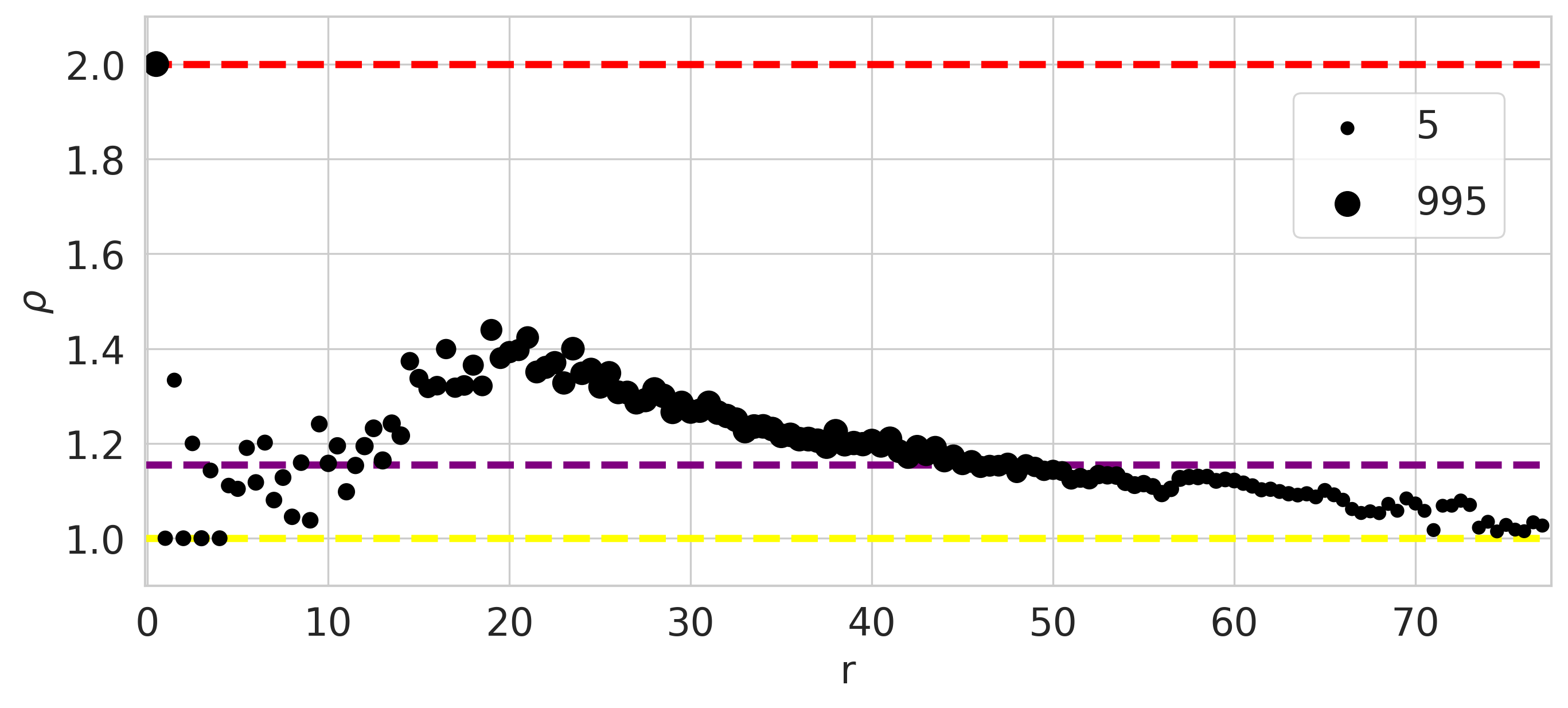}
      &  \includegraphics[width=0.25\linewidth]{images/networks/WS_network/WS_10000_6_c_1_m_0.1_avg.png} \\
     \bottomrule
     \end{tabular}
 \end{table}

\subsection{Curvature profile for metric spaces}
 The next phase computes $\rho$ on some datasets that are initially presented as finite metric spaces. We start with toy examples consisting of samples drawn from a $2-$dimensional Euclidean plane, circle, and a graph tree, each representing a canonical model of flat, highly positive, or highly negative curvature. The curvature profiles for these toy examples are shown in \cref{tab:toyex}. The close agreement between the box plots and the averaged and non-averaged curvature profiles illustrates the structural rigidity inherent in these geometries. The rigid clustering of curvature values in the extreme cases, i.e. tree (most negatively curved) and circle (most positively curved), emphasizes the distinctive nature of these spaces. While the Euclidean plane's distribution deviates slightly from the expected flat geometry benchmark at $\rho=\frac{3}{\sqrt{2}}$, likely due to a sampling artifact, the averaged $\rho$ profile offers a clearer alignment with flat geometry, affirming that our method effectively captures key geometric characteristics.\\
 \begin{table}[H]
 \caption{Curvature profile for reference spaces.}
  \label{tab:toyex}
\centering
\begin{tabular}{ c | c c c}
    \toprule
      \multicolumn{2}{c}{averaged}   & non-averaged &  boxplot \\		
       \midrule
       \includegraphics[width=0.10\linewidth]{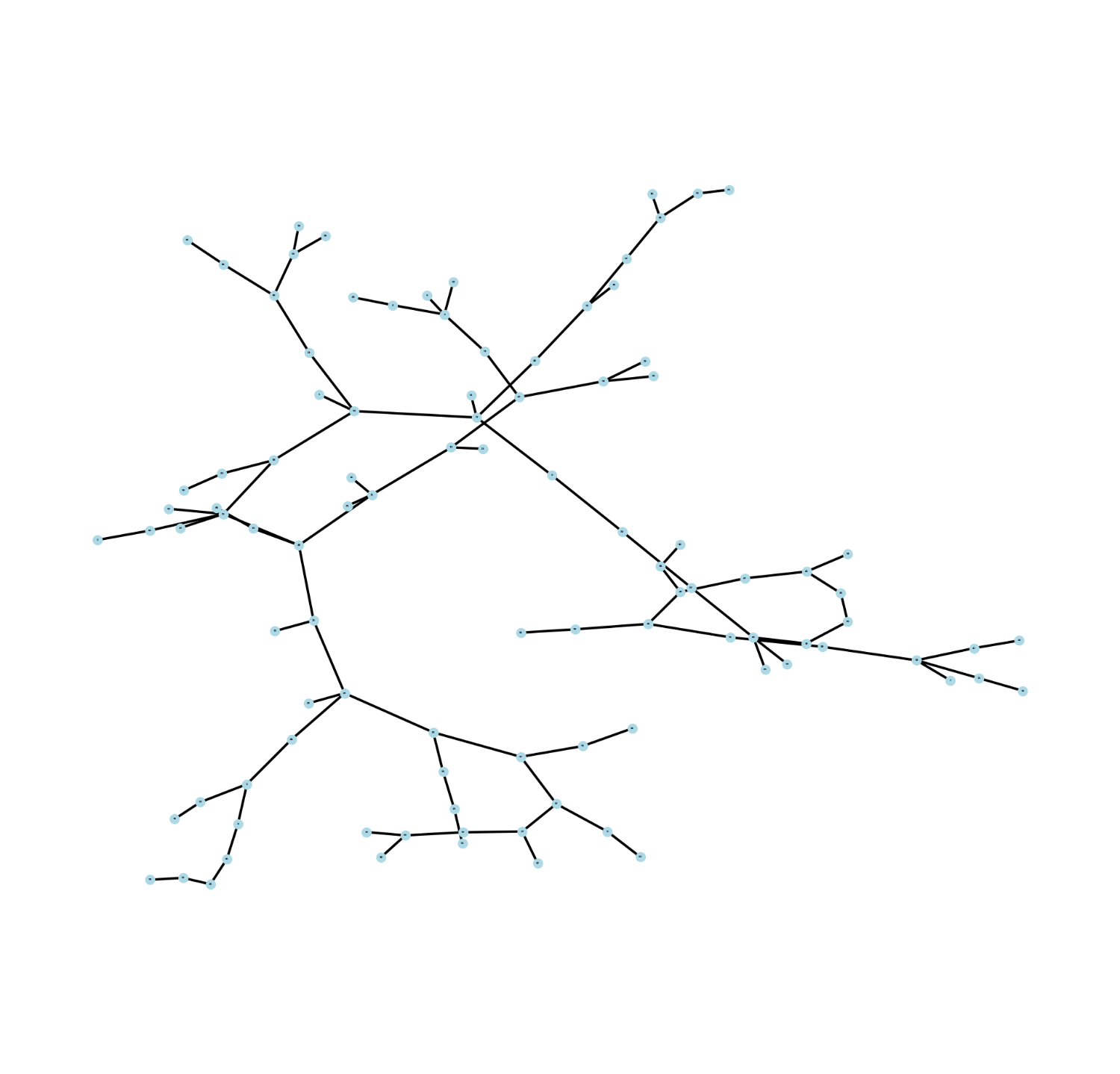}  & \includegraphics[width=0.25\linewidth]{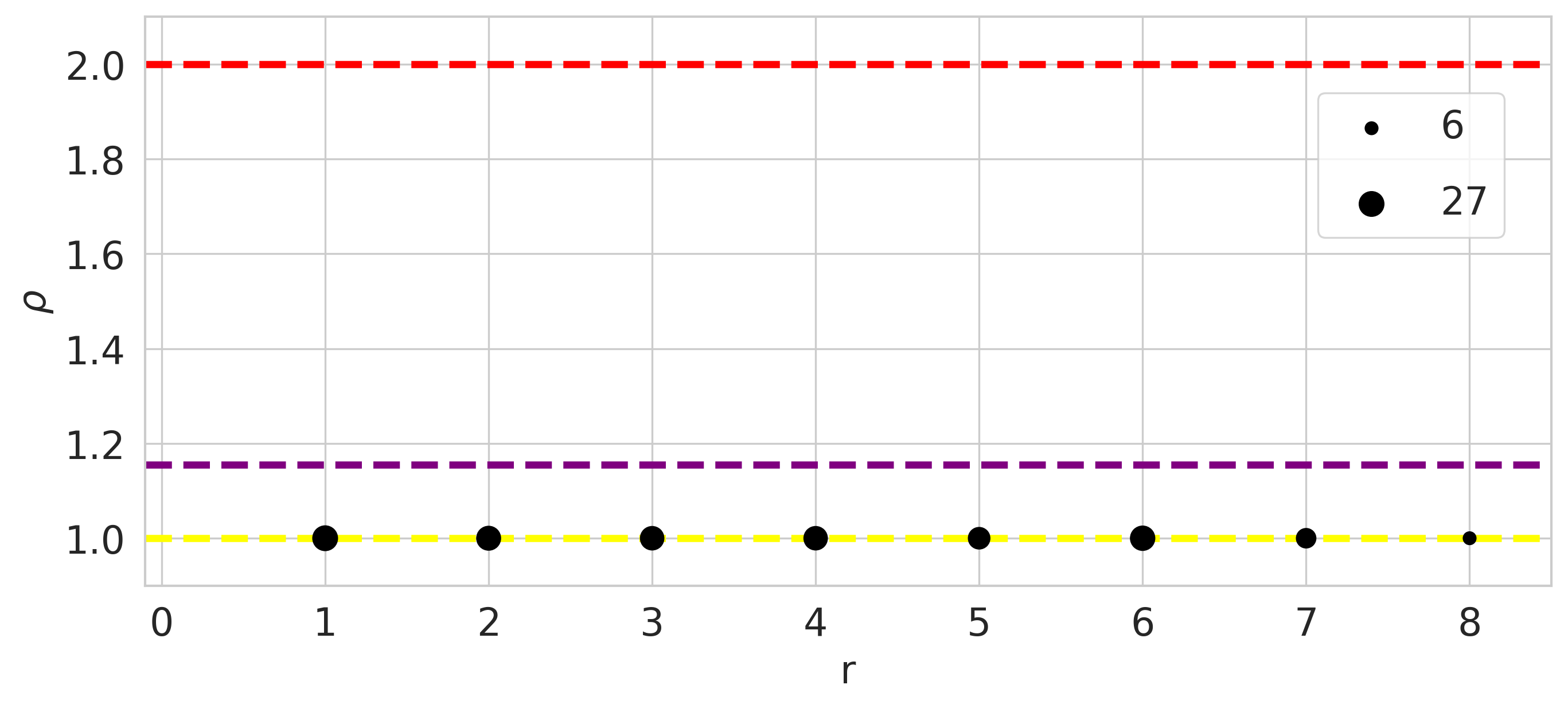}
      & \includegraphics[width=0.25\linewidth]{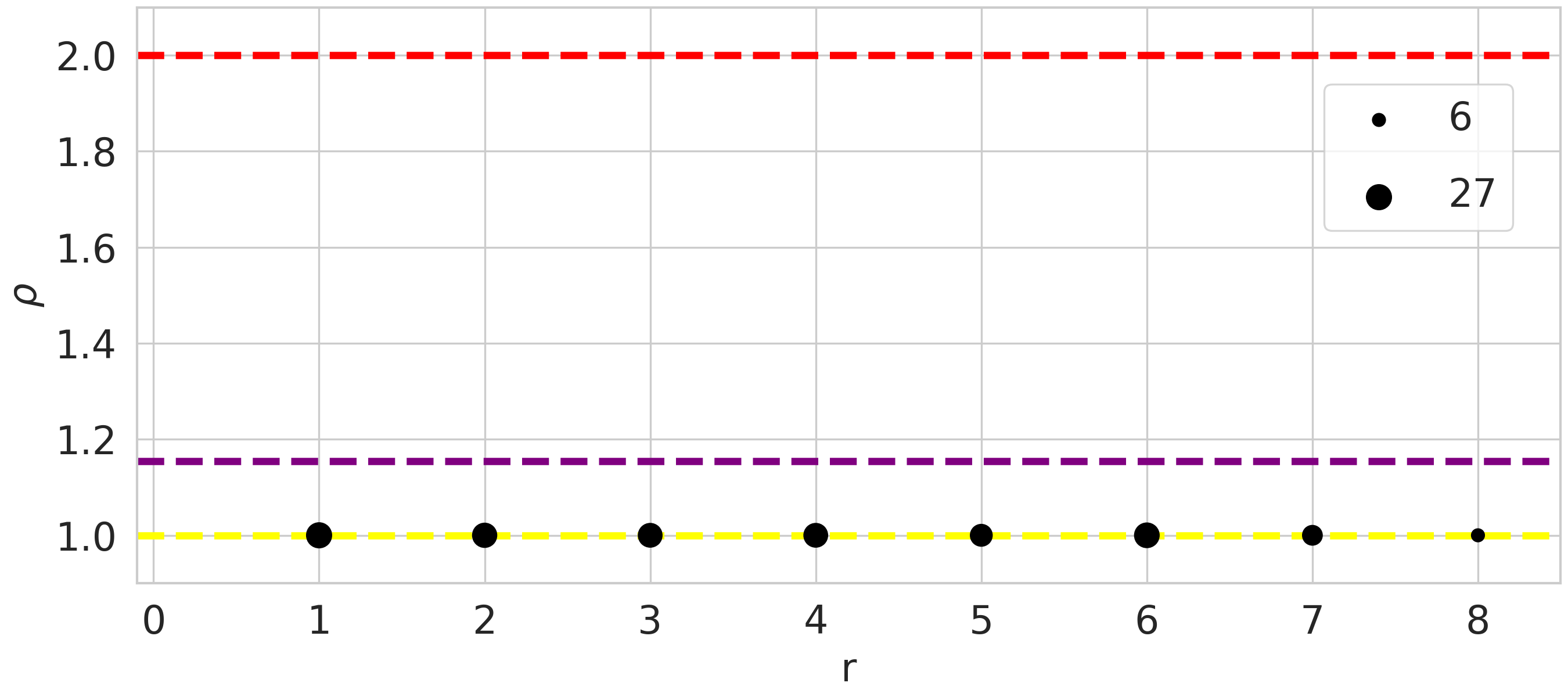}
      &  \includegraphics[width=0.25\linewidth]{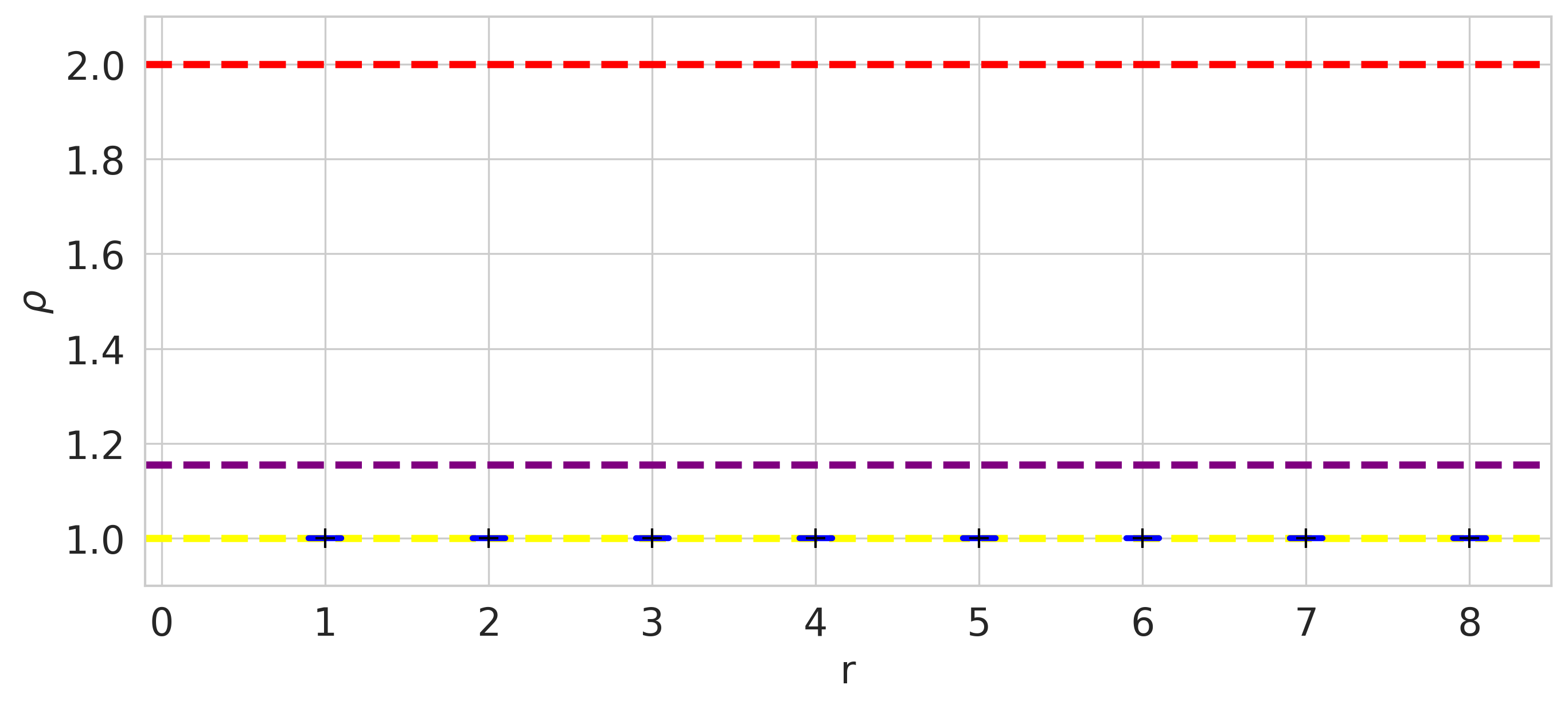} 
      \\ 
      \midrule 
       \includegraphics[width=0.10\linewidth]{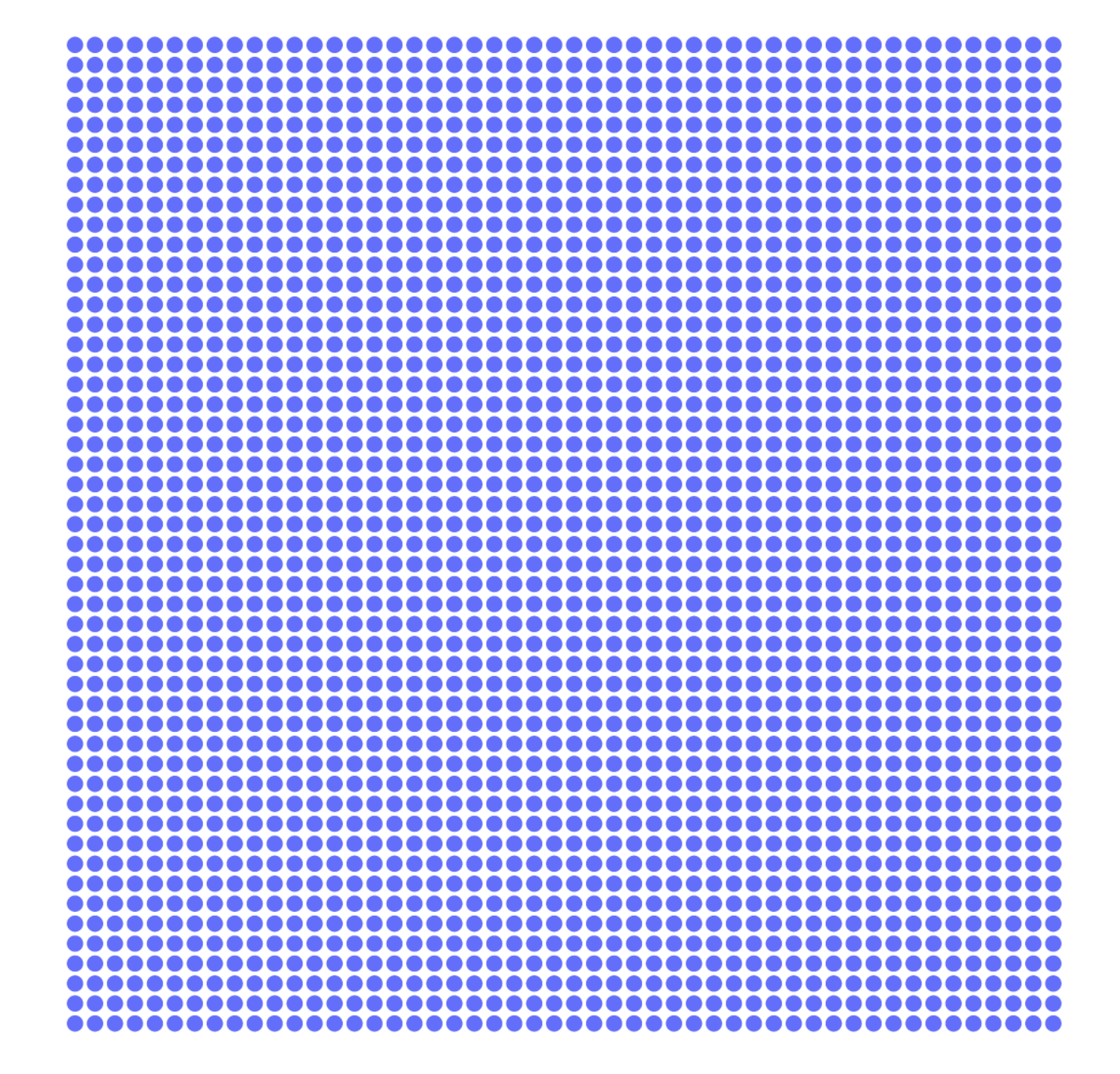}  & \includegraphics[width=0.25\linewidth]{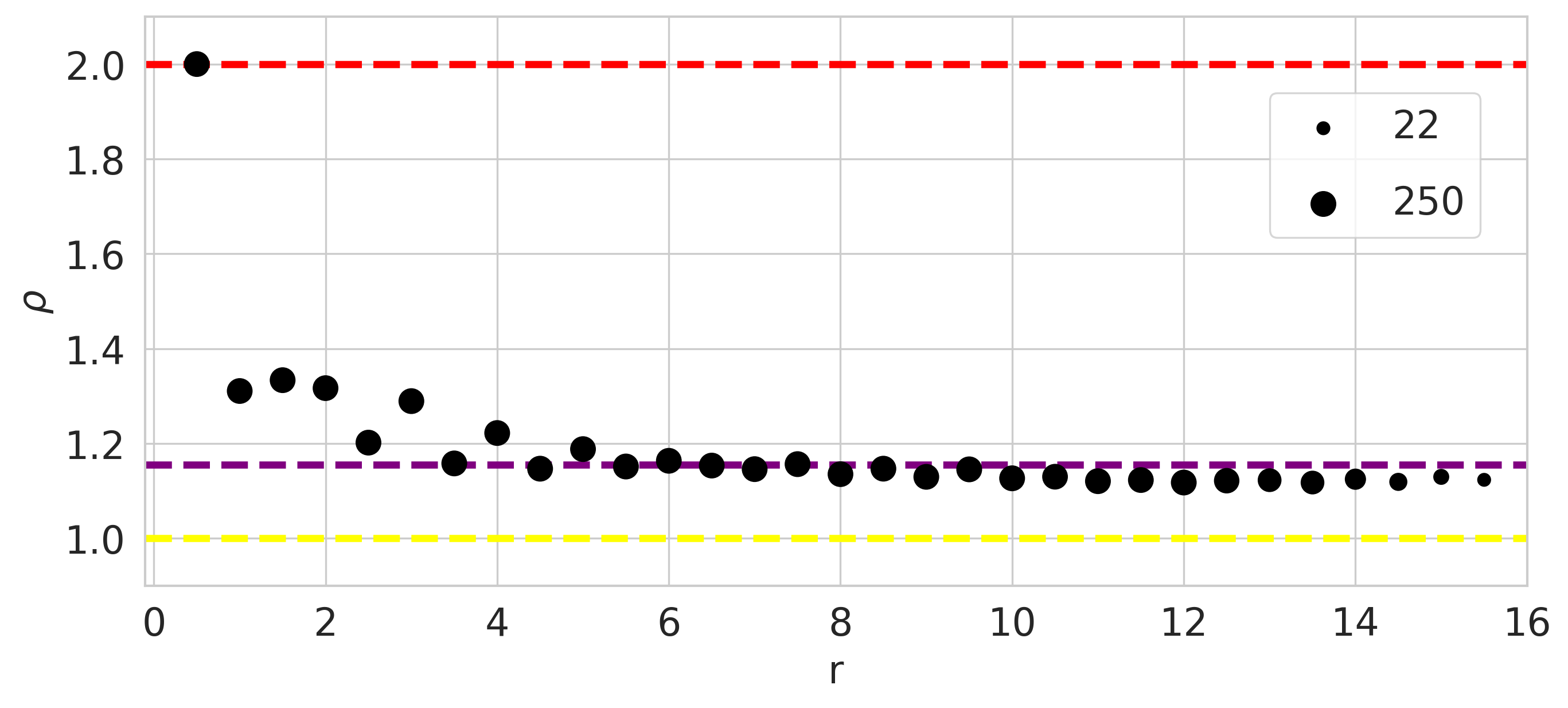}
      & \includegraphics[width=0.25\linewidth]{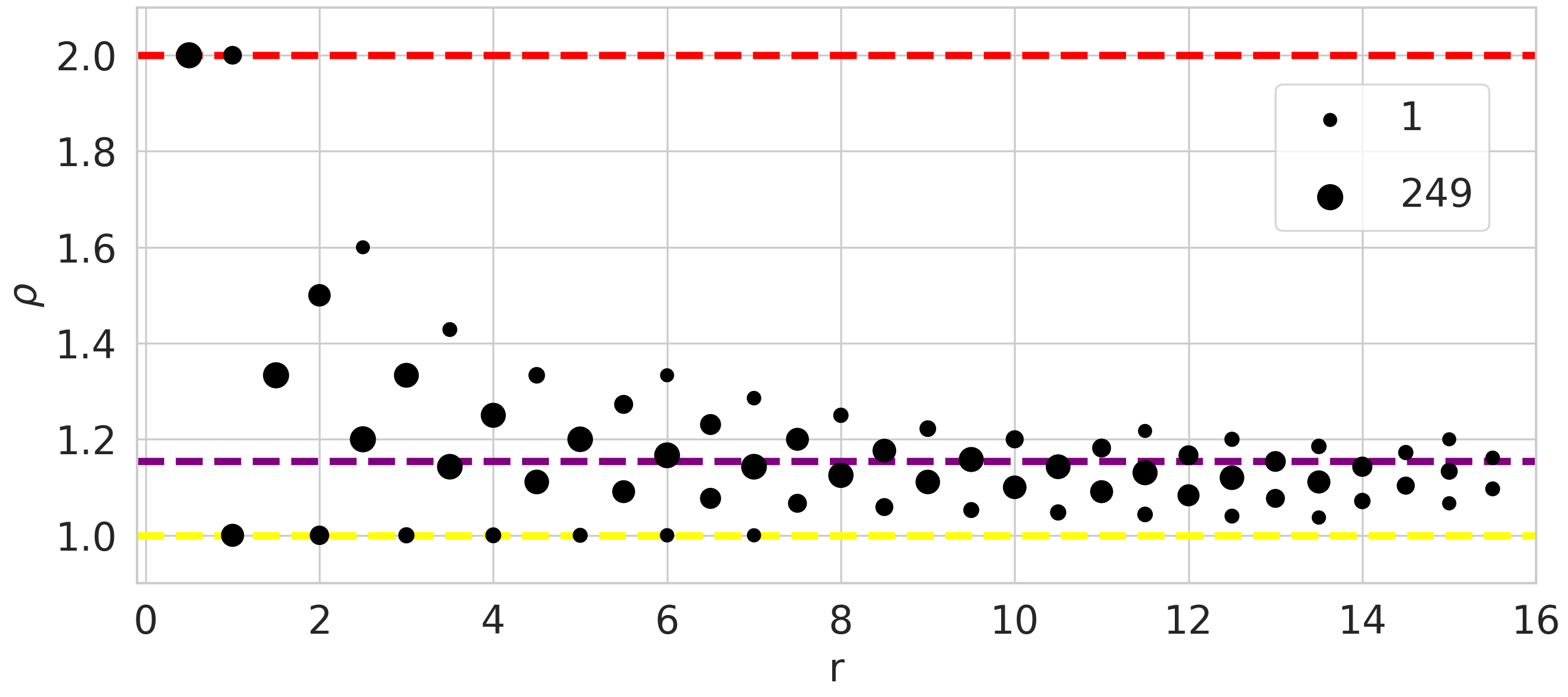}
      &  \includegraphics[width=0.25\linewidth]{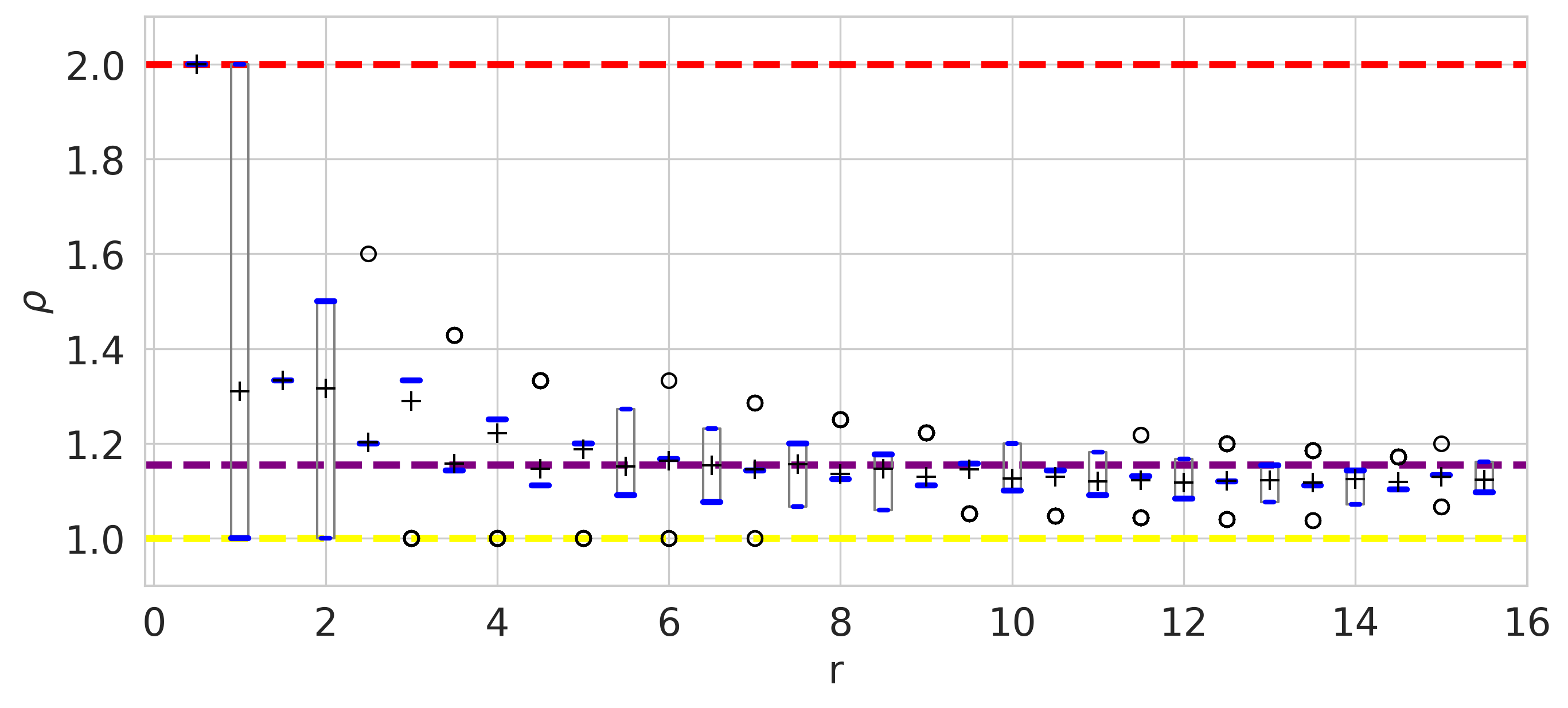} \\
      \midrule
      \includegraphics[width=0.10\linewidth]{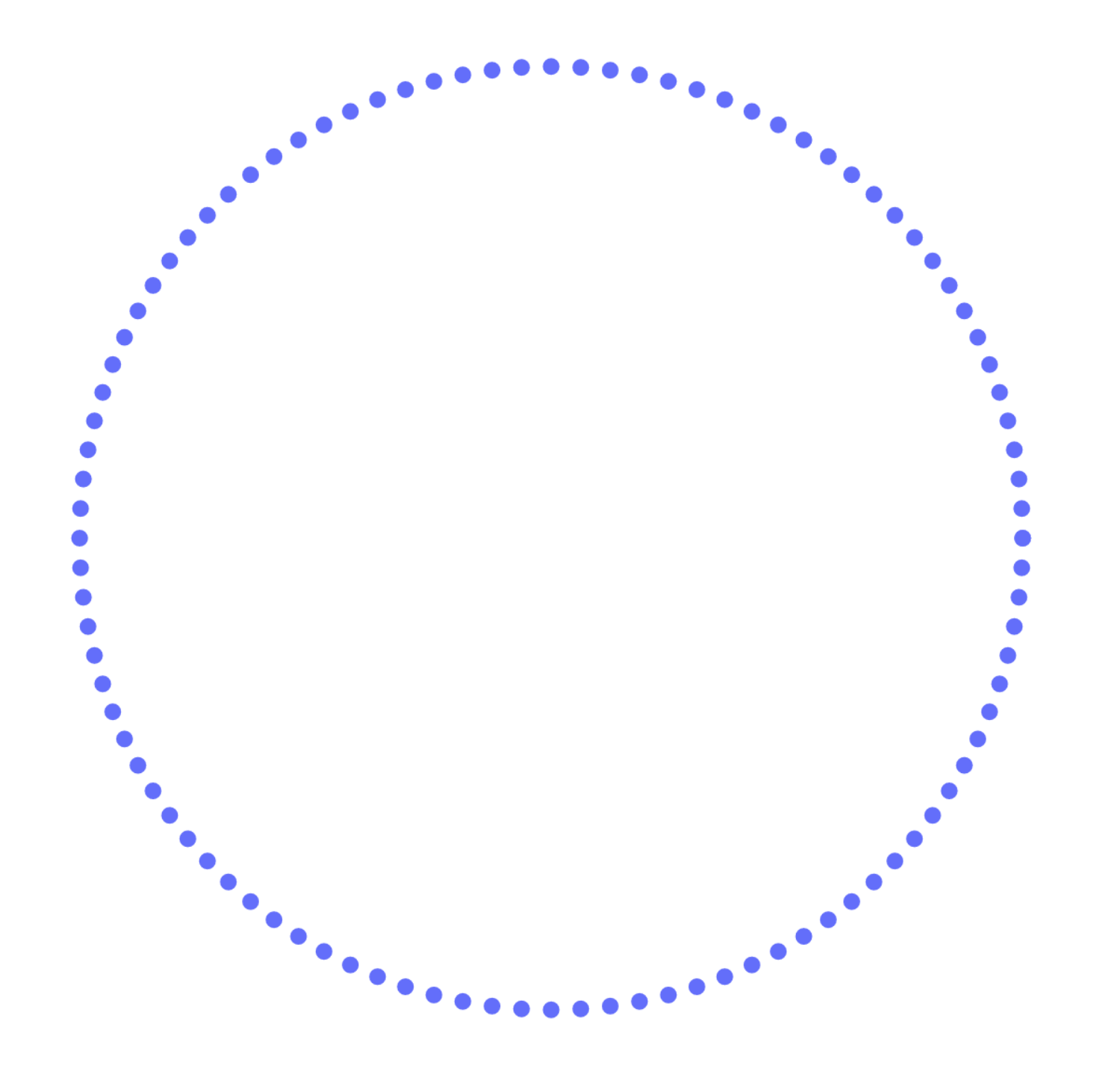}   & \includegraphics[width=0.25\linewidth]{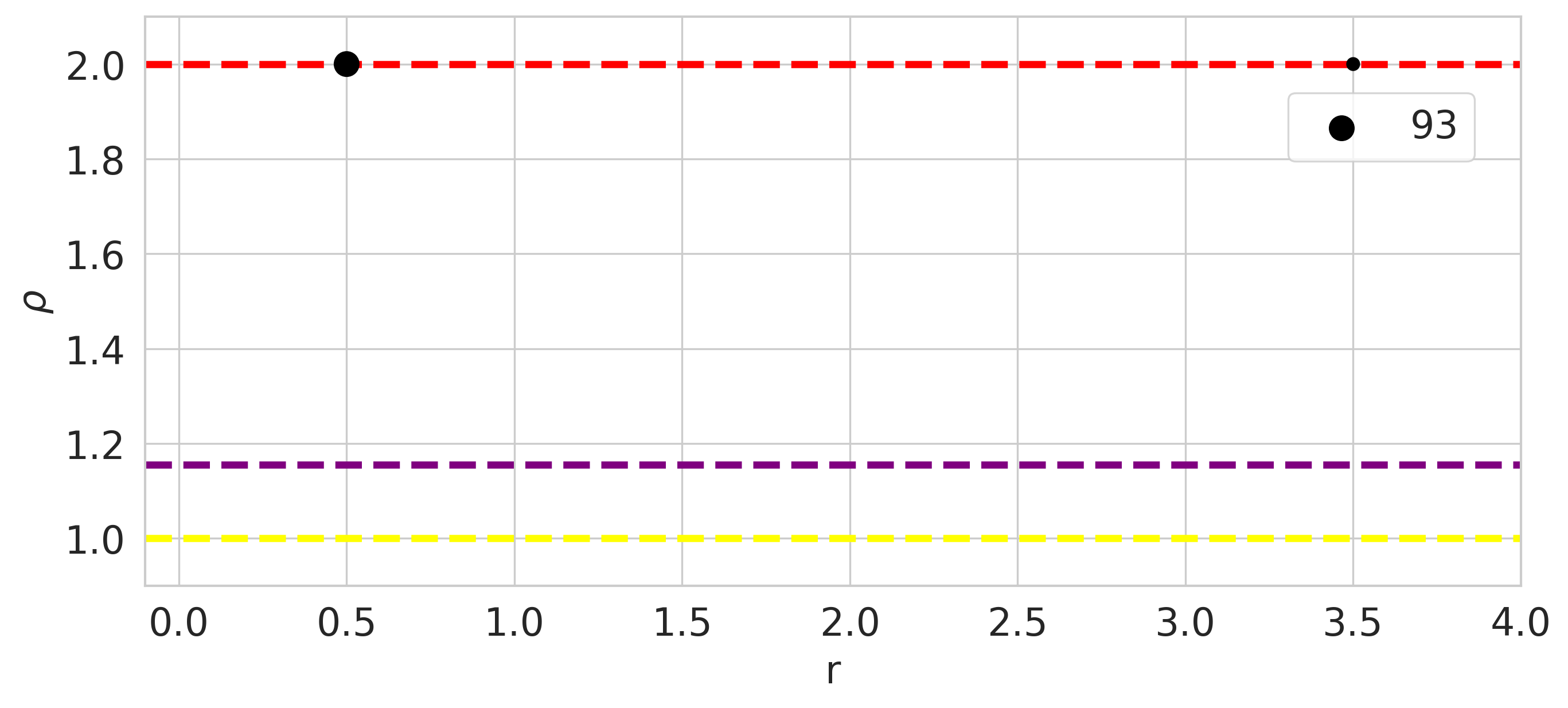}
      & \includegraphics[width=0.25\linewidth]{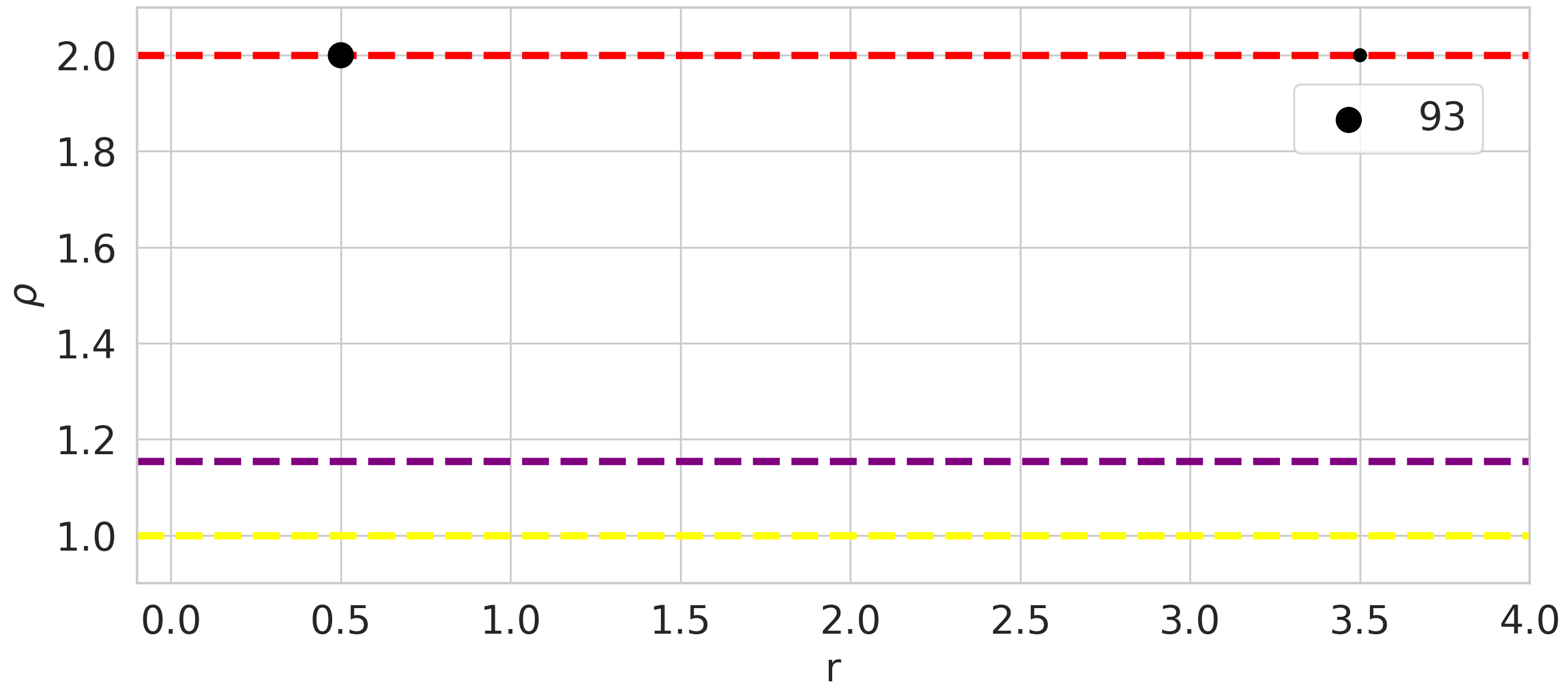}
      &  \includegraphics[width=0.25\linewidth]{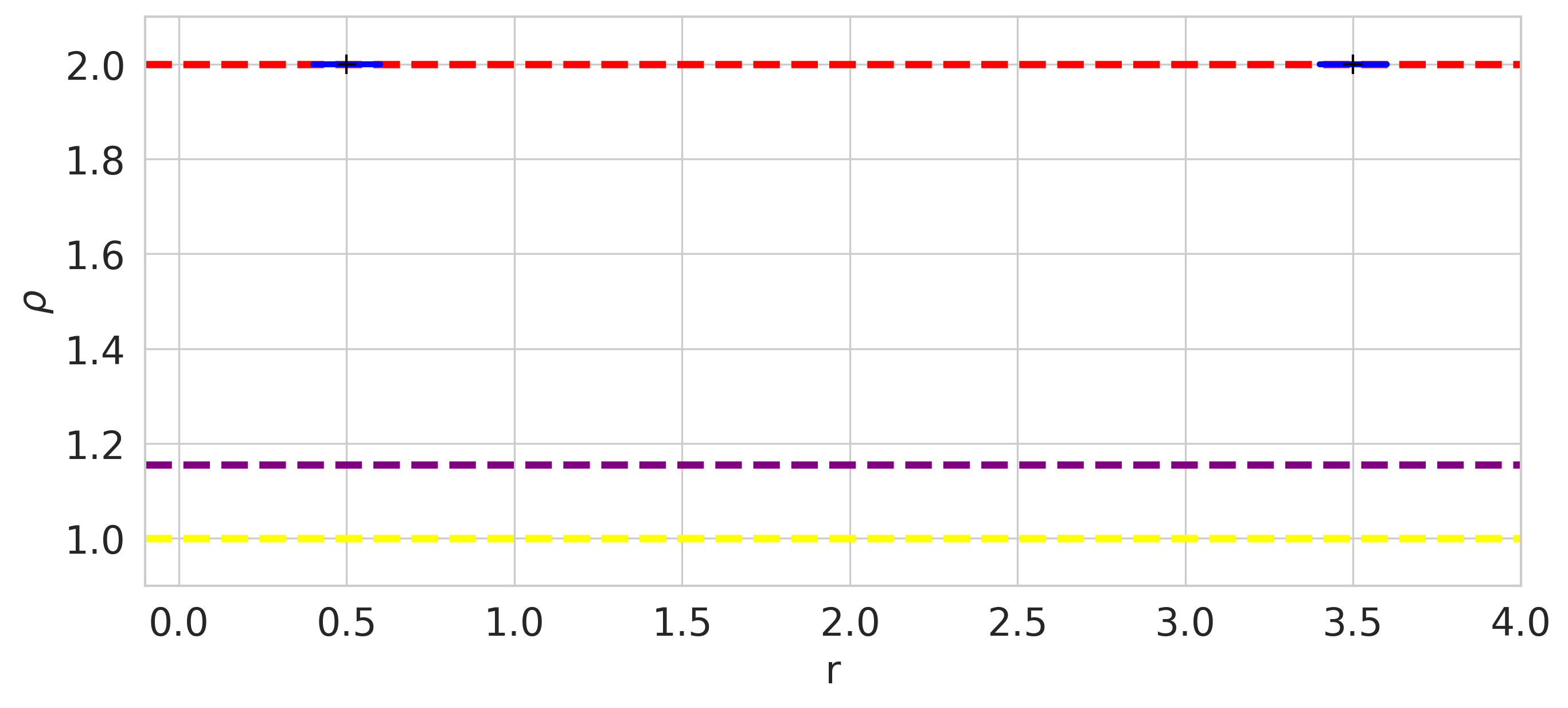} 
      \\ 
      \\ 
 \bottomrule
     \end{tabular}
 \end{table}
Having validated the method on controlled toy examples that represent canonical geometric models, we now turn to more complex, high-dimensional datasets to assess its performance in practical settings. In particular, we compute curvature profiles for benchmark datasets including MNIST \cite{lecun1998gradient}, FashionMNIST \cite{xiao2017fashionmnist}, and  CIFAR10 \cite{Krizhevsky09} datasets. For each of these datasets, we consider a sample of size 
$10,000$, employing the adaptive neighborhood structure to define the underlying geometry, since the ambient distribution density is unknown. The resulting curvature profiles are summarized in  \cref{tab:metprof}. Along with these benchmark datasets \cref{tab:metprof} also includes two additional datasets: mammoth (10000 samples of 3-dimensional coordinates), and breast cancer (570 samples, 32 features). 
\begin{table}[H]
 \caption{Curvature profile for metric datasets with $(k_{\min}, k_{\max})=(15,20)$.}
  \label{tab:metprof}
\centering
\begin{tabular}{ c | c c c}
    \toprule
      \multicolumn{2}{c}{averaged}   & non-averaged &  boxplot \\		
       \midrule
       MNIST   & \includegraphics[width=0.25\linewidth]{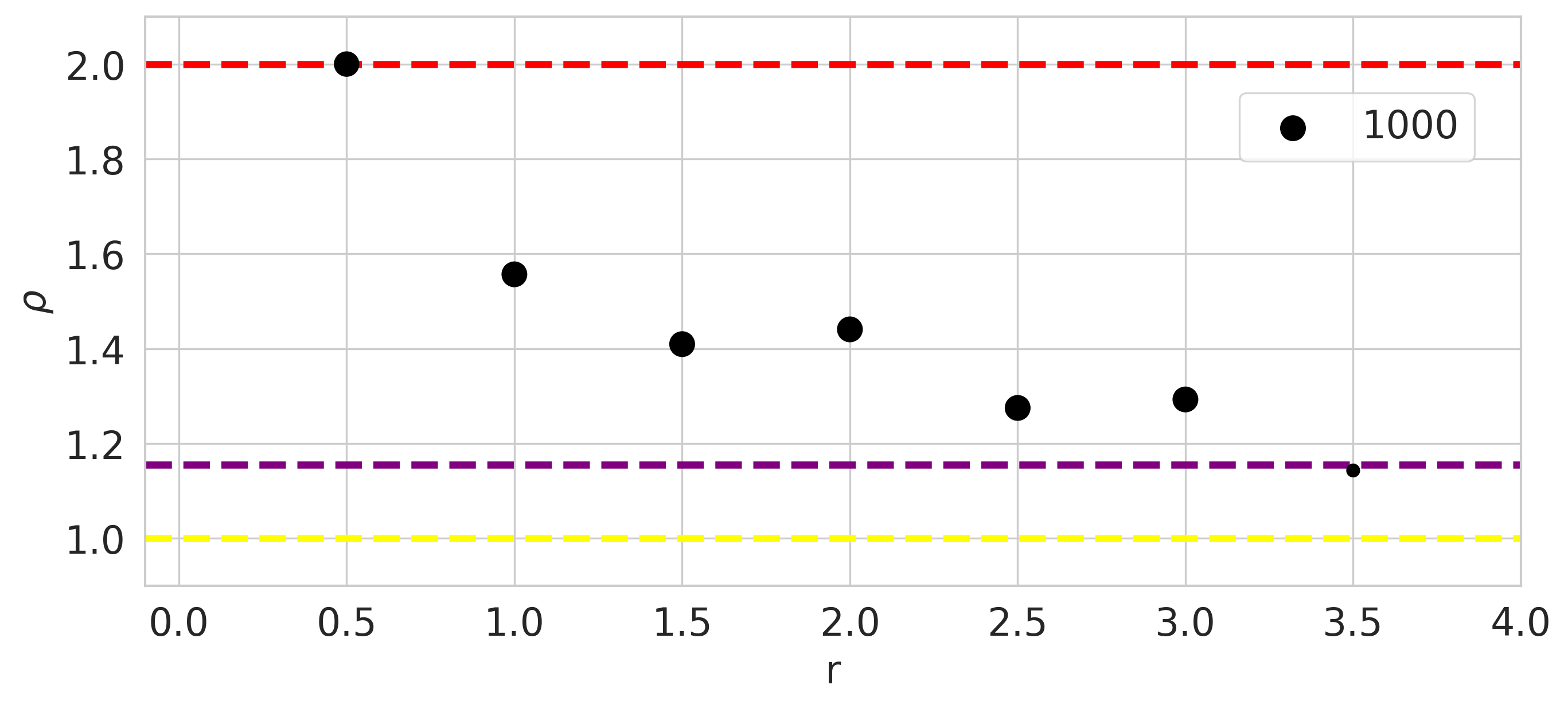}
      & \includegraphics[width=0.25\linewidth]{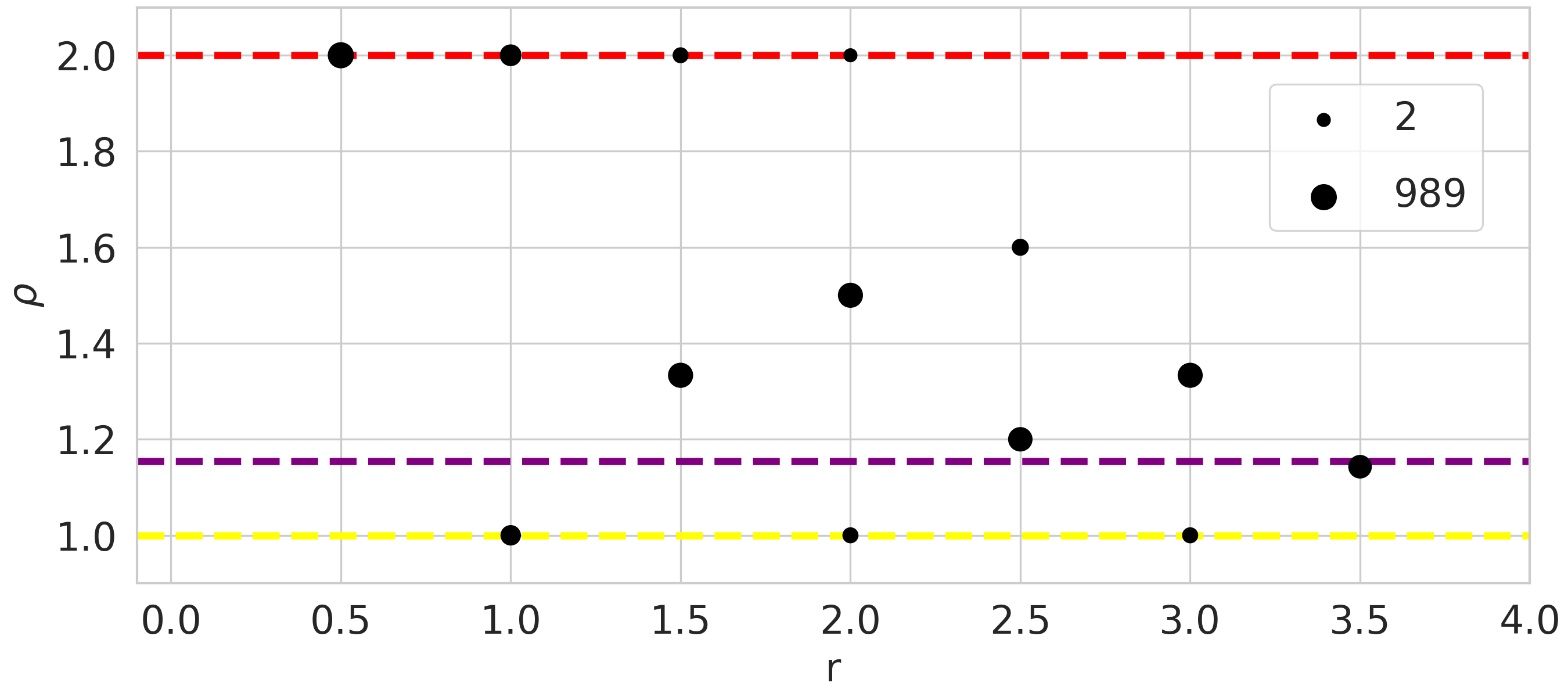}
      &  \includegraphics[width=0.25\linewidth]{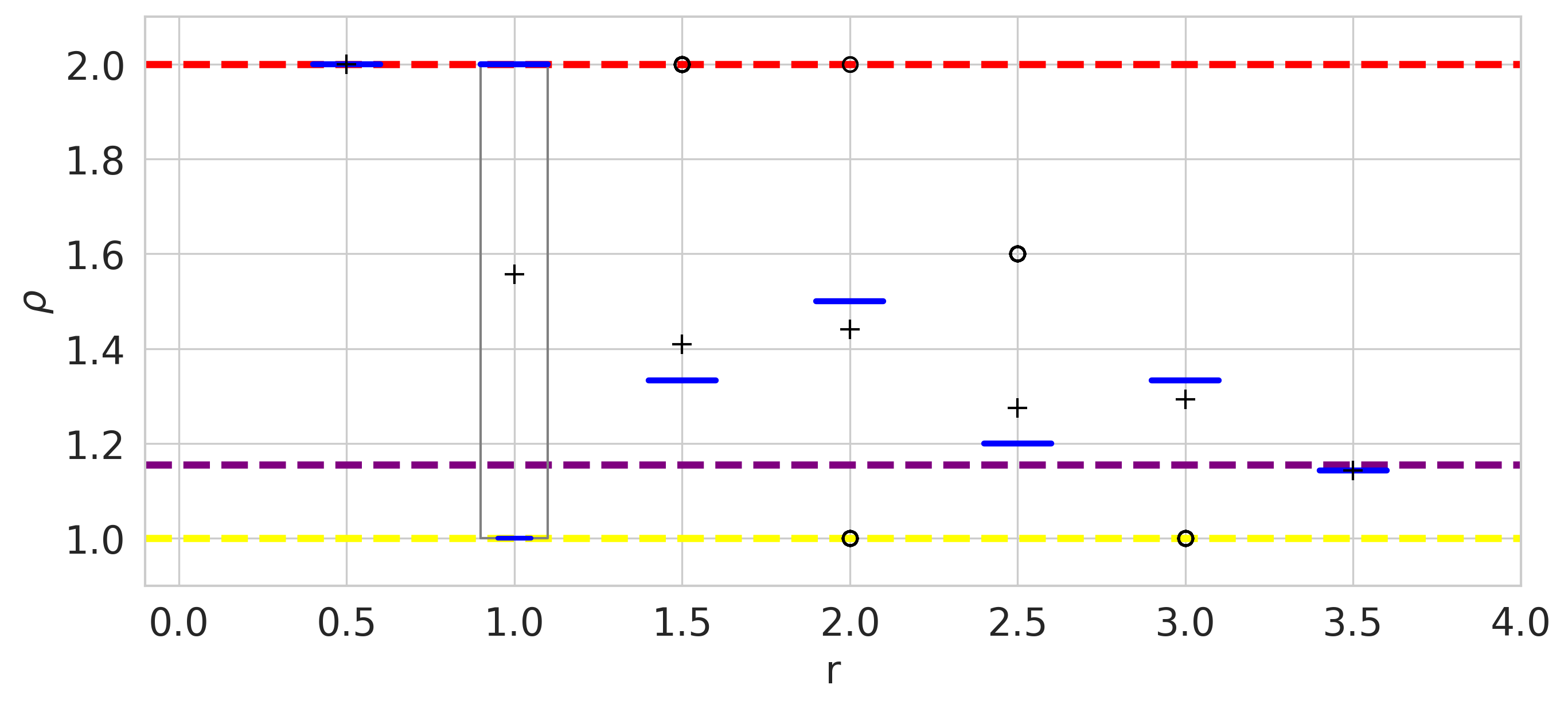} 
      \\ 
       \midrule 
       FMNIST   & \includegraphics[width=0.25\linewidth]{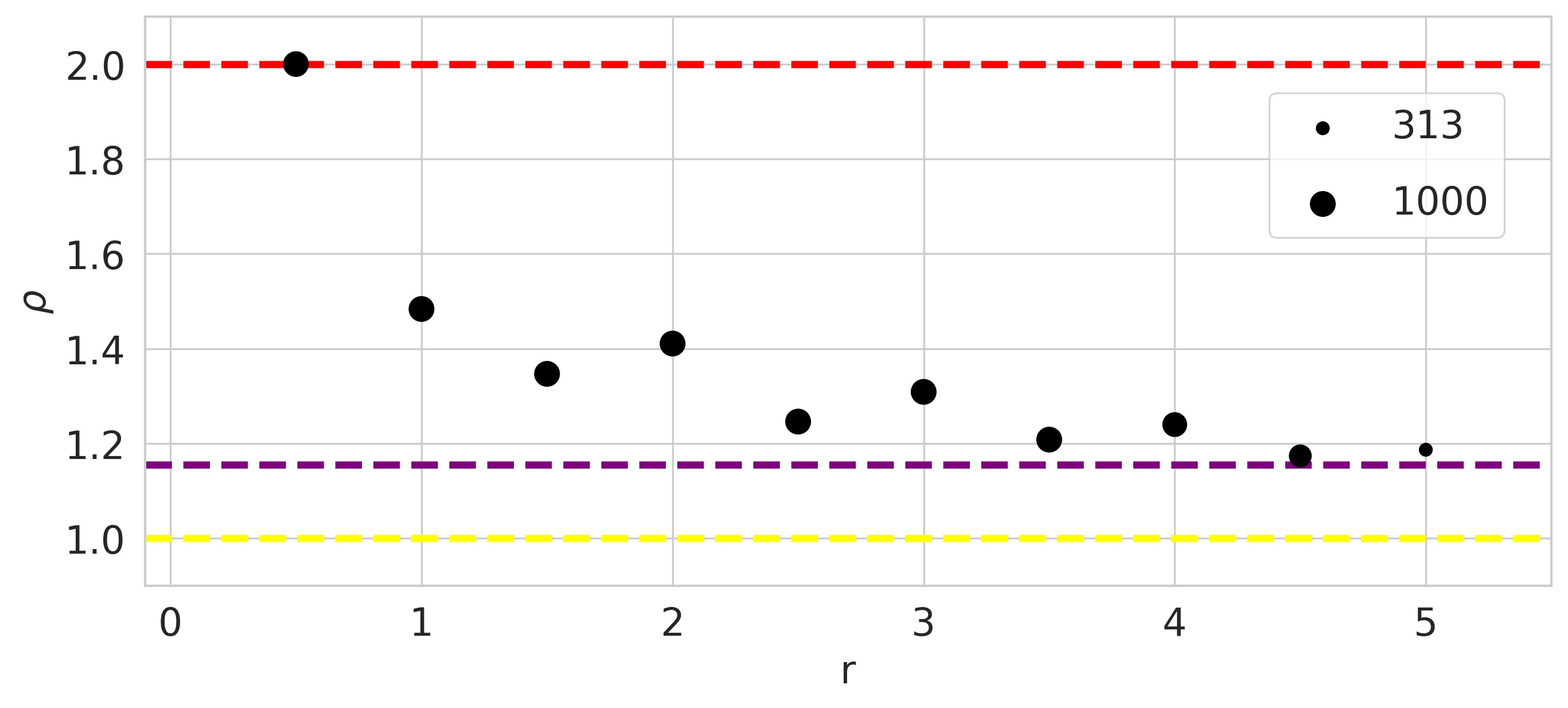}
       & \includegraphics[width=0.25\linewidth]{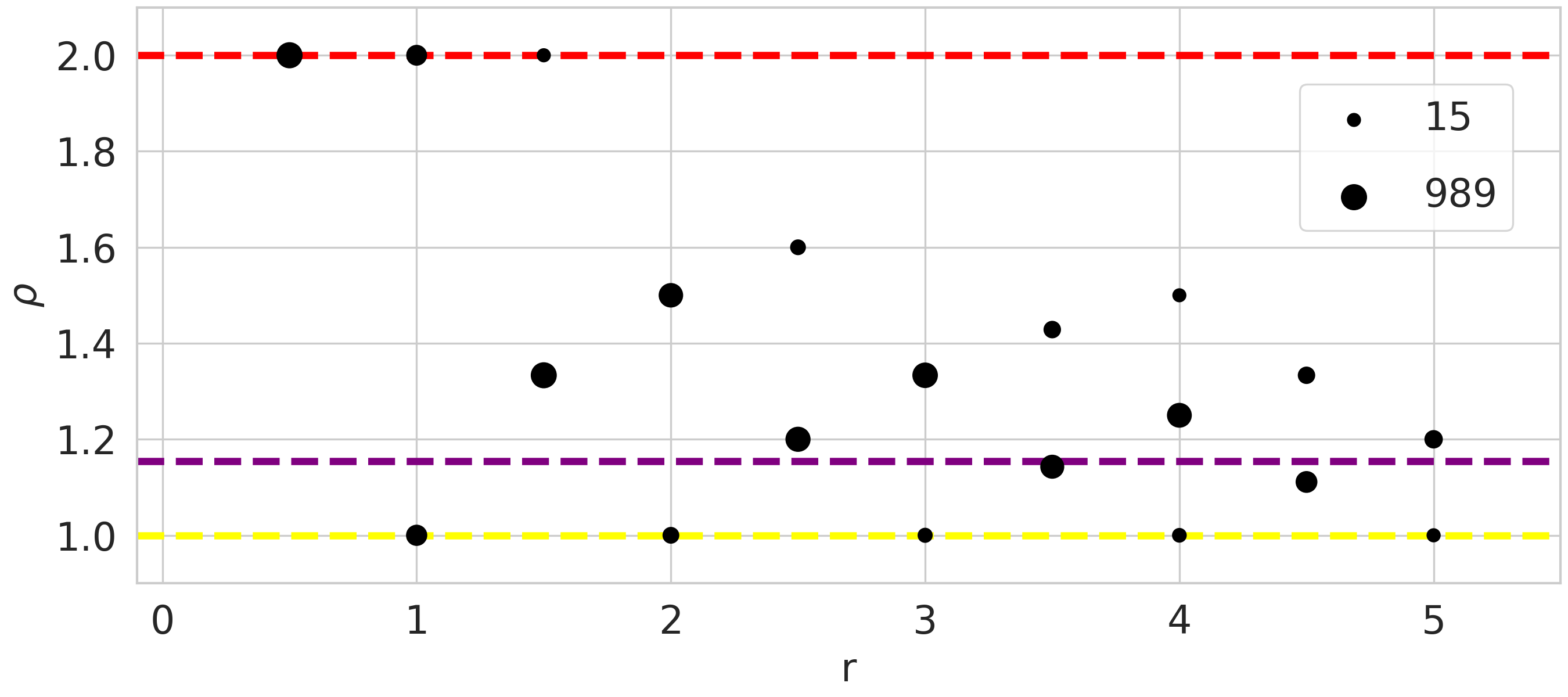}
      &  \includegraphics[width=0.25\linewidth]{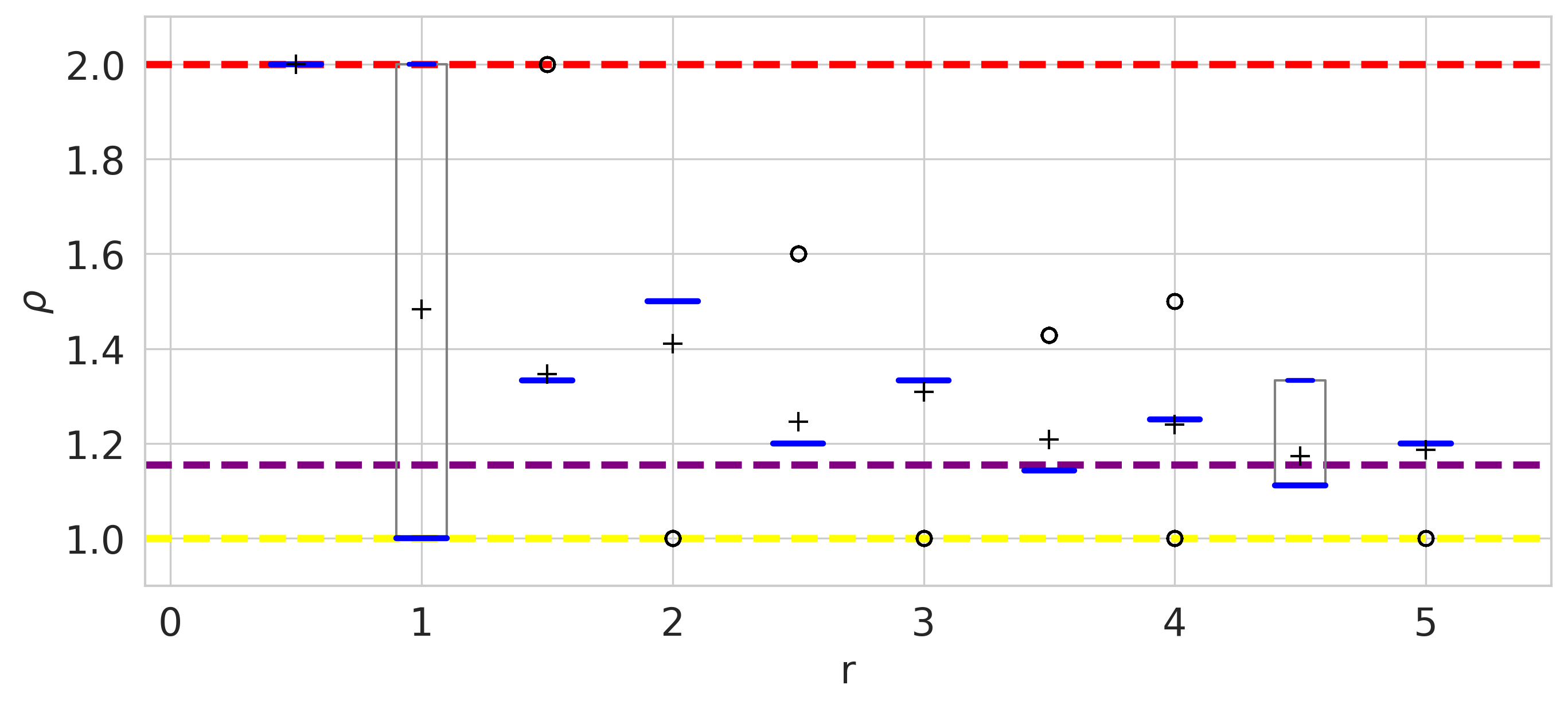} 
      \\ 
       \midrule 
       CIFAR10   & \includegraphics[width=0.25\linewidth]{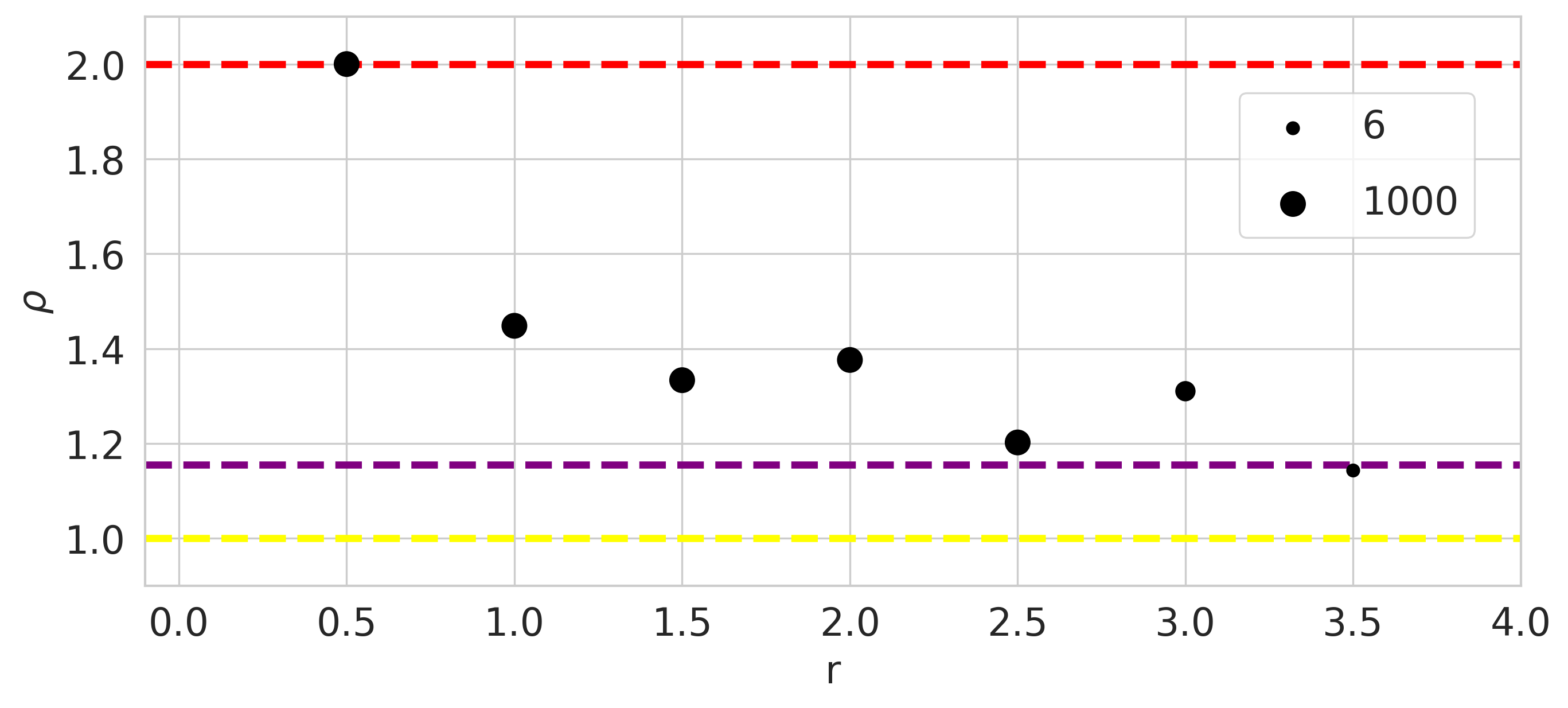}
       & \includegraphics[width=0.25\linewidth]{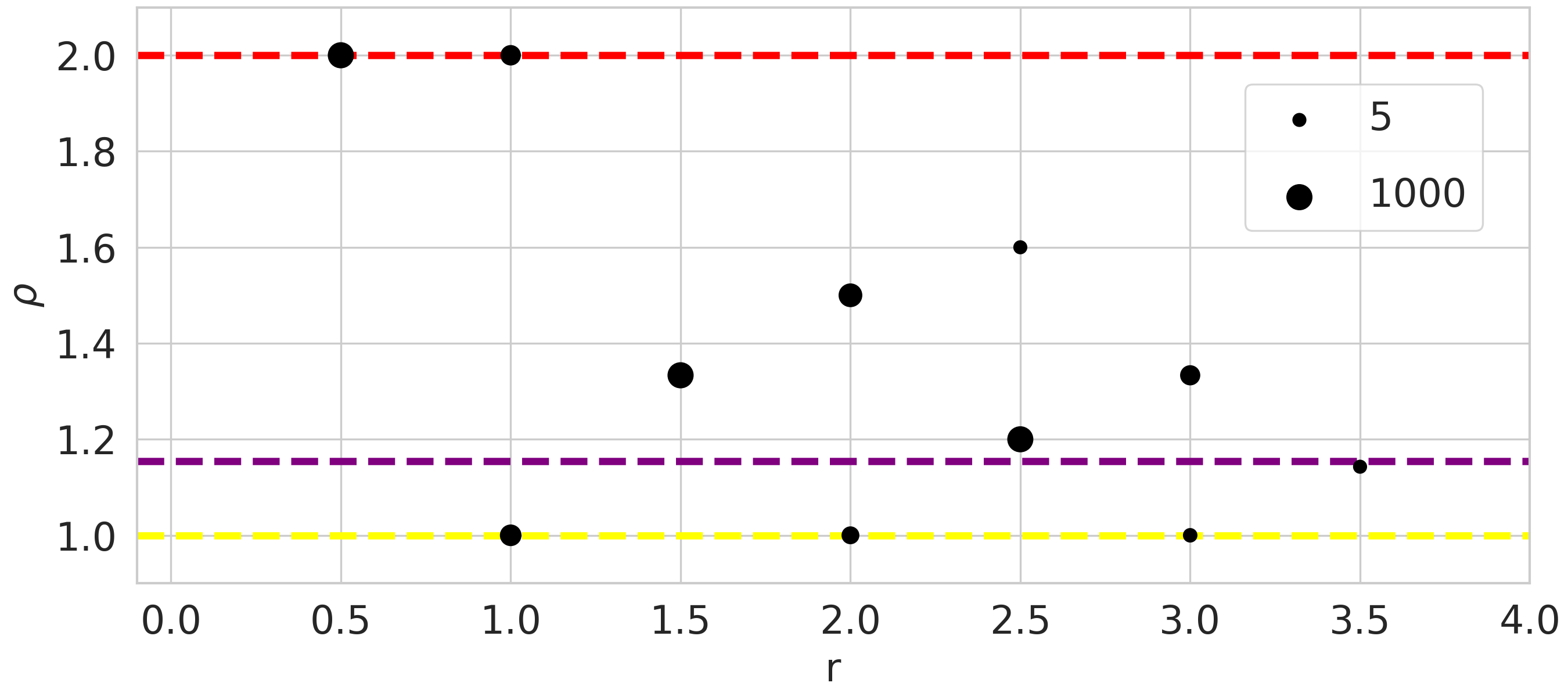}
      &  \includegraphics[width=0.25\linewidth]{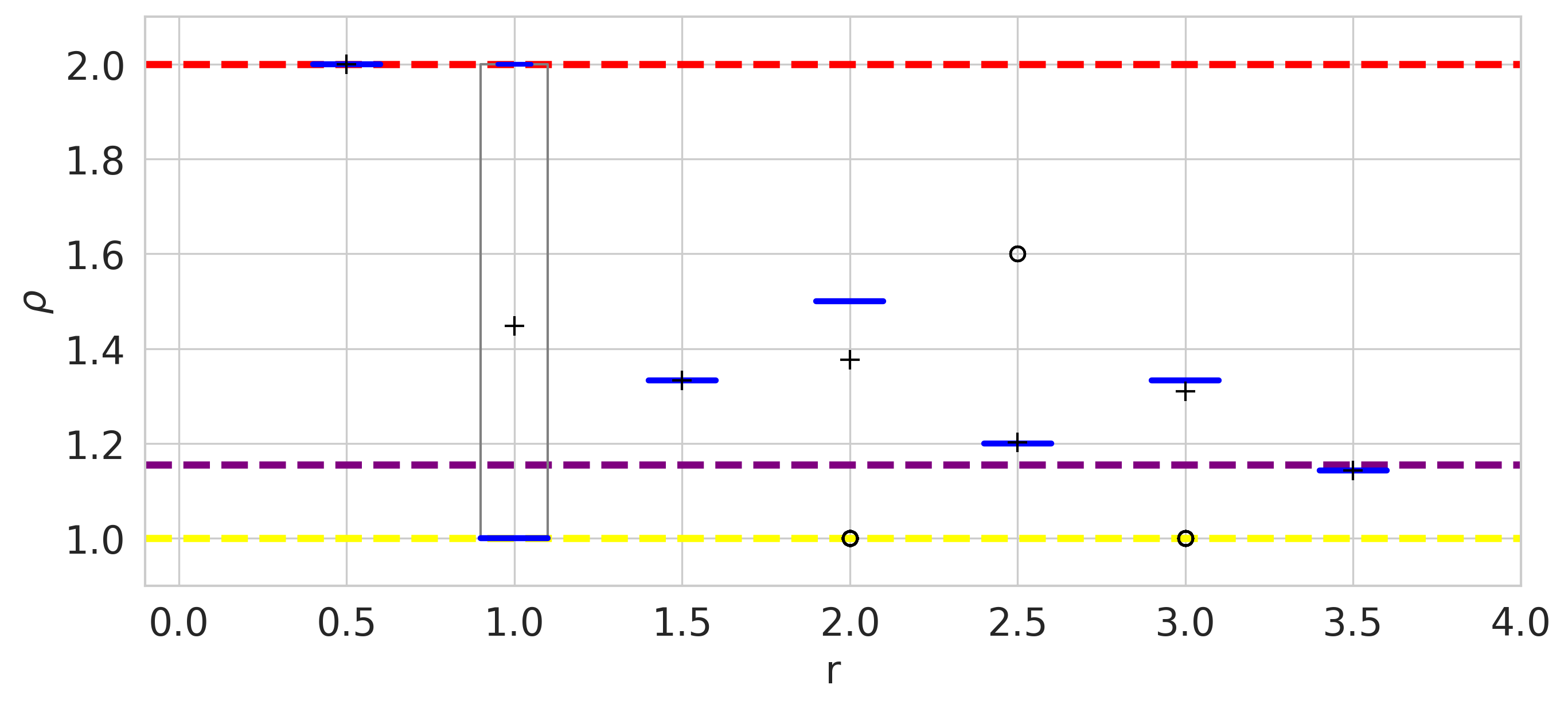} 
      \\ 
      \midrule 
       BreastC   & \includegraphics[width=0.25\linewidth]{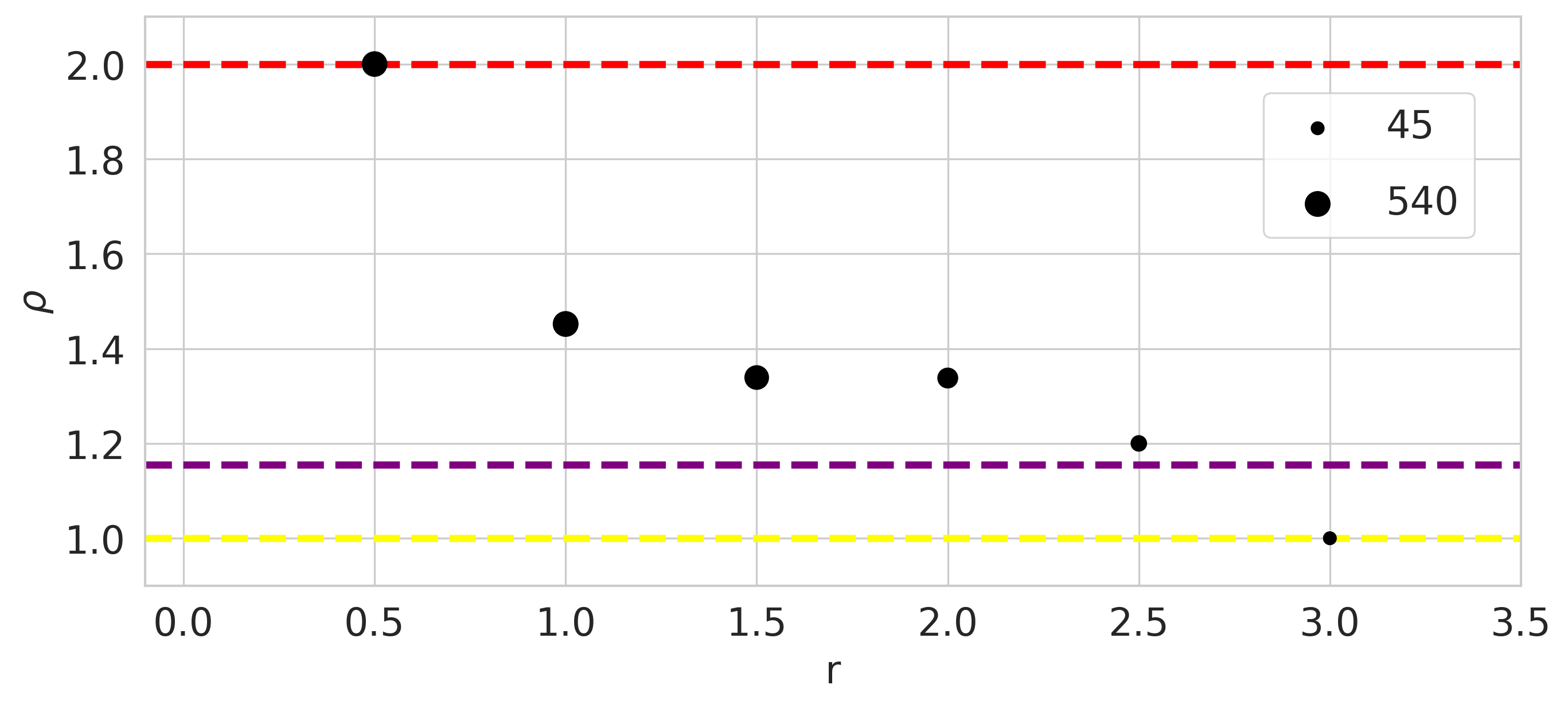}
       & \includegraphics[width=0.25\linewidth]{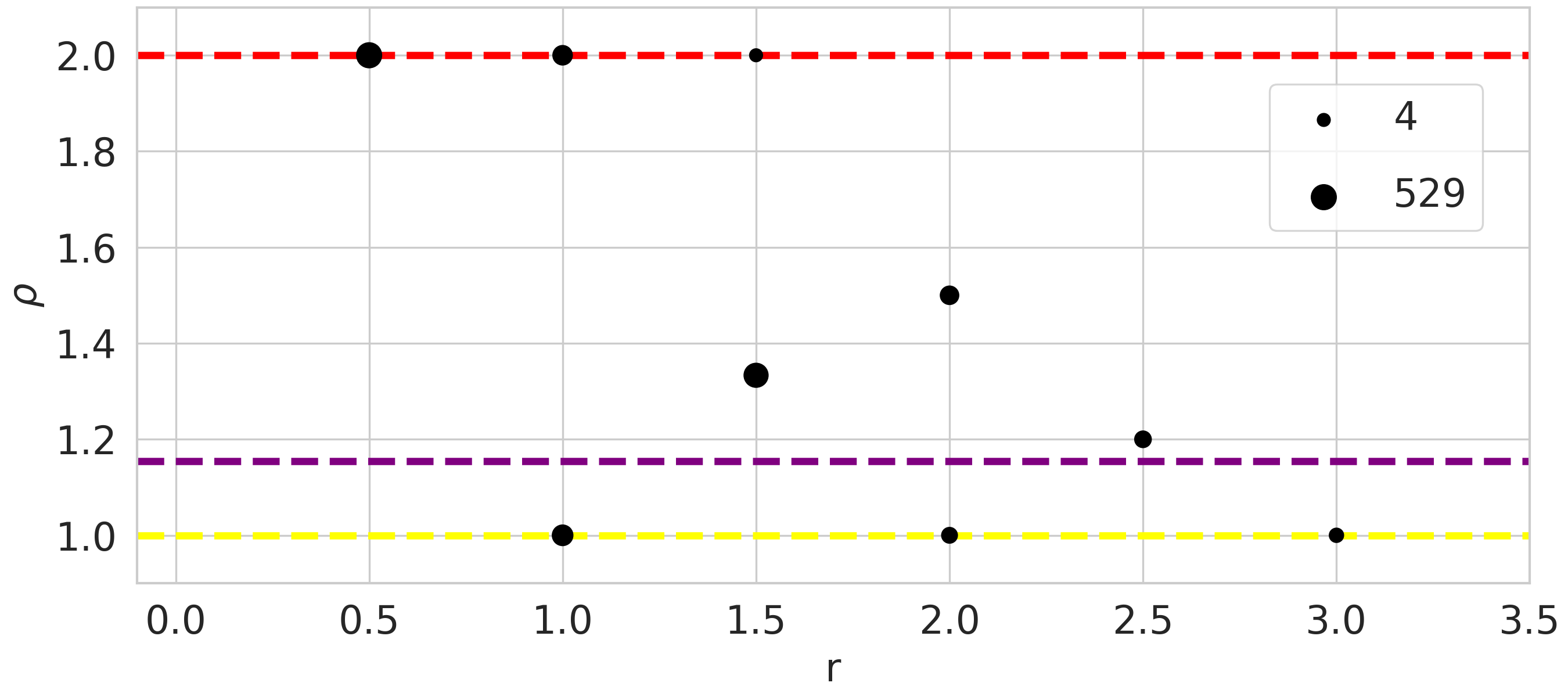}
      &  \includegraphics[width=0.25\linewidth]{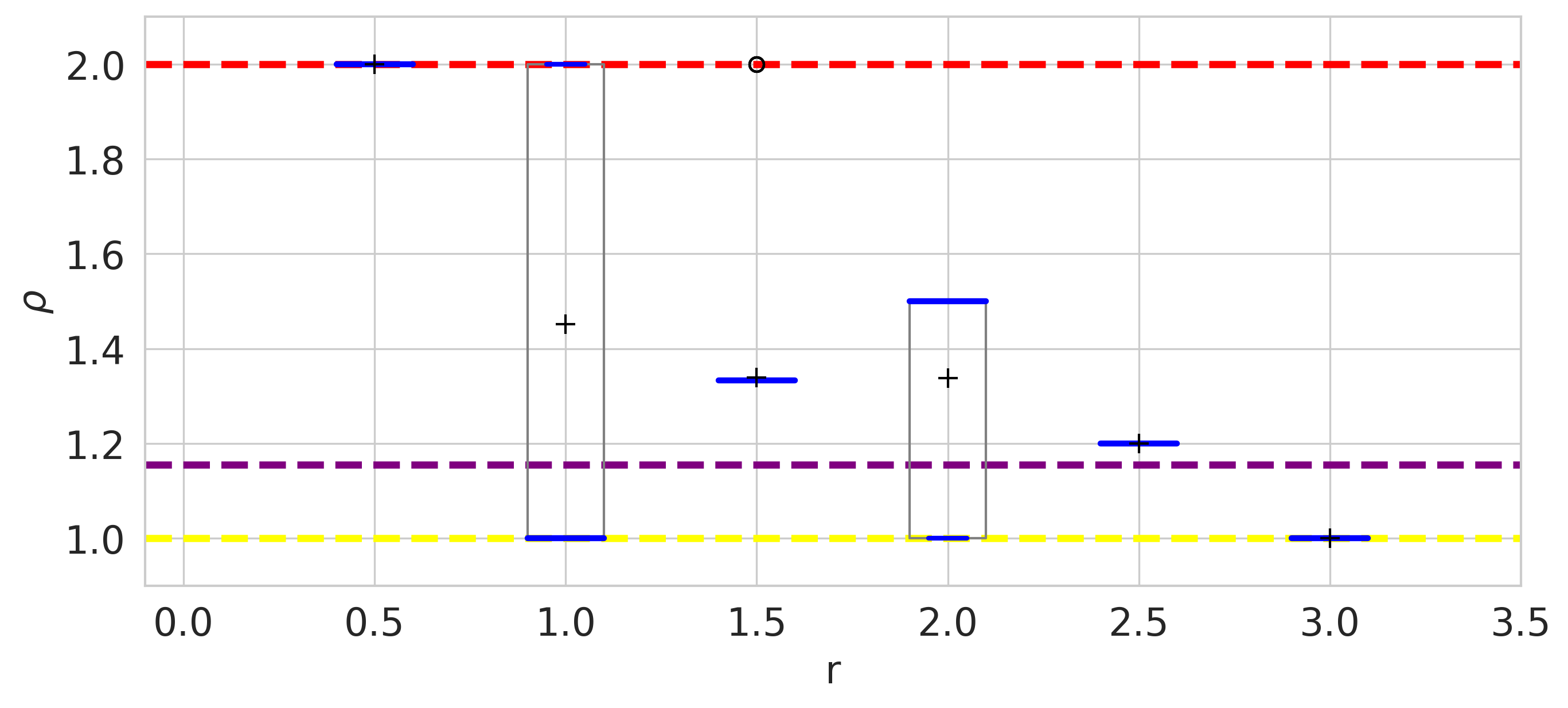} 
      \\ 
      \midrule 
       Mammoth   & \includegraphics[width=0.25\linewidth]{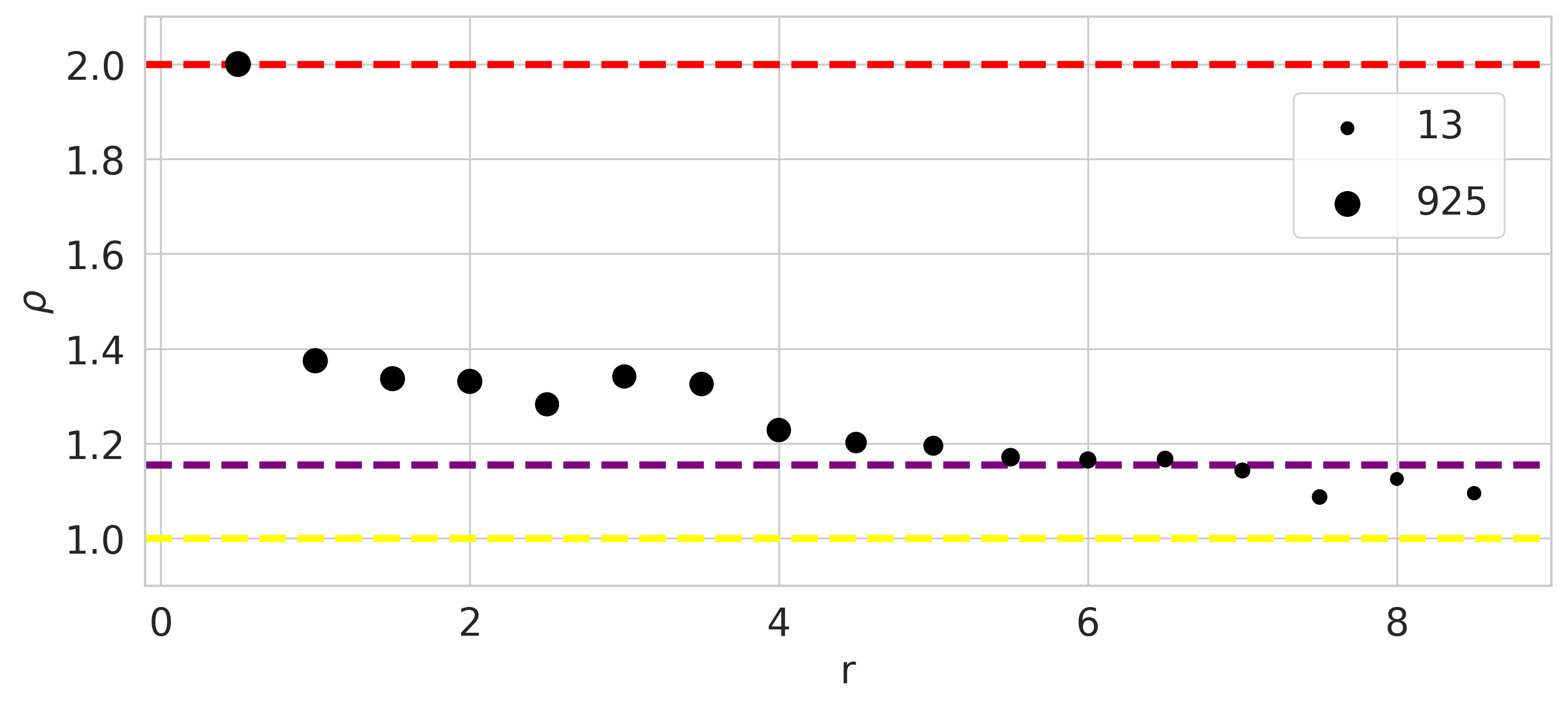}
       & \includegraphics[width=0.25\linewidth]{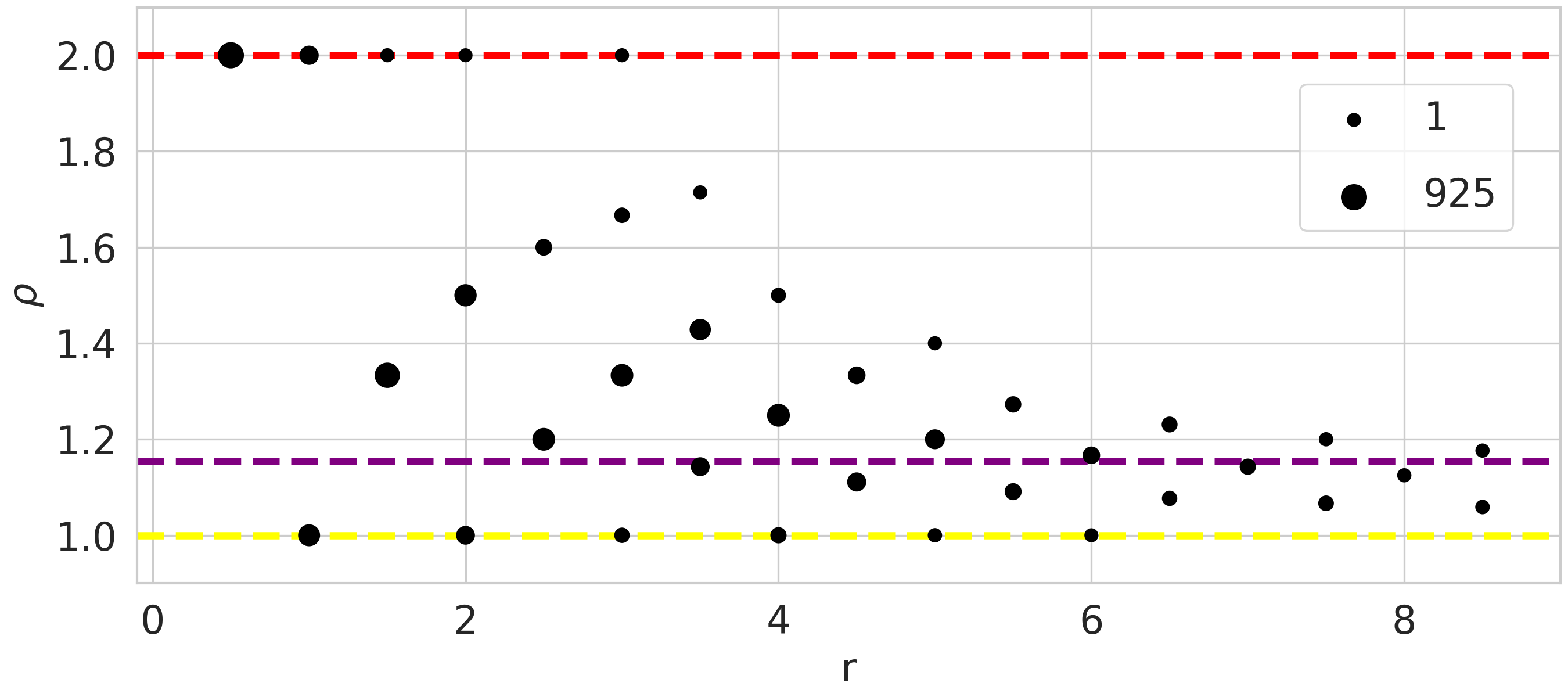}
      &  \includegraphics[width=0.25\linewidth]{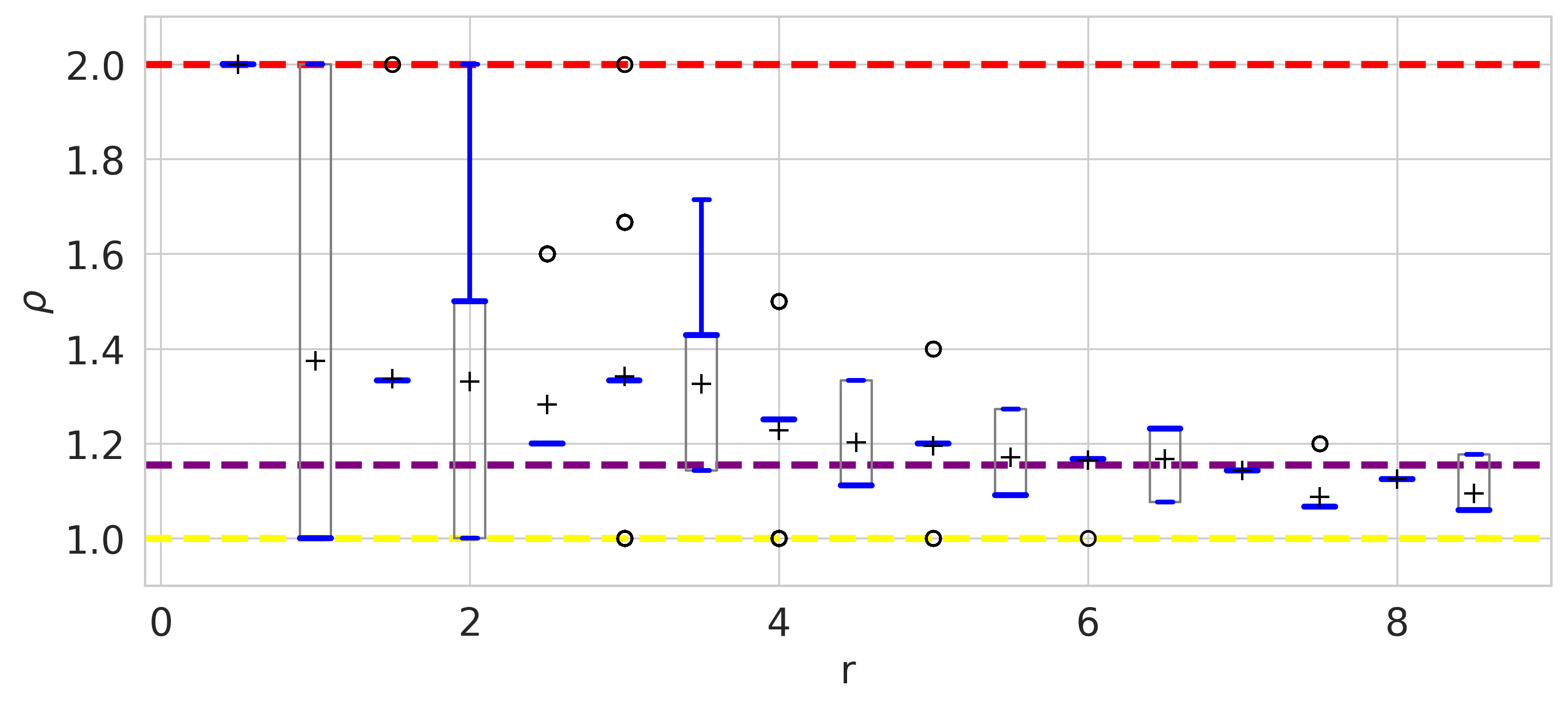} 
      \\ 
 \bottomrule
     \end{tabular}
 \end{table}
How should we interpret the curvature proxy statistics?  We see here only  how curvature behavior evolves across different scales. Nevertheless, this scale-dependent analysis provides a preliminary understanding of the geometric organization of these datasets, potentially offering insights into their intrinsic structure which we can be utilized to check the efficiency of representations of datasets or inferring the intrinsic dimension as we will investigate in the following. 
\subsection{Evaluation of dimensionality reduction methods via the curvature profile}\label{sec:dimreduction} 
Empirical datasets usually consist of a finite set $X$ of  elements associated with some $n$ observed scalar values, and so, $X$ can be embedded in the Cartesian space $\R^n$, and distances are often computed using the Euclidean metric. This metric need not be intrinsic to the data -- Euclidean distances between pairs of points may not reflect connectivity paths that traverse through other points in $X$.  Thus, for capturing local connectivity among data points, neighborhood graphs are commonly constructed. Most dimensionality reduction techniques aim to preserve this graph-based connectivity structure. Our curvature profile inference method is designed to extract geometric information from such graph representations. Analysis of the curvature profile (through averaged or unaveraged distributions of $\rho$) offers a principled way to compare the geometry of the original dataset with that of its low-dimensional embedding. 
To compare  curvature profiles, we use the $1$-Wasserstein distance  (also known as the Earth Mover's Distance or optimal transport distance), which quantifies the minimal effort required to transform one distribution into another \cite{villani2008optimal}. For this purpose, the $\rho$ distributions of both the original dataset and its embedding (obtained via a dimensionality reduction method)are first interpolated on a common meshgrid constructed using a $kd-$tree \cite{2020SciPy-NMeth,maneewongvatana1999analysis}. The $W_{1}$ distance is then computed between these fitted distributions, with probability masses defined by the number of triangles associated with each $(r,\rho)-$coordinates. This procedure is repeated for all pairs  $(k_{min},k_{max})$ used in constructing the graphs, across all embedding dimensions. In this way, we quantify the impact of these experimental choices on the consistency of curvature profiles under embedding. \\

Here we select three dimensionality reduction methods,  UMAP \cite{McInnes18}, Isomap \cite{Tenenbaum00}, and IsUMap \cite{Joharinad25}, and plot $W_1$ against dimension for benchmark datasets, including MNIST, FashionMNIST, CIFAR10, as well as and Breast cancer patients' dataset \cite{wolberg1993diagnostic} and Mammoth \cite{smithsonian_mammuthus_2020} for multiple parameters $(k_{\min},k_{\max})$. The plots are shown in \cref{tab:Wbench}. Having tried multiple choices of the parameter for constructing the neighborhood graph of the input data and the output of the respective dimensionality reduction method, eventually, we chose the adaptive model for the former and the vanilla model (the typical $k$-neighborhood graph setting $k=k_{\min}$) for the latter.

\begin{table}[ht]
  \caption{The $W_1$ distance between the curvature profiles before and after dimensionality reduction across dimension $d$}
  \label{tab:Wbench}
  \centering
  \begin{tabular}{ c  c c c }
    \toprule
    \multicolumn{1}{c}{} & \multicolumn{1}{c}{10–15} & \multicolumn{1}{c}{15–20} & \multicolumn{1}{c}{20–30} \\
    \midrule
    \raisebox{3.5\height}{MNIST} & 
    \includegraphics[width=0.26\linewidth]{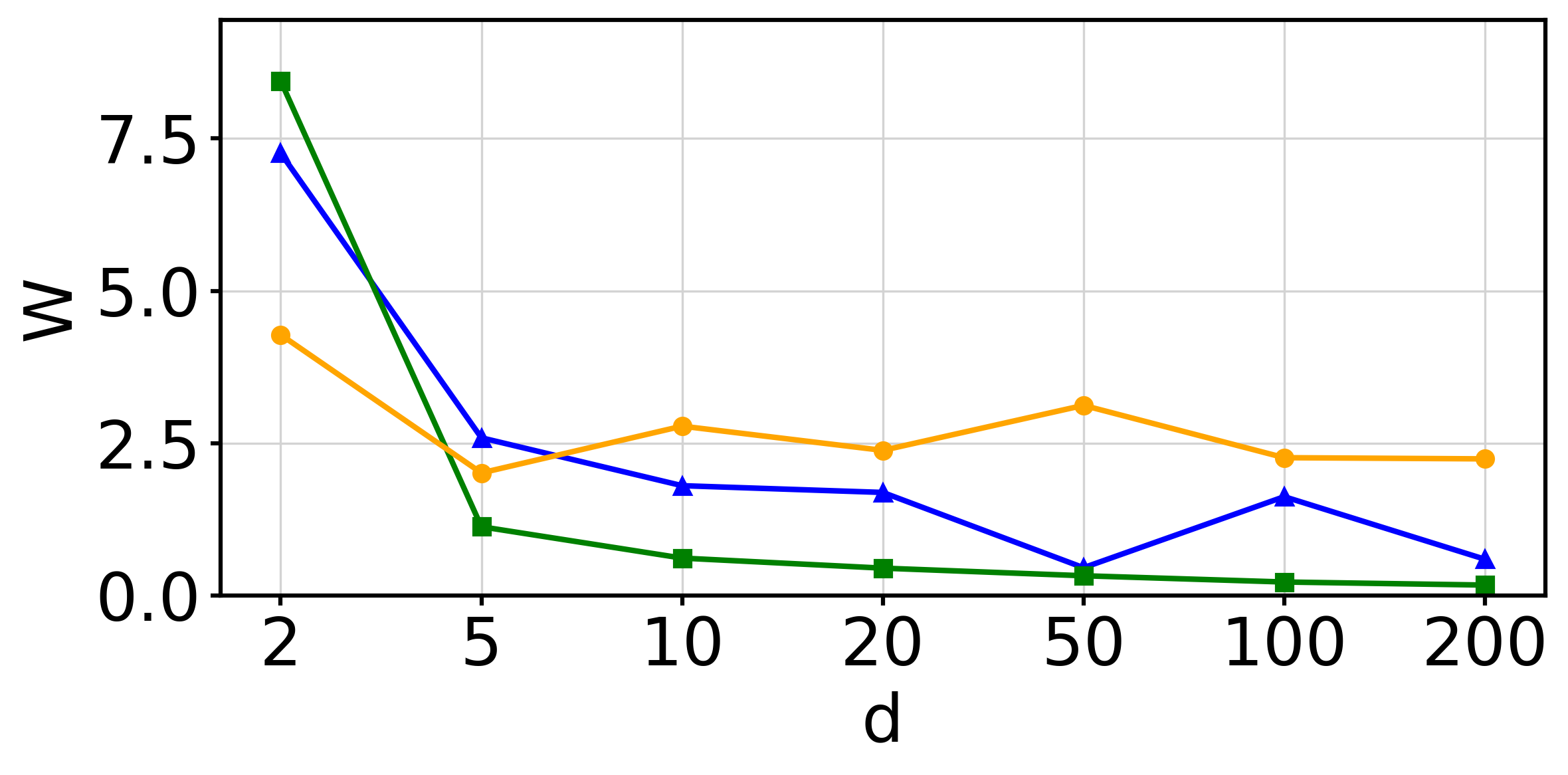} &
    \includegraphics[width=0.26\linewidth]{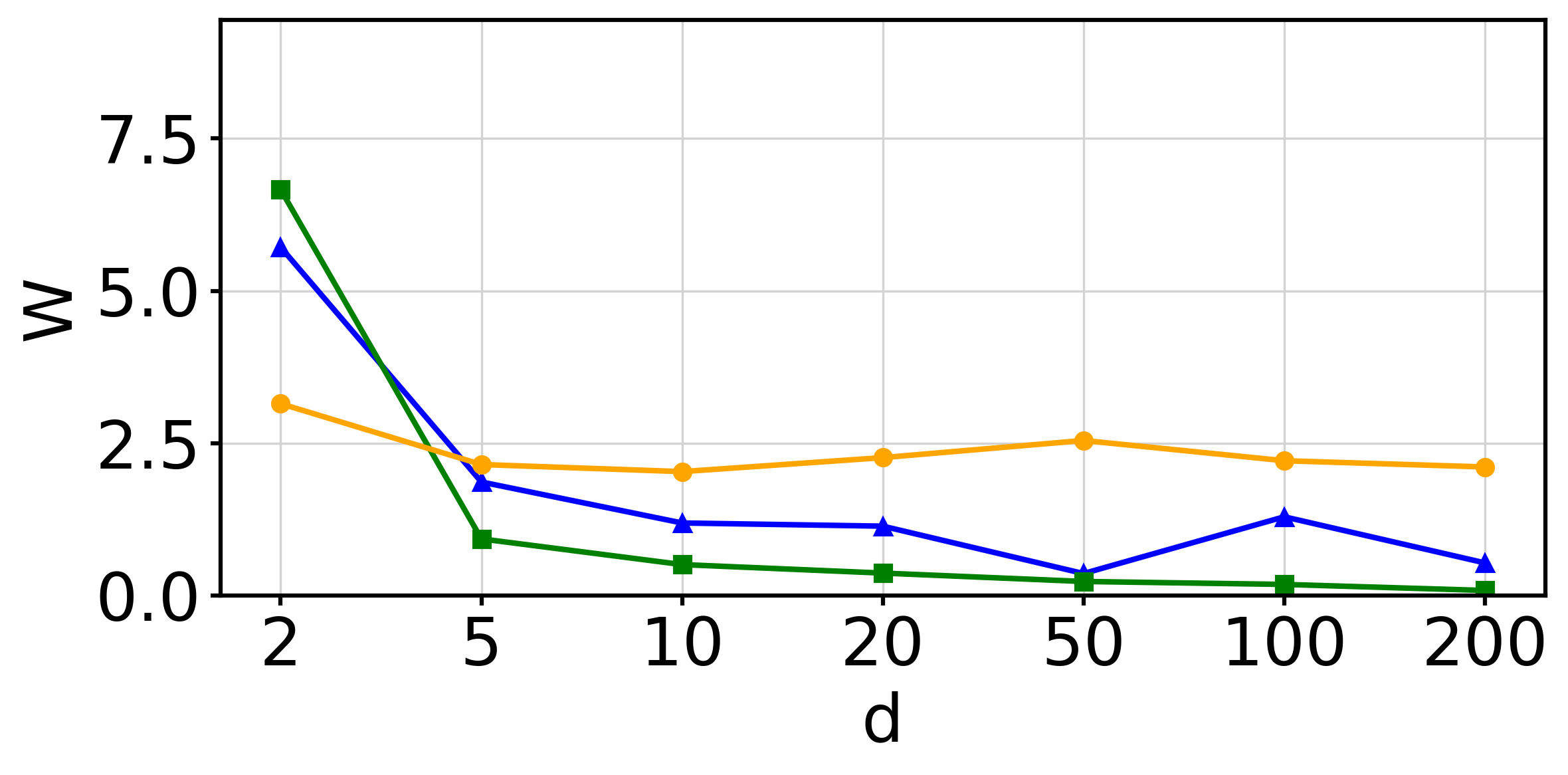} &
    \includegraphics[width=0.26\linewidth]{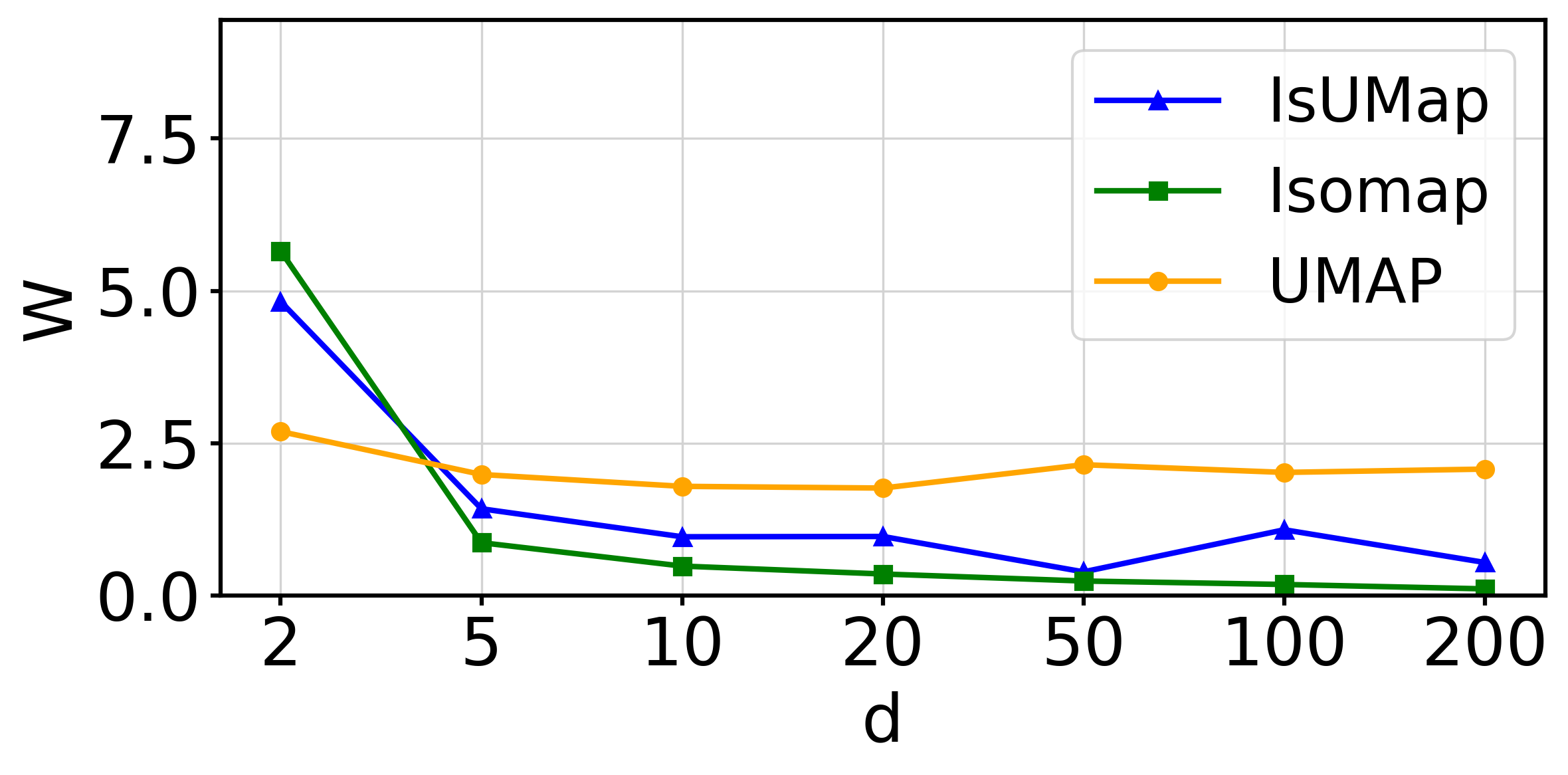} \\
    \midrule
    \raisebox{3.5\height}{FMNIST} & 
    \includegraphics[width=0.26\linewidth]{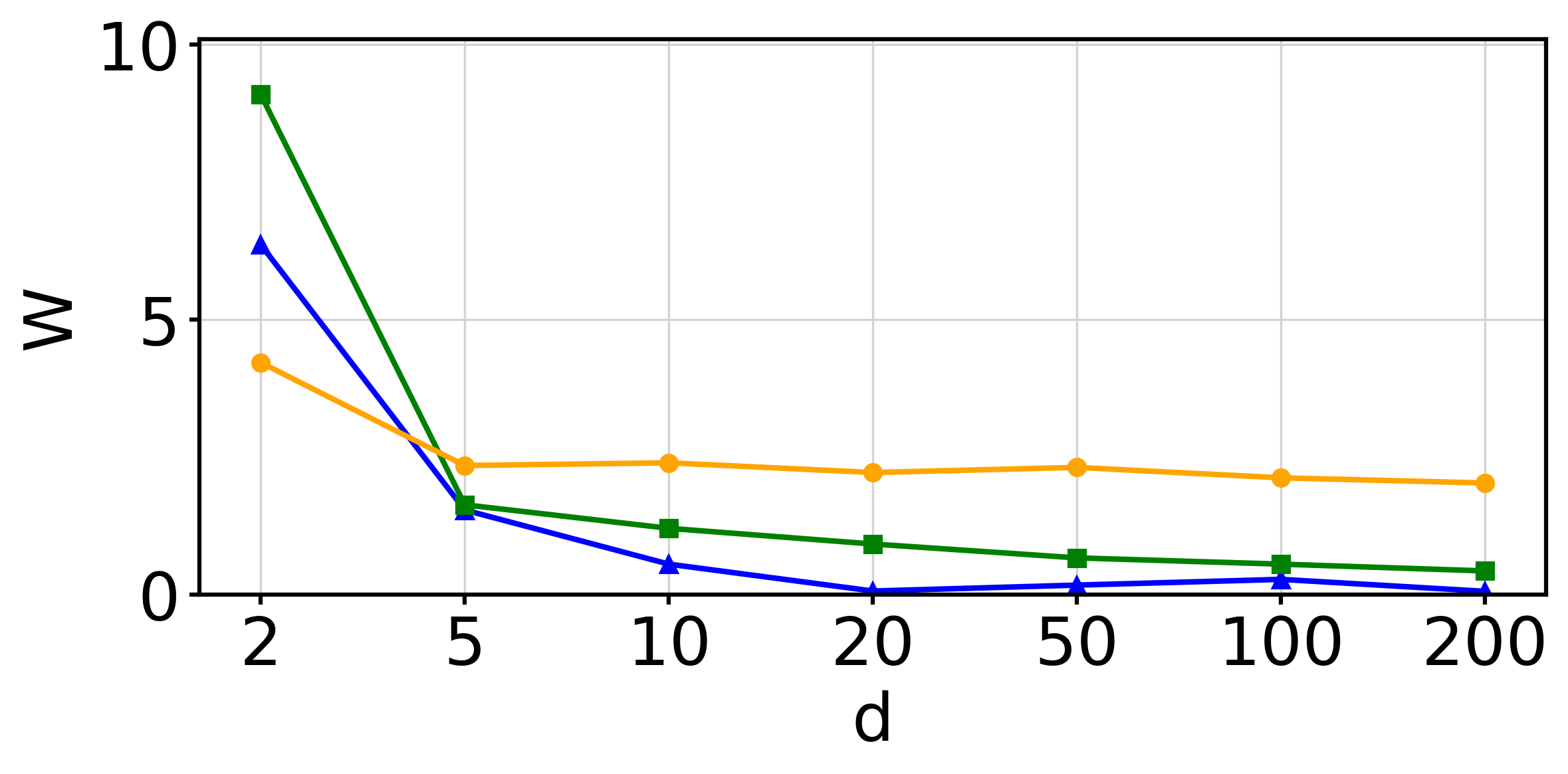} &
    \includegraphics[width=0.26\linewidth]{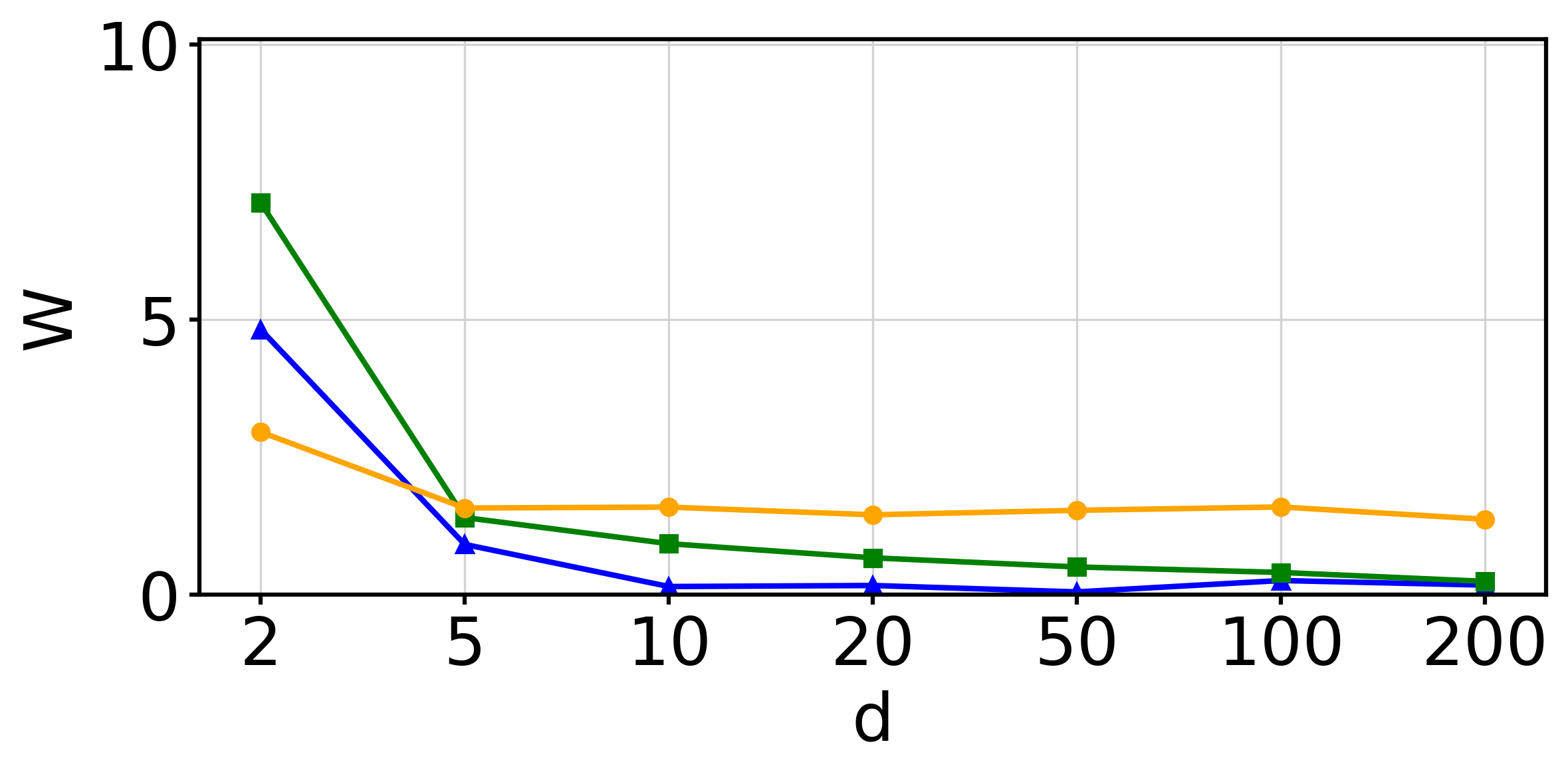} &
    \includegraphics[width=0.26\linewidth]{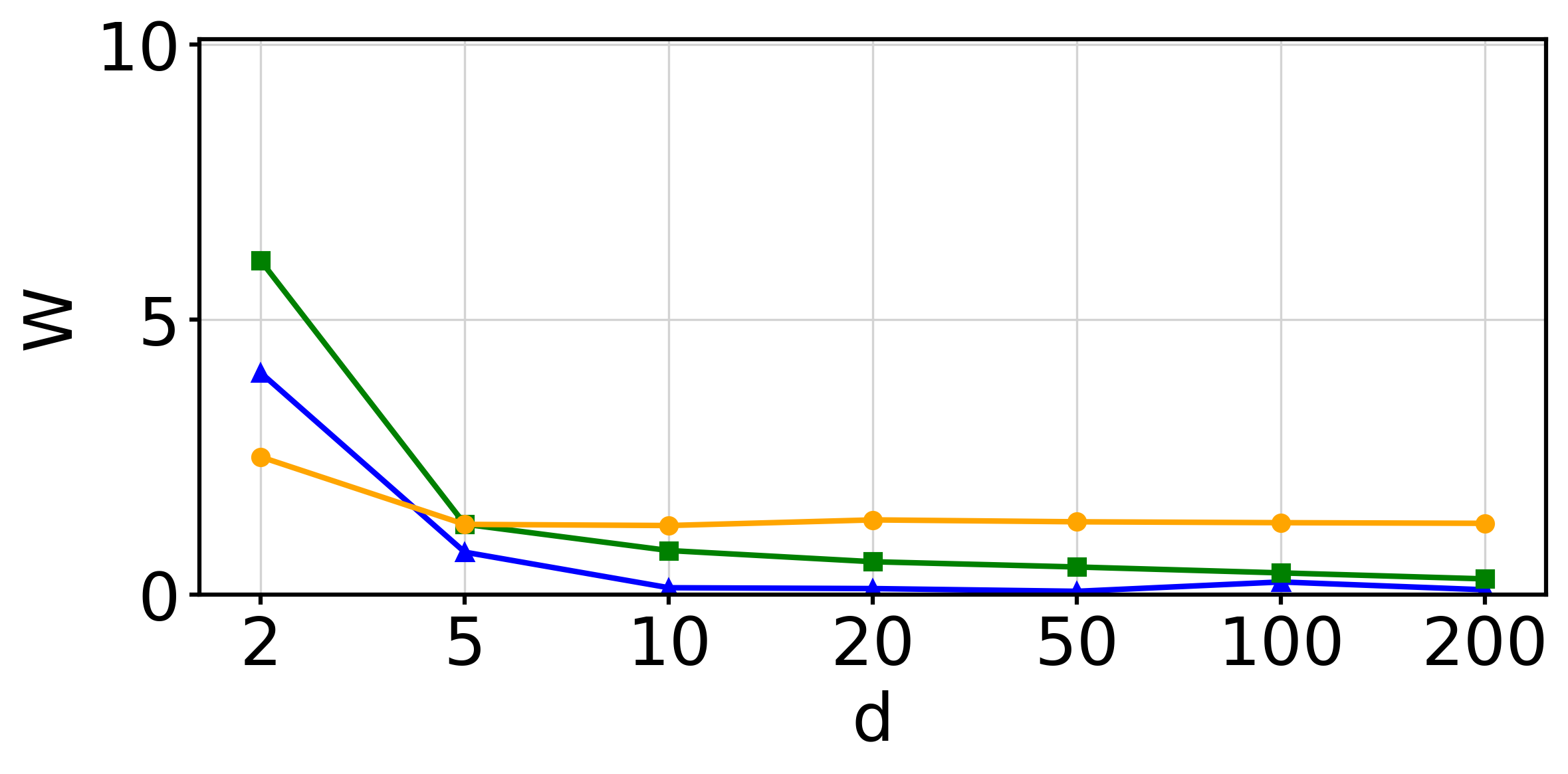} \\
    \midrule
    \raisebox{3.5\height}{CIFAR10} & 
    \includegraphics[width=0.26\linewidth]{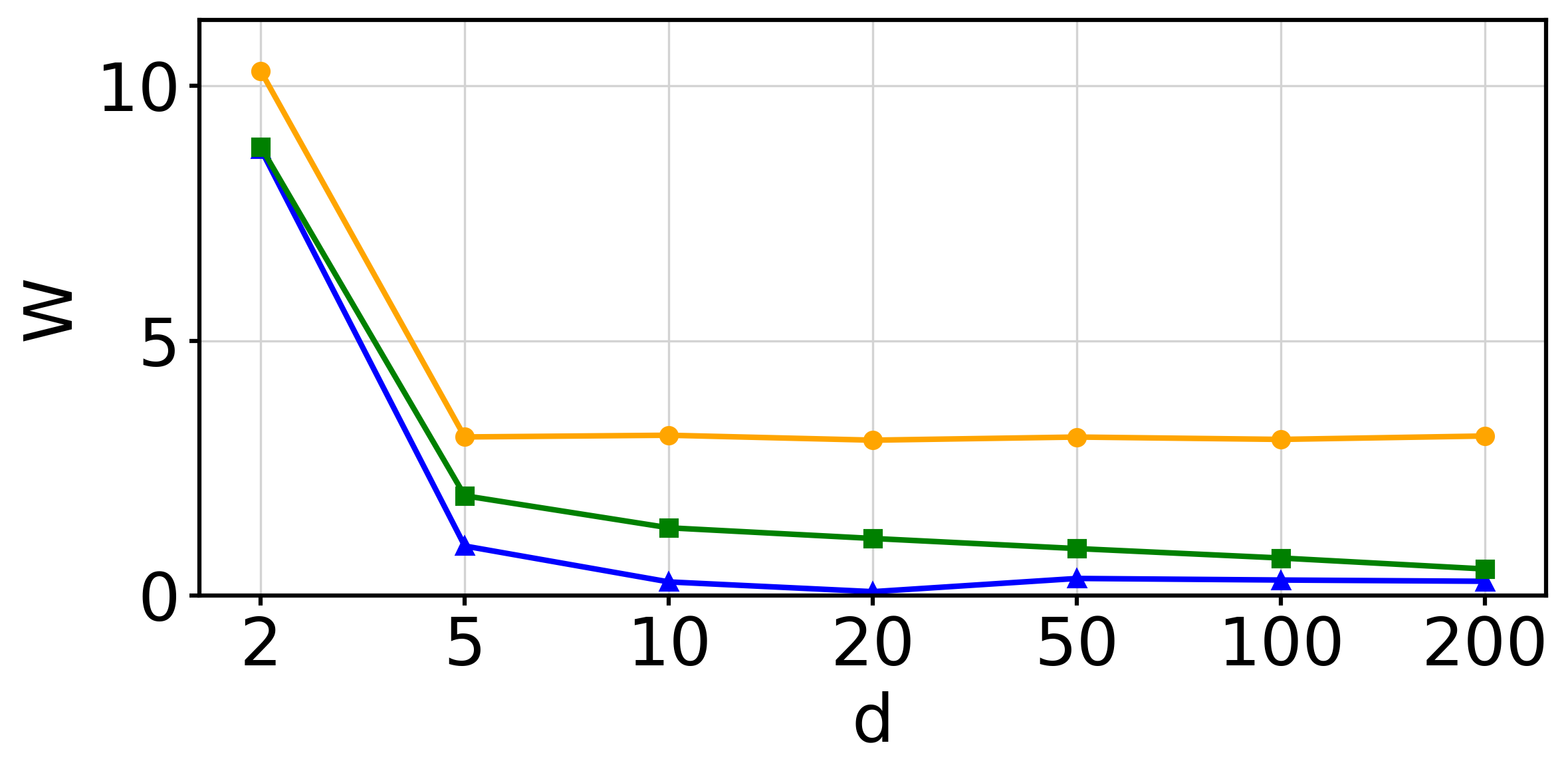} &
    \includegraphics[width=0.26\linewidth]{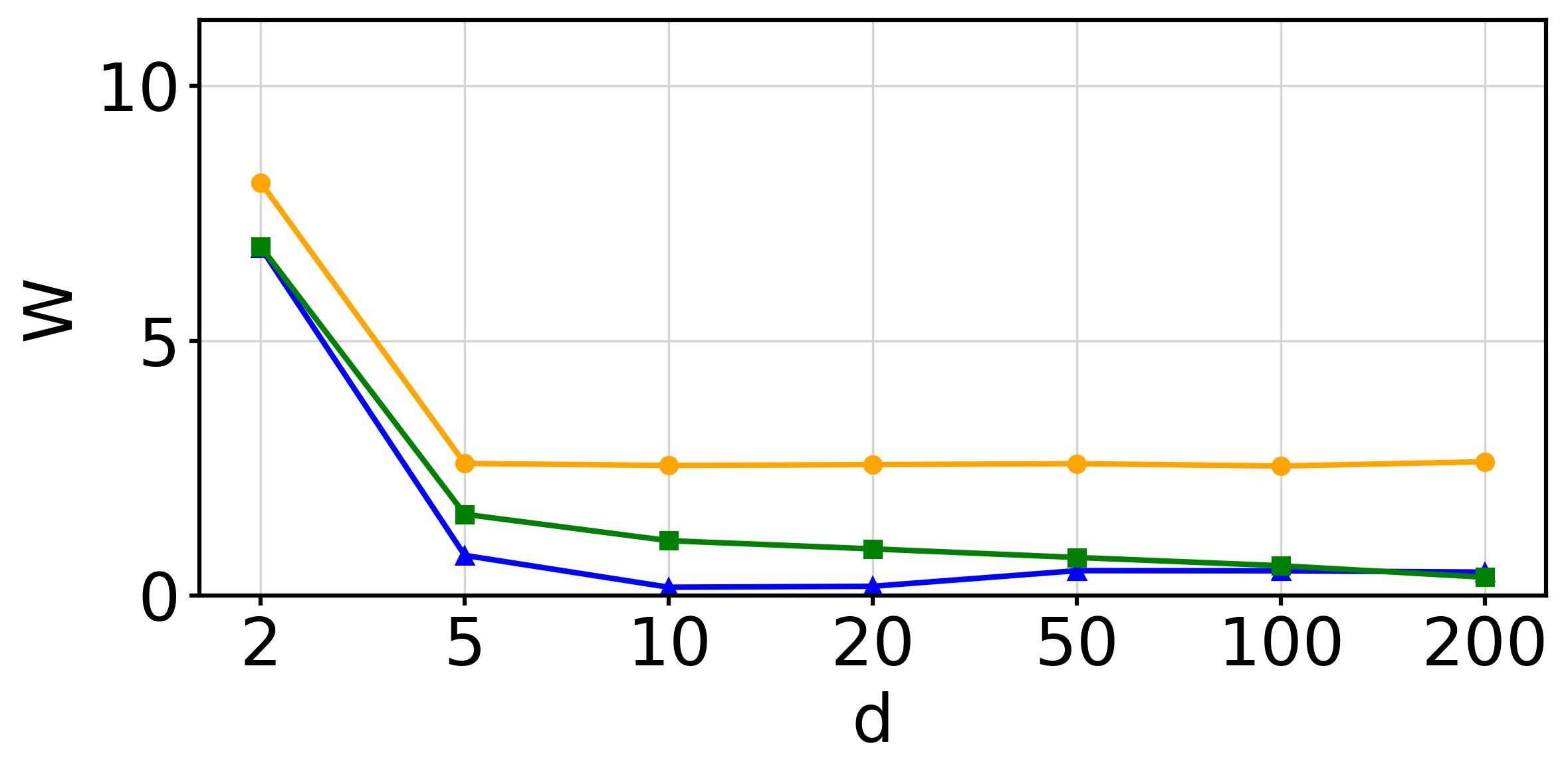} &
    \includegraphics[width=0.26\linewidth]{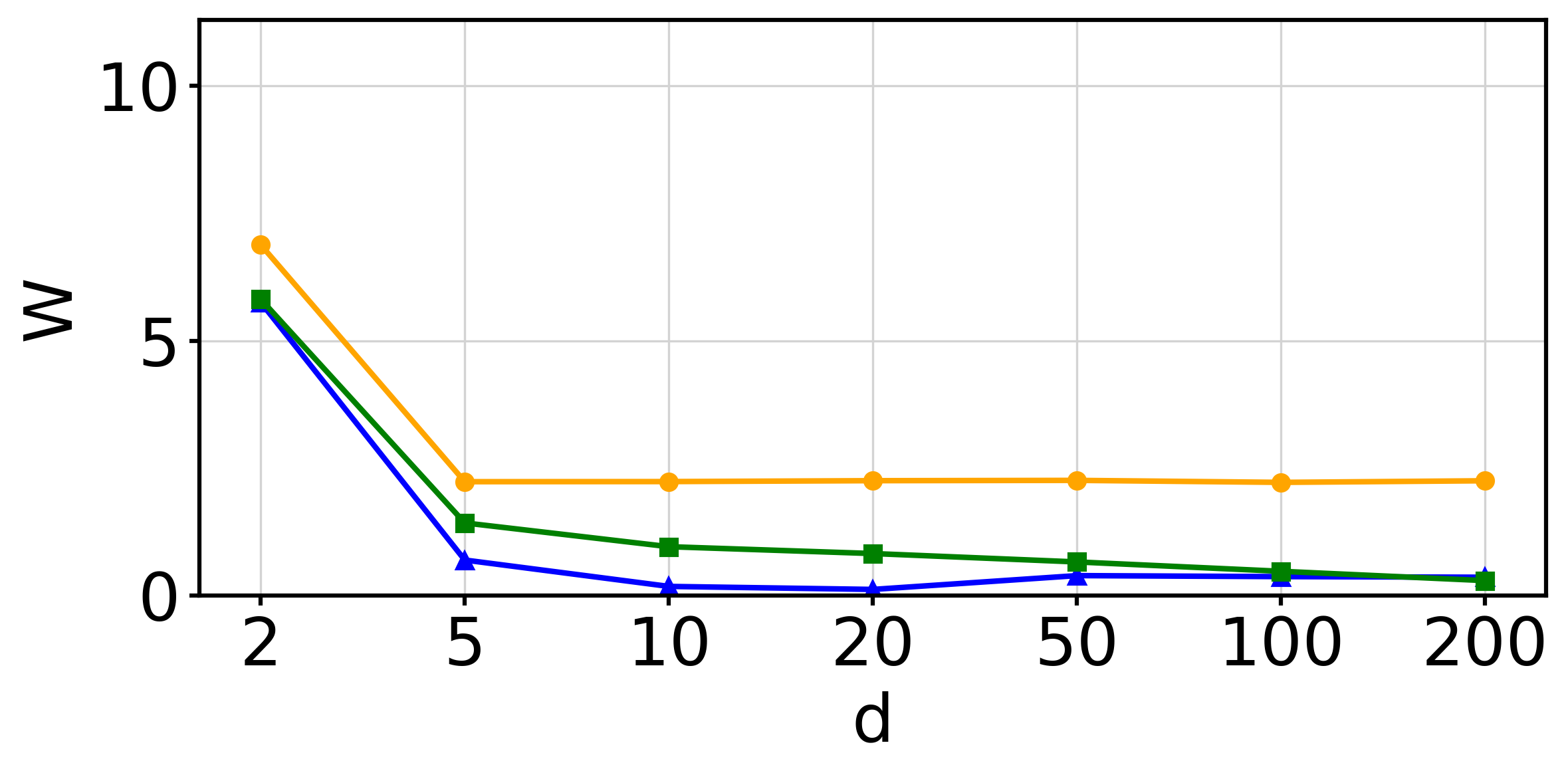} \\
    \midrule
    \raisebox{3.5\height}{Breast cancer} & 
    \includegraphics[width=0.26\linewidth]{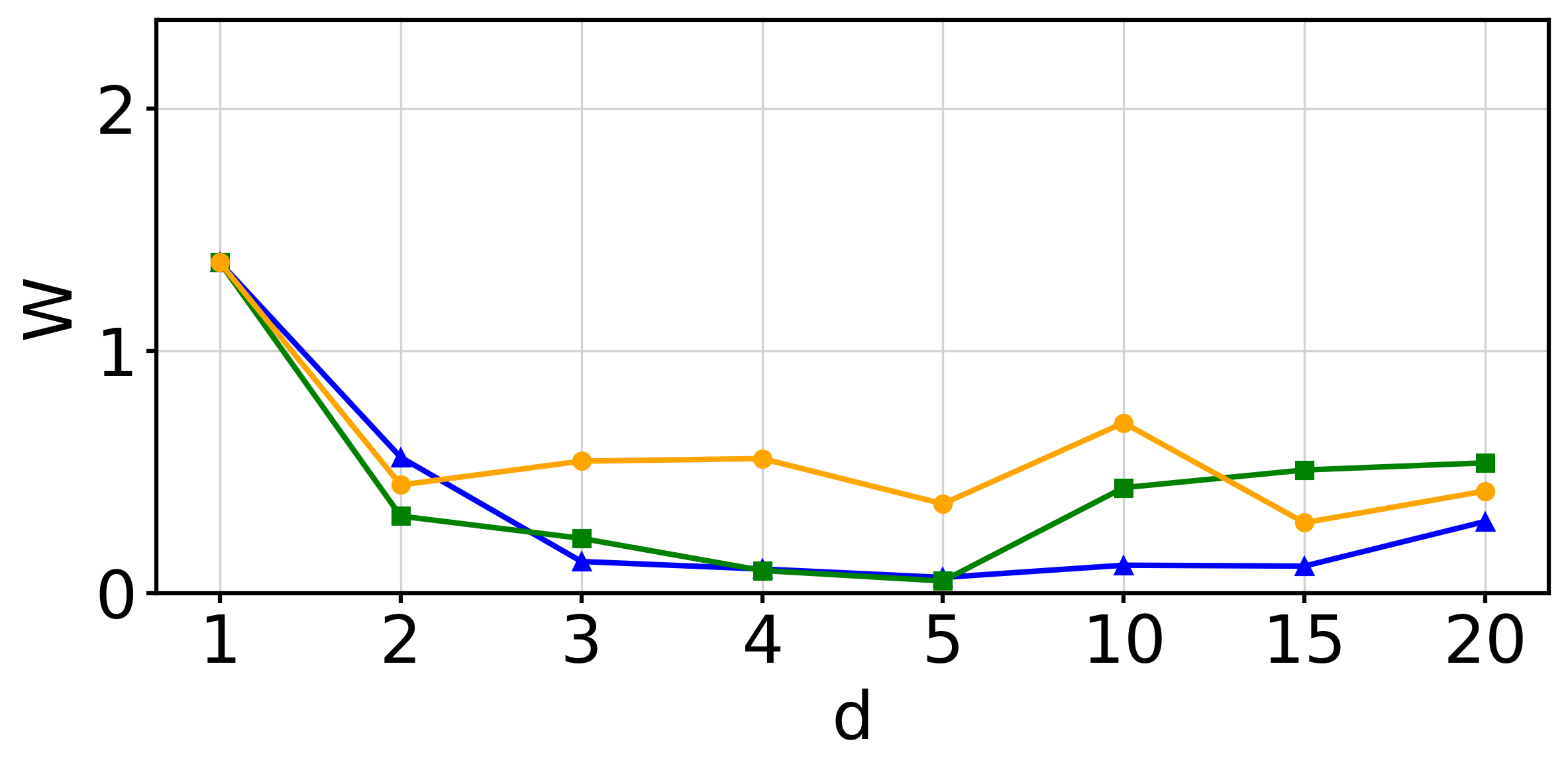} &
    \includegraphics[width=0.26\linewidth]{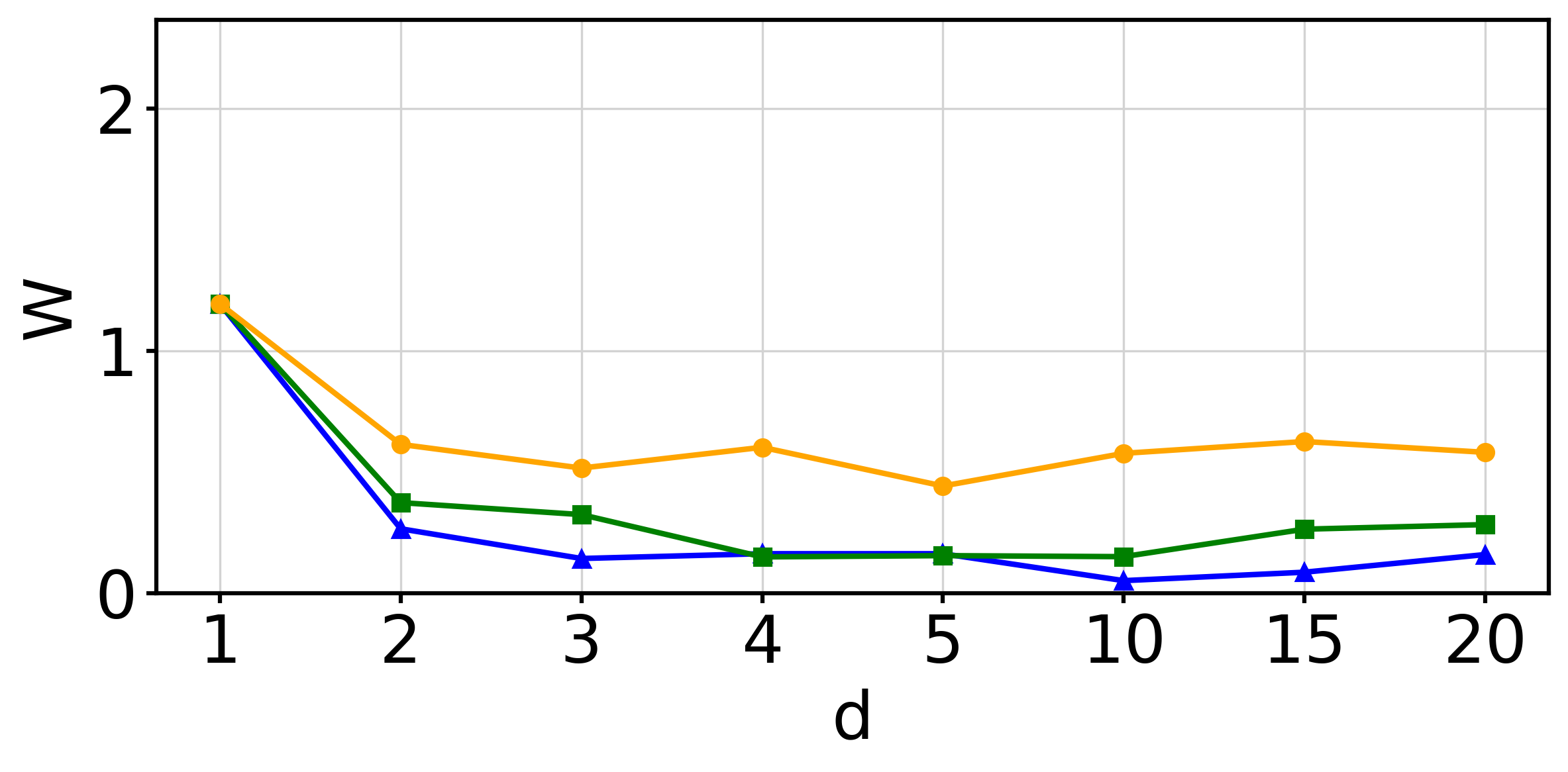} &
    \includegraphics[width=0.26\linewidth]{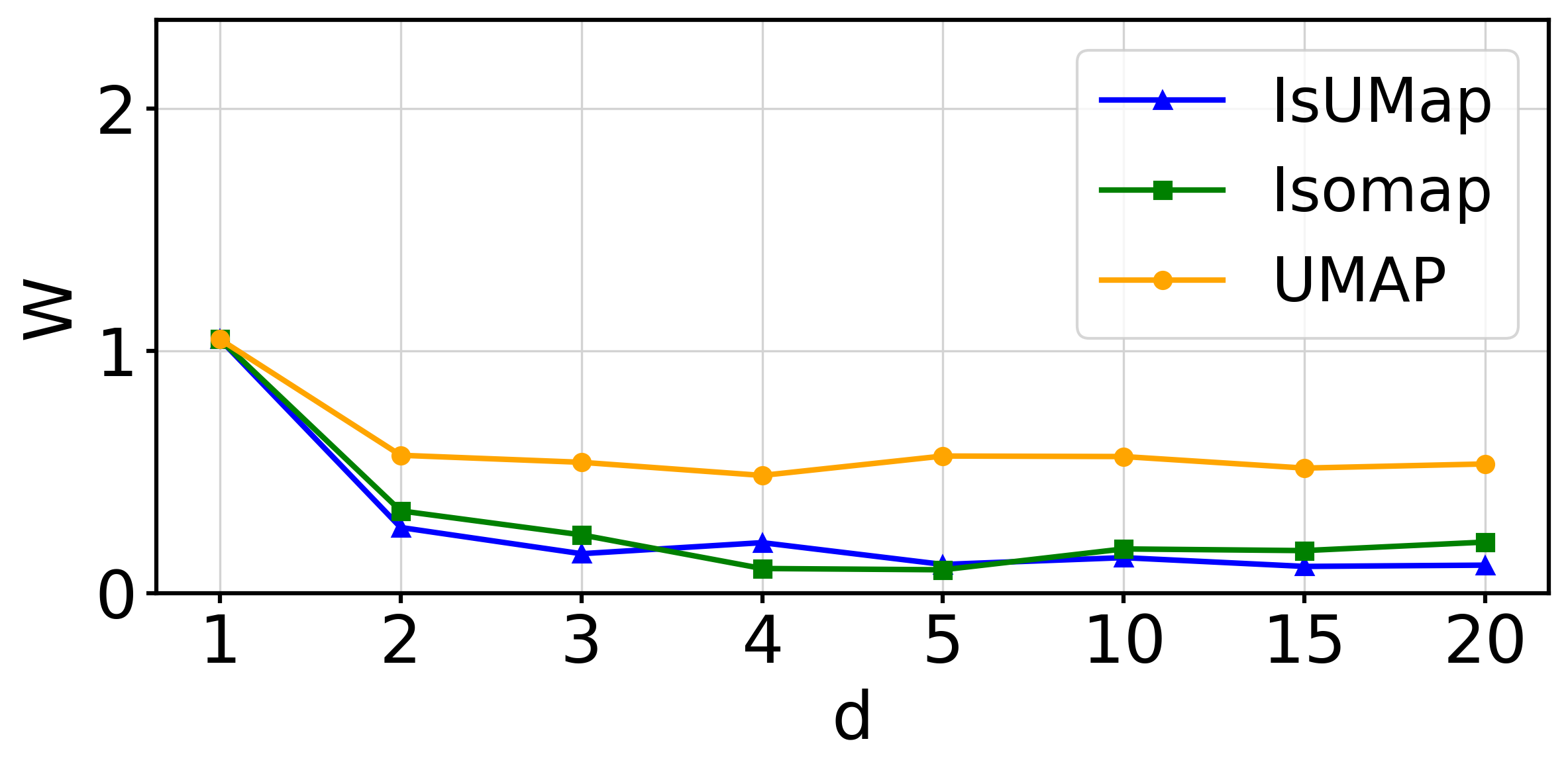} \\
    \midrule
    \raisebox{3.5\height}{Mammoth} & 
    \includegraphics[width=0.26\linewidth]{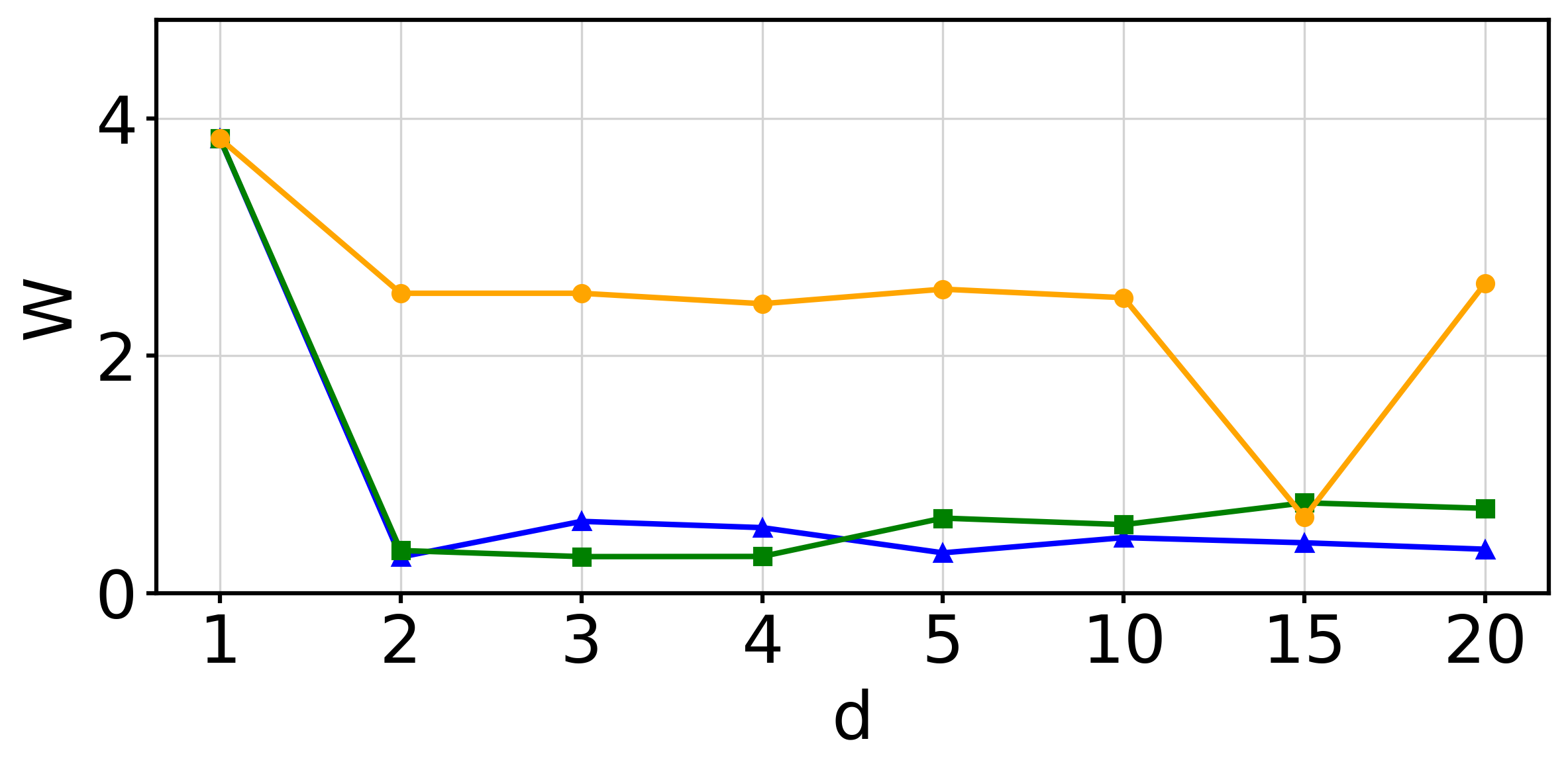} &
    \includegraphics[width=0.26\linewidth]{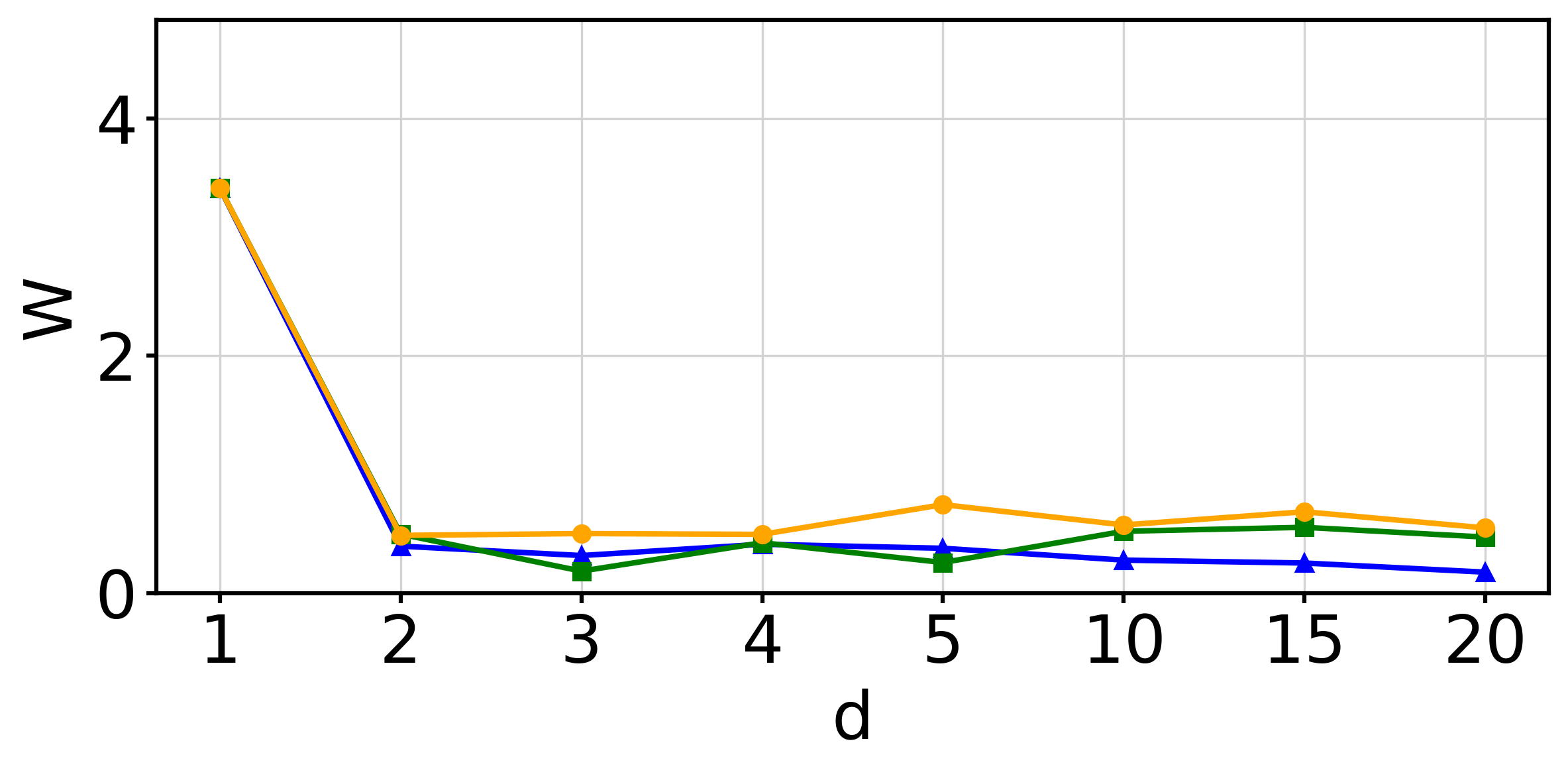} &
    \includegraphics[width=0.26\linewidth]{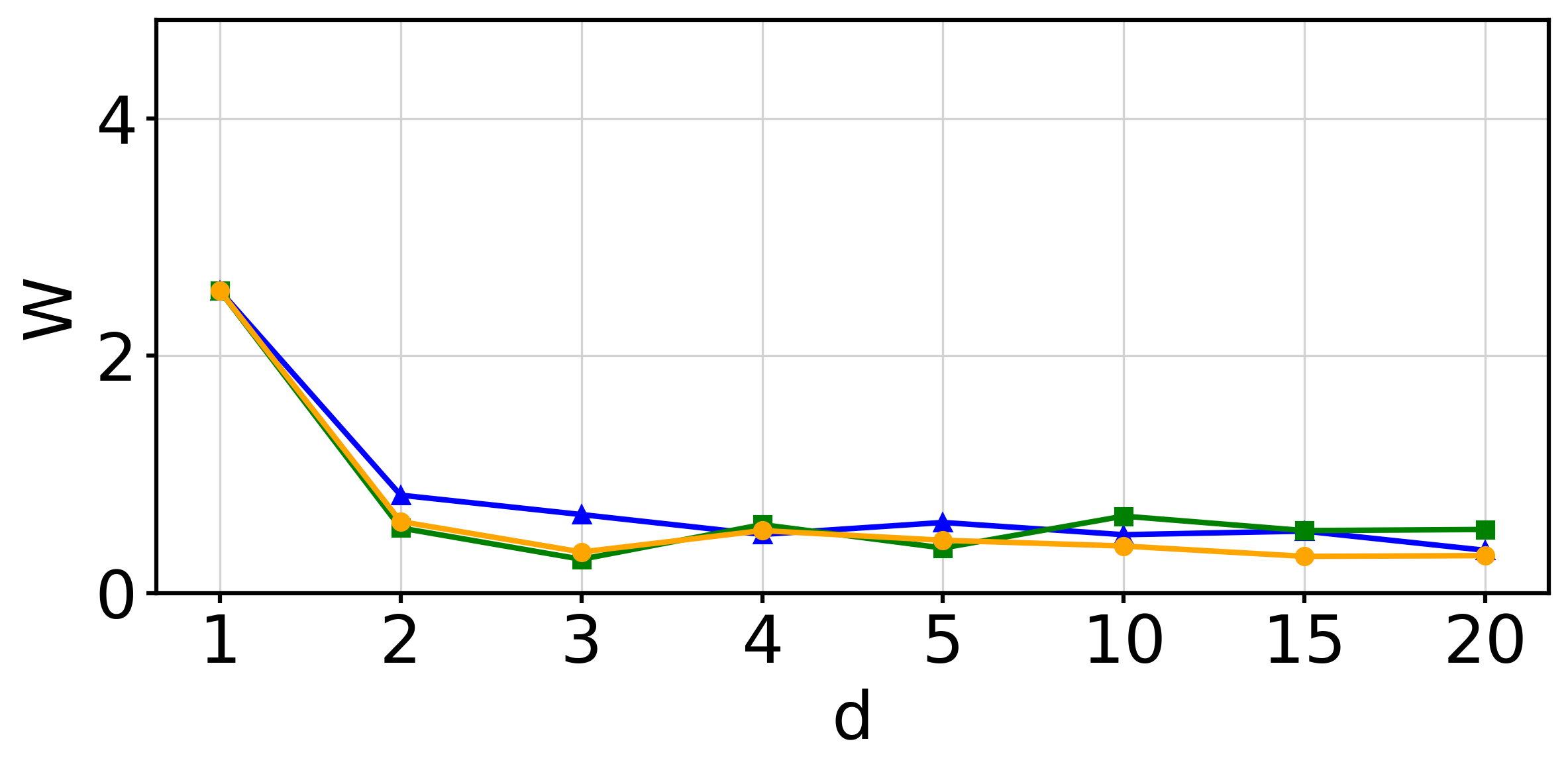} \\
    \bottomrule
  \end{tabular}
\end{table}
 The $W_1$ plots in \cref{tab:Wbench} indicate that Isomap and IsUMap outperform UMAP in preserving the intrinsic geometry, as captured by the curvature profile. \\
 We further apply our evaluation framework to three high-dimensional experimental datasets,  namely Trefoil-Knotted Protein Chains \cite{Benjamin23}, single-cell RNA sequence (scRNA-seq) \cite{goekeri2023microrna}(biological datasets), and gridcells \cite{Gardner22} (neural data). The sRNA dataset was originally stored in the RDS format, native to the R programming language. Since our analysis pipeline is implemented in Python, we preprocessed the data to ensure compatibility with our framework. The Trefoil-Knotted Protein dataset is of particular interest, as its initial representation is not given by an embedding into a high-dimensional Euclidean space via scalar features, but rather by a distance matrix. This matrix encodes topological similarity, computed by comparing the persistence diagrams and persistence landscapes associated with each datapoint.\\
  The $W_1$ plots applying the three chosen embedding methods are presented in \cref{tab:Wexp}. The results indicate that, for instance on gridcells, IsUmap and Isomap outperform UMAP significantly in preserving the underlying geometry. 

\begin{table}[ht]
  \caption{$W_1$-distance between the $\rho$ distributions before and after DR for experimental datasets}
  \label{tab:Wexp}
  \centering
  \begin{tabular}{ c  c c c}
    \toprule
    \multicolumn{1}{c}{} & \multicolumn{1}{c}{\small{10–15}} & \multicolumn{1}{c}{\small{15–20}} & \multicolumn{1}{c}{\small{20–30}} \\ 
    \midrule
    \raisebox{3.5\height}{\small{Proteins}}
    & \includegraphics[width=0.26\linewidth]{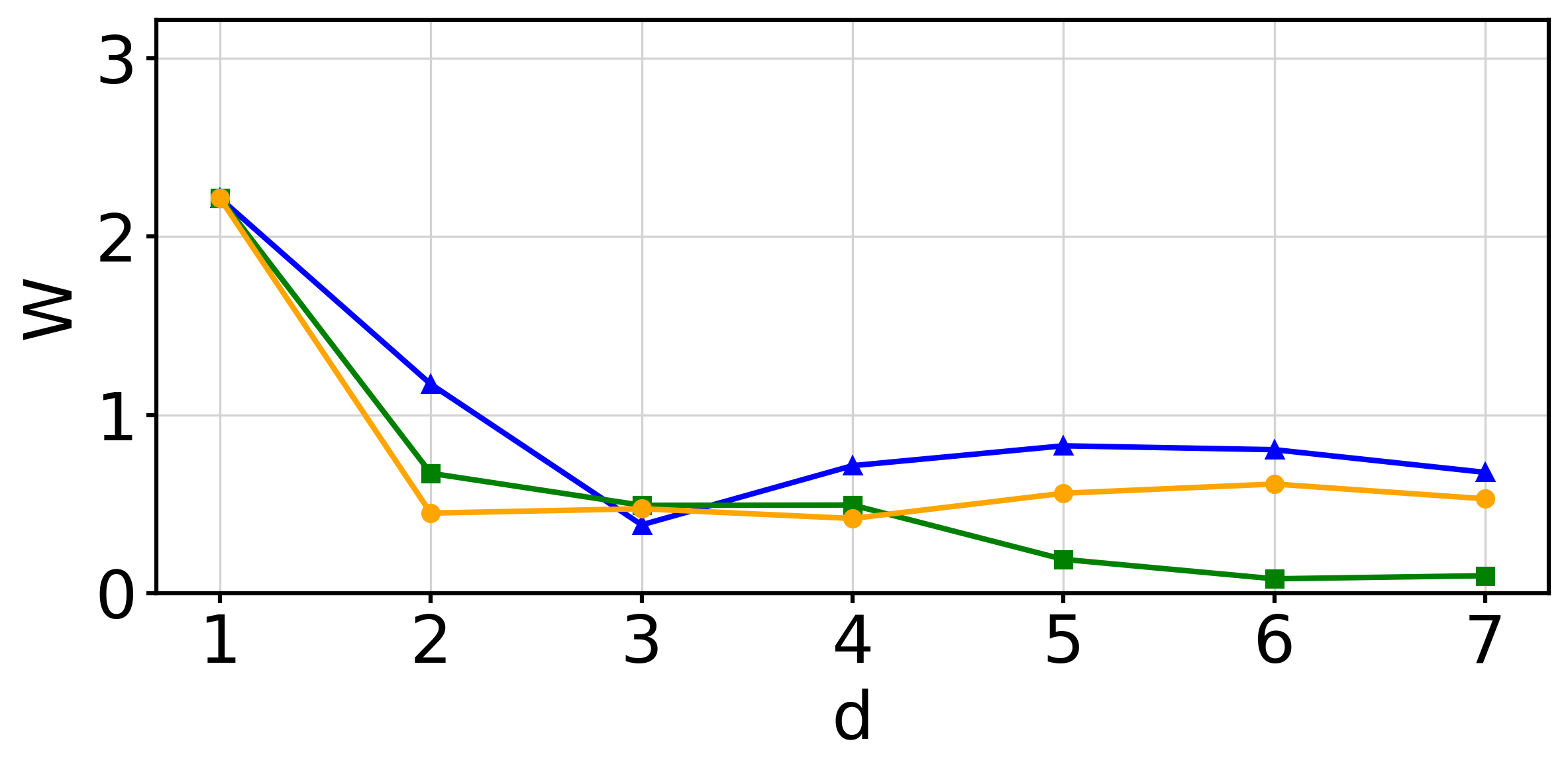}
    & \includegraphics[width=0.26\linewidth]{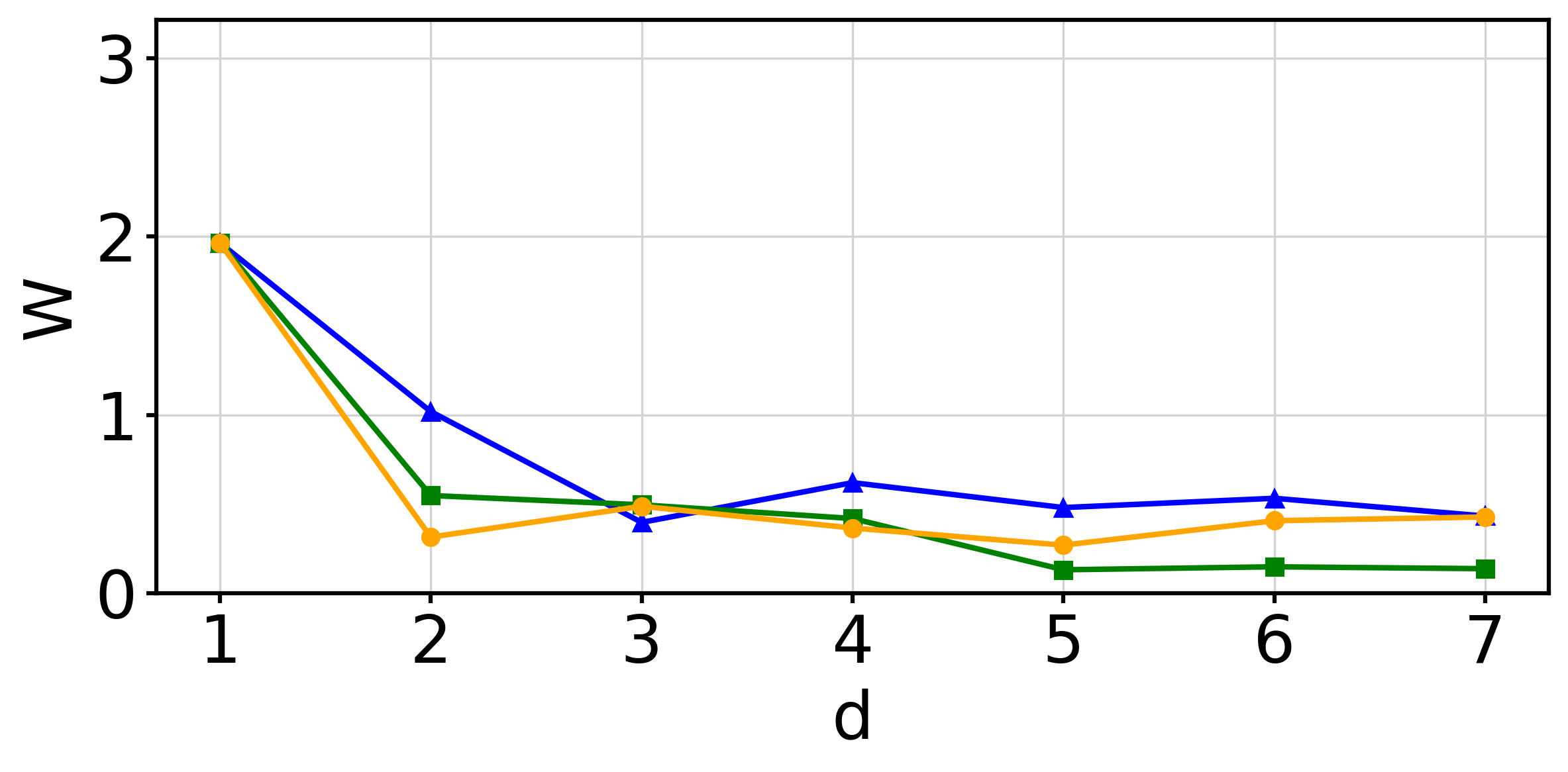}
      &  \includegraphics[width=0.26\linewidth]{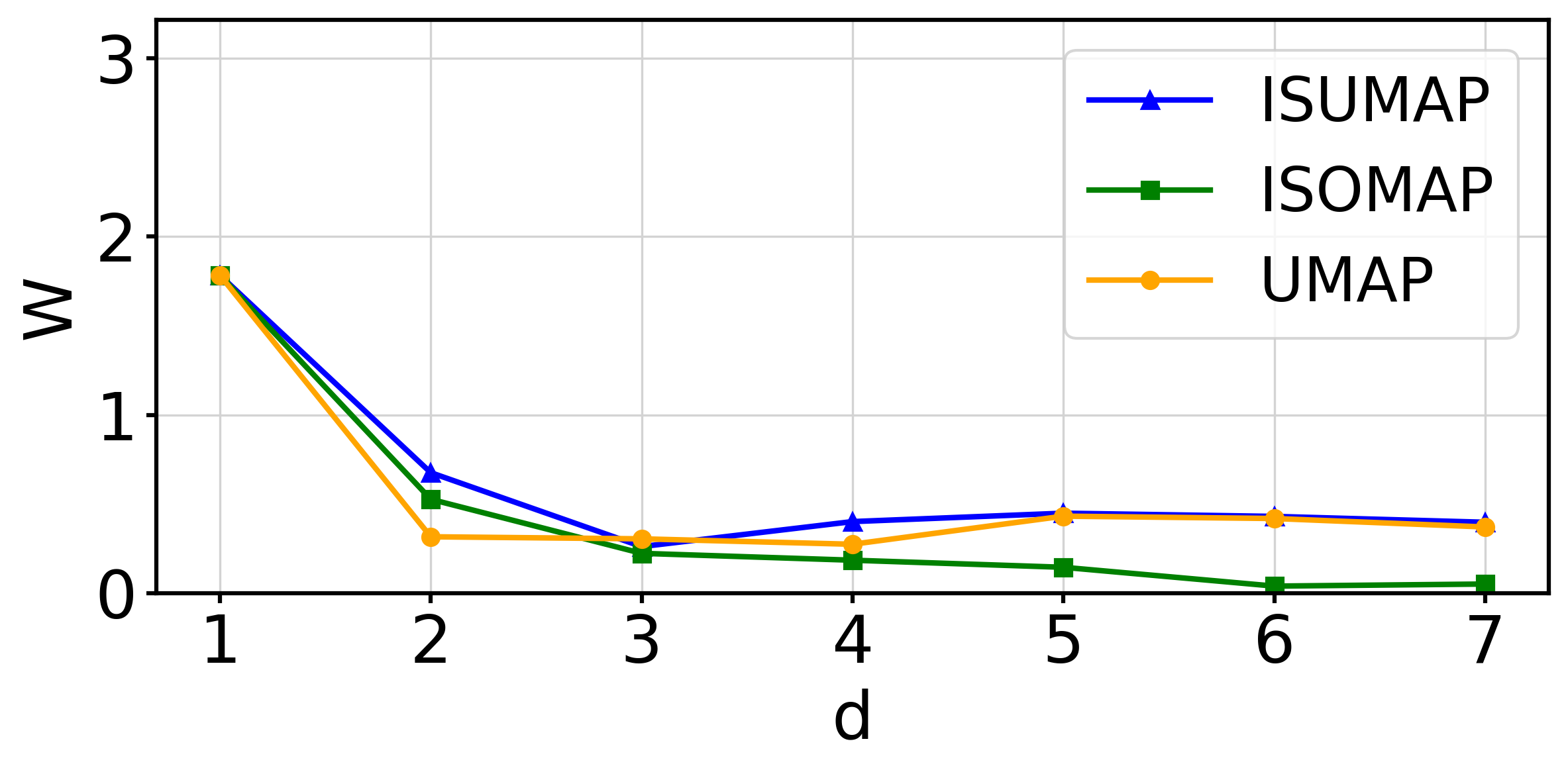}         \\
    \midrule
  \raisebox{3.5\height}{\small{Gridcells}}    
  & \includegraphics[width=0.26\textwidth]{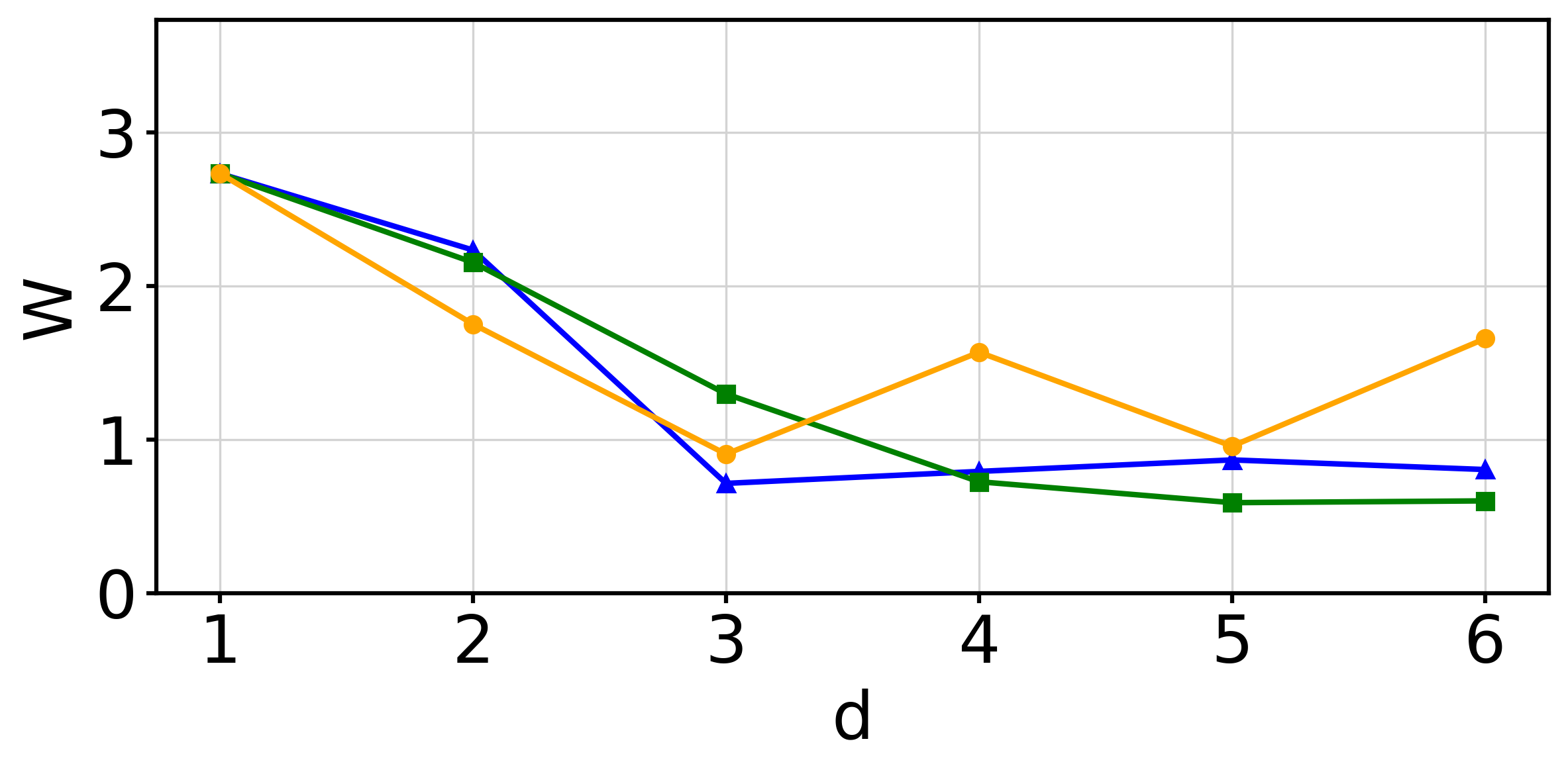} & \includegraphics[width=0.26\textwidth]{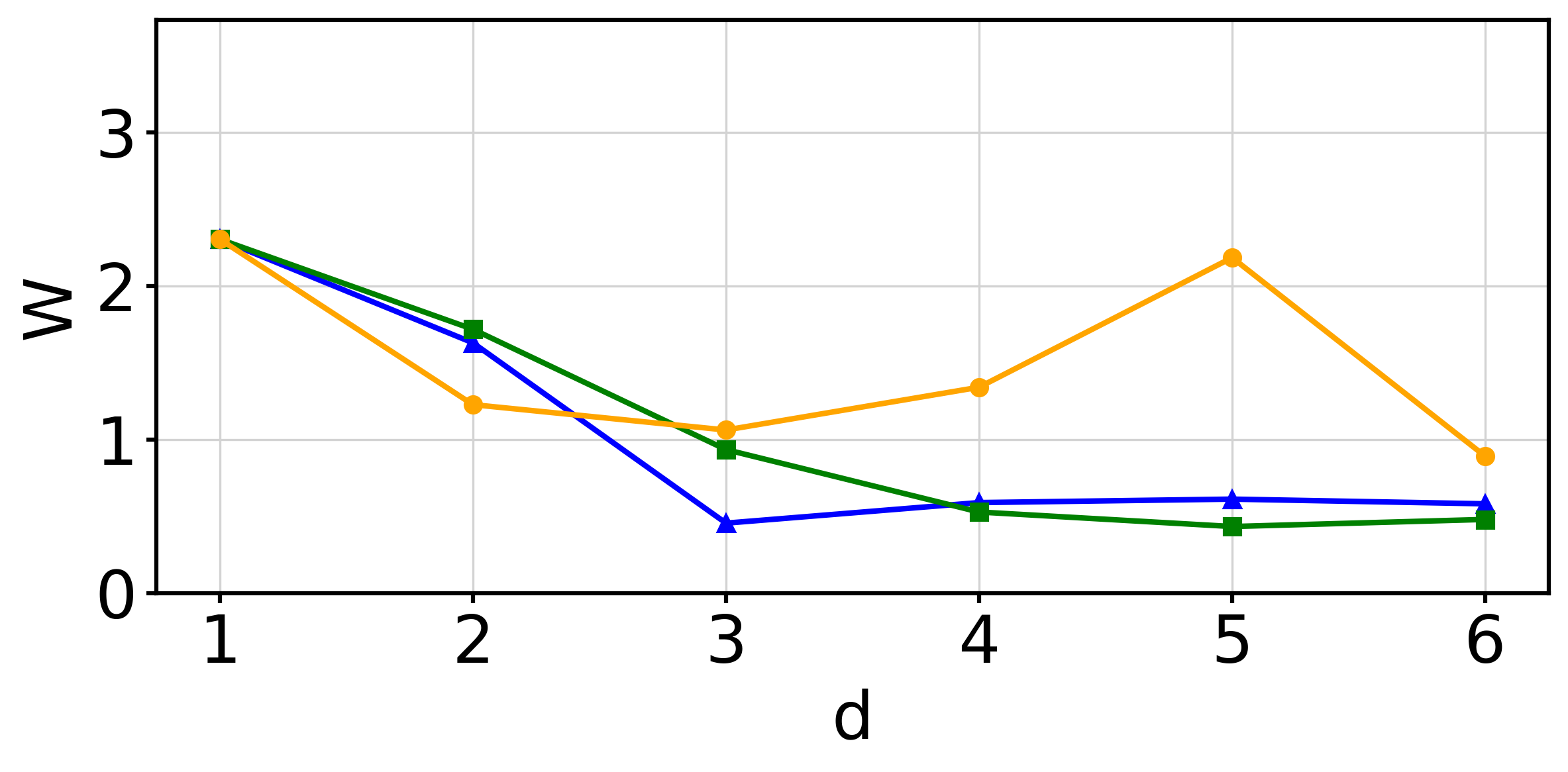} & \includegraphics[width=0.26\textwidth]{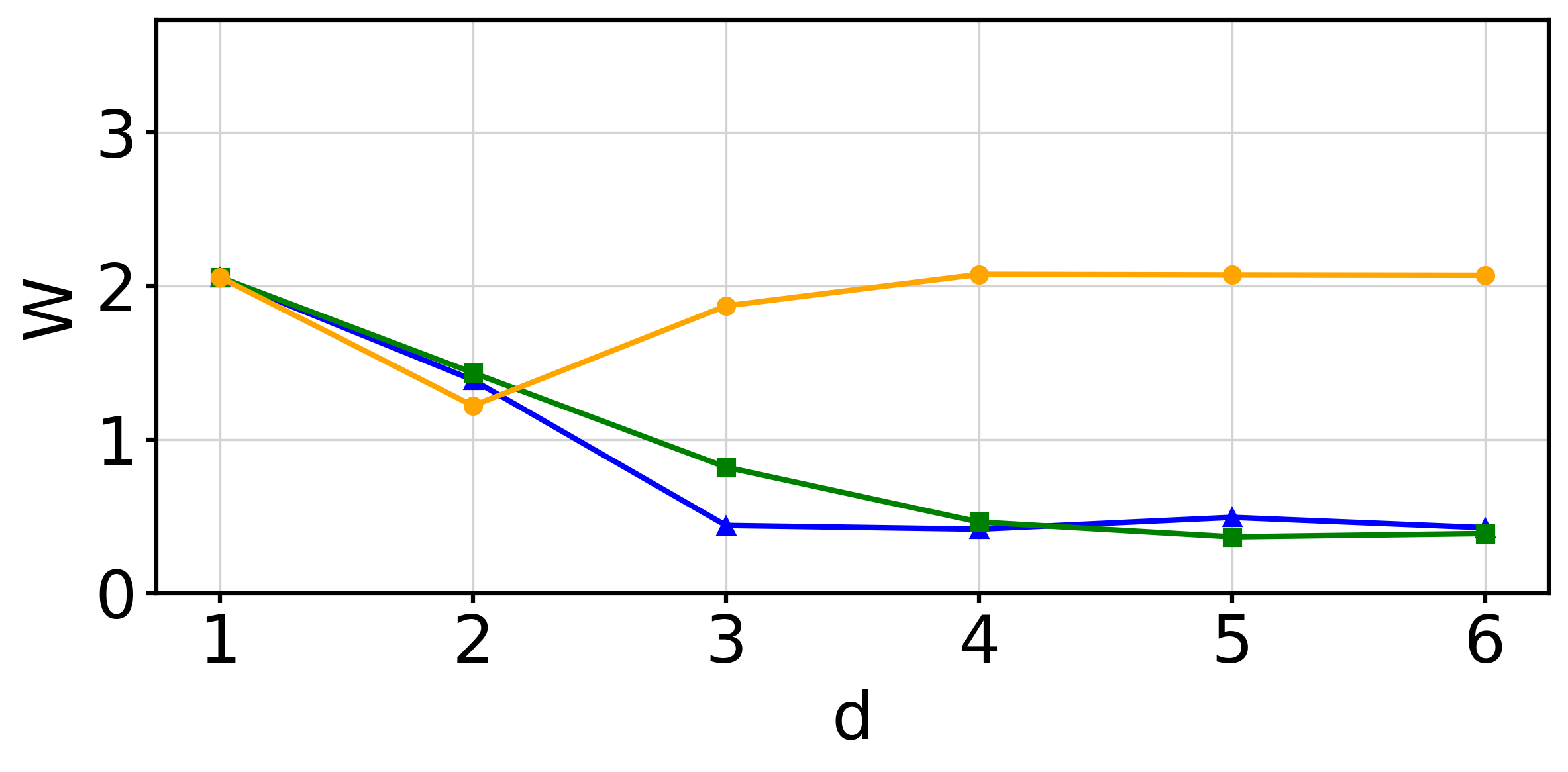}    \\
  \midrule
  \raisebox{3.5\height}{\small{sRNA}}    
  & \includegraphics[width=0.26\textwidth]{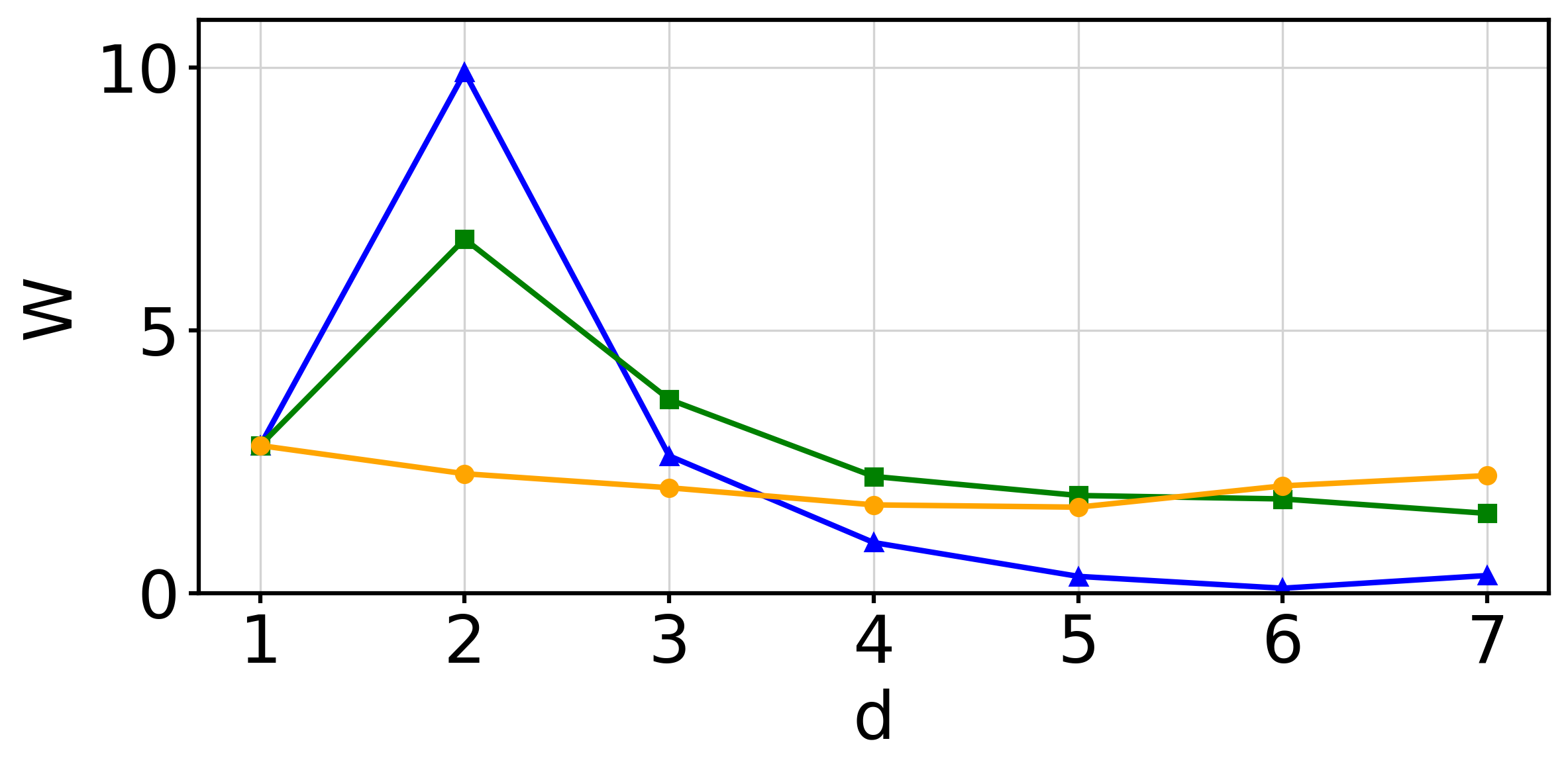} & \includegraphics[width=0.26\textwidth]{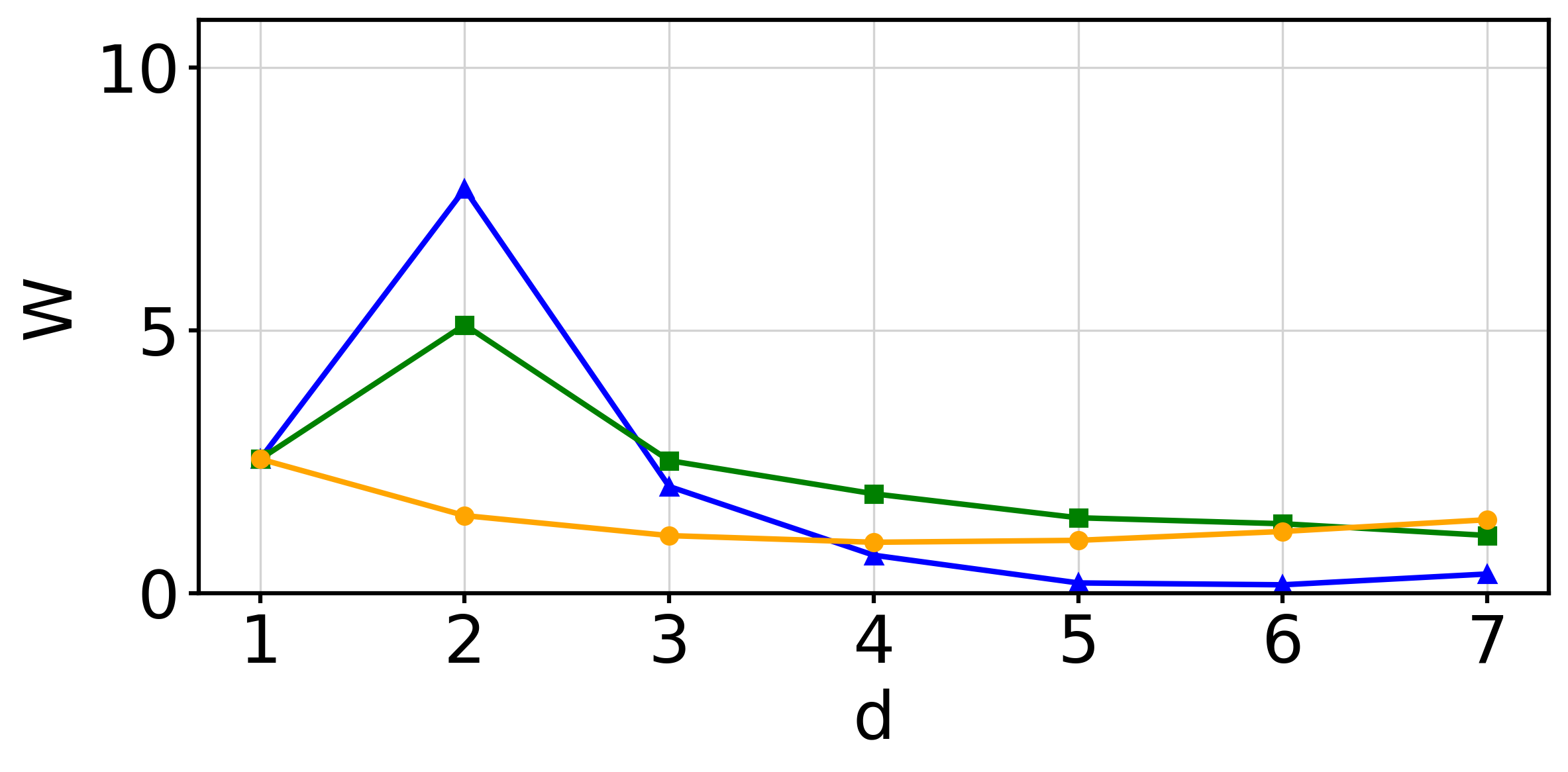} & \includegraphics[width=0.26\textwidth]{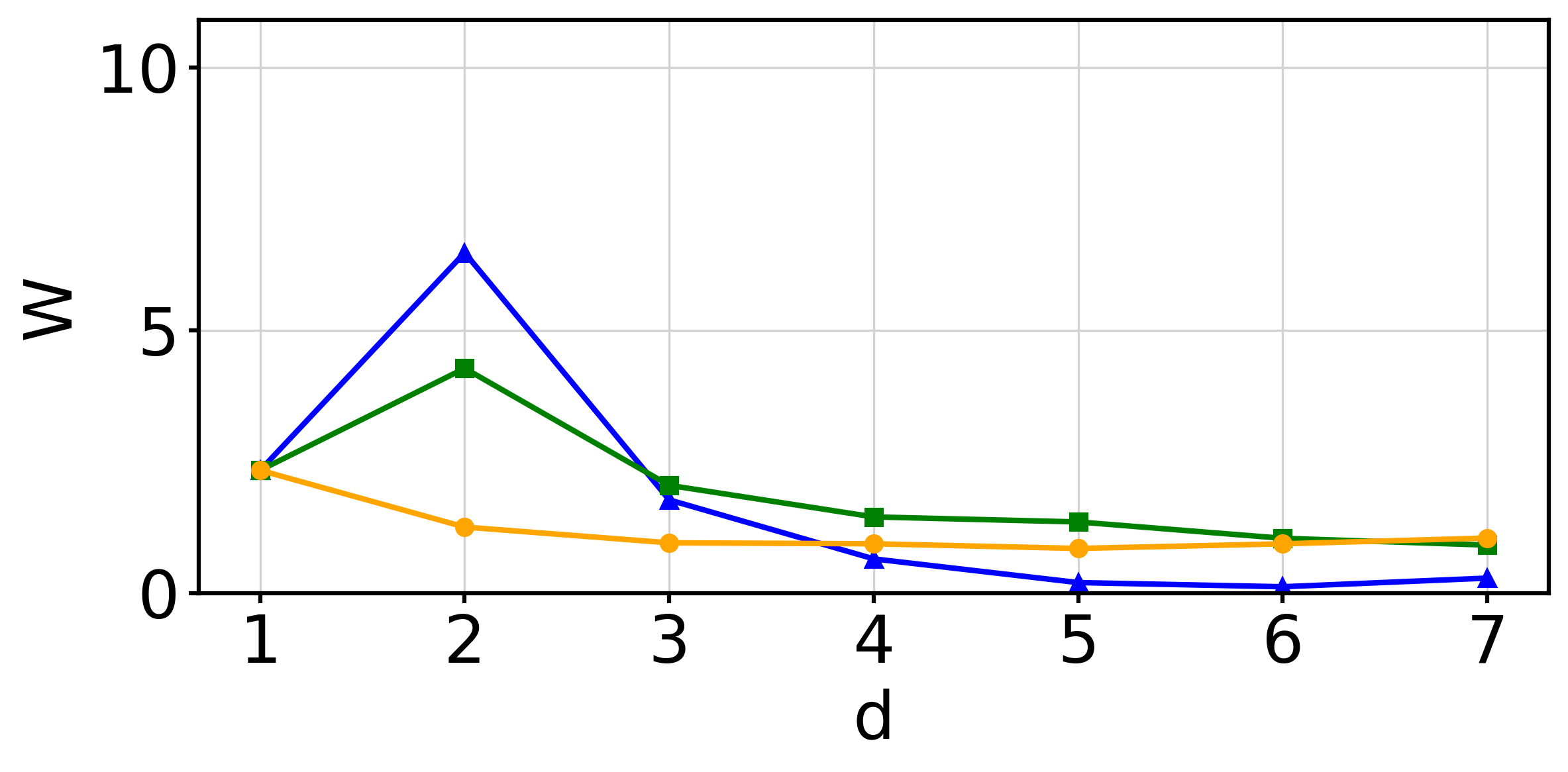}    \\
  \bottomrule
  \end{tabular}
\end{table}

\begin{figure}[ht] 
\centering
   { \includegraphics[width=0.32\textwidth]
		 {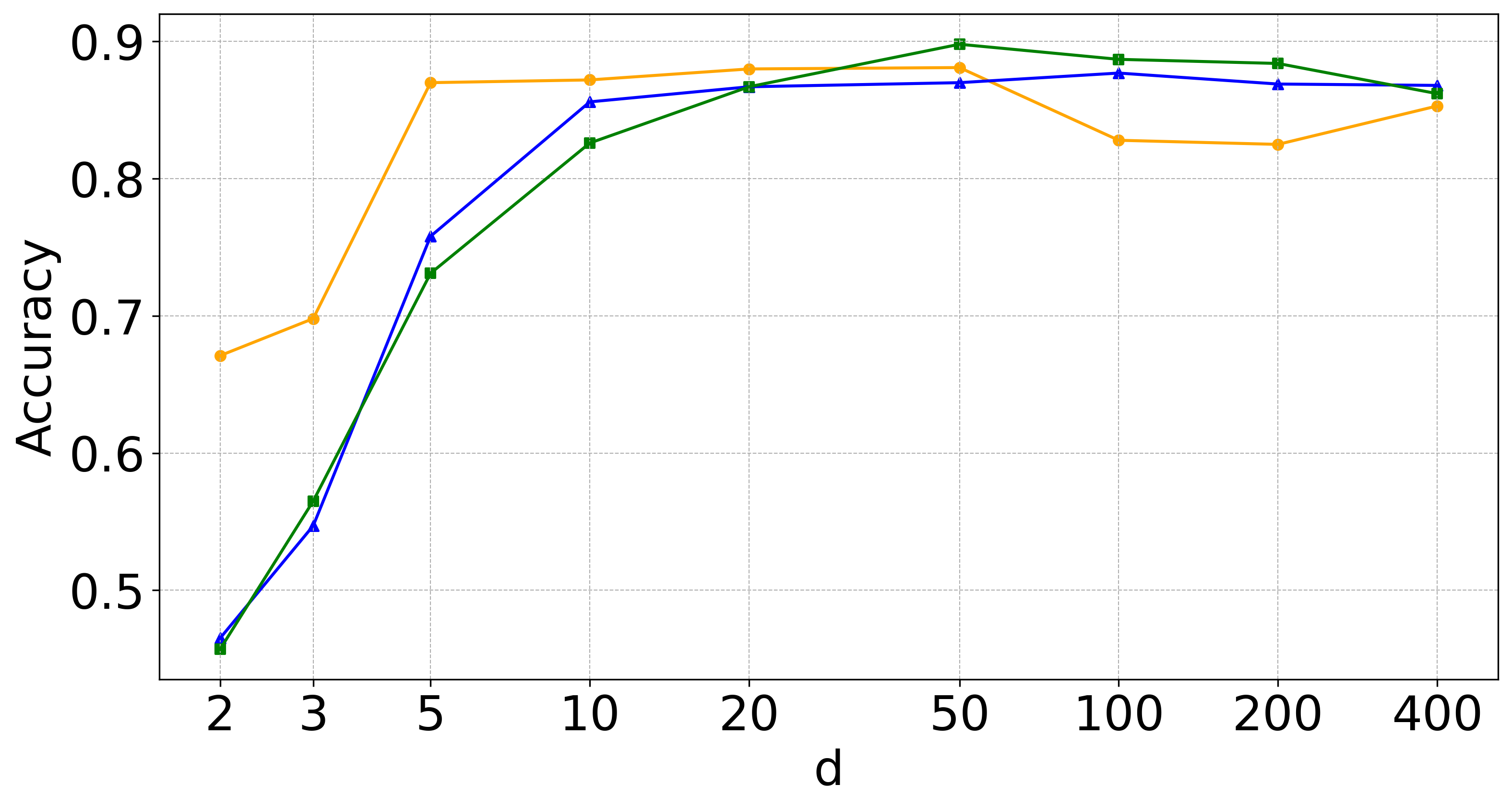}}
    {\includegraphics[width=0.32\textwidth]{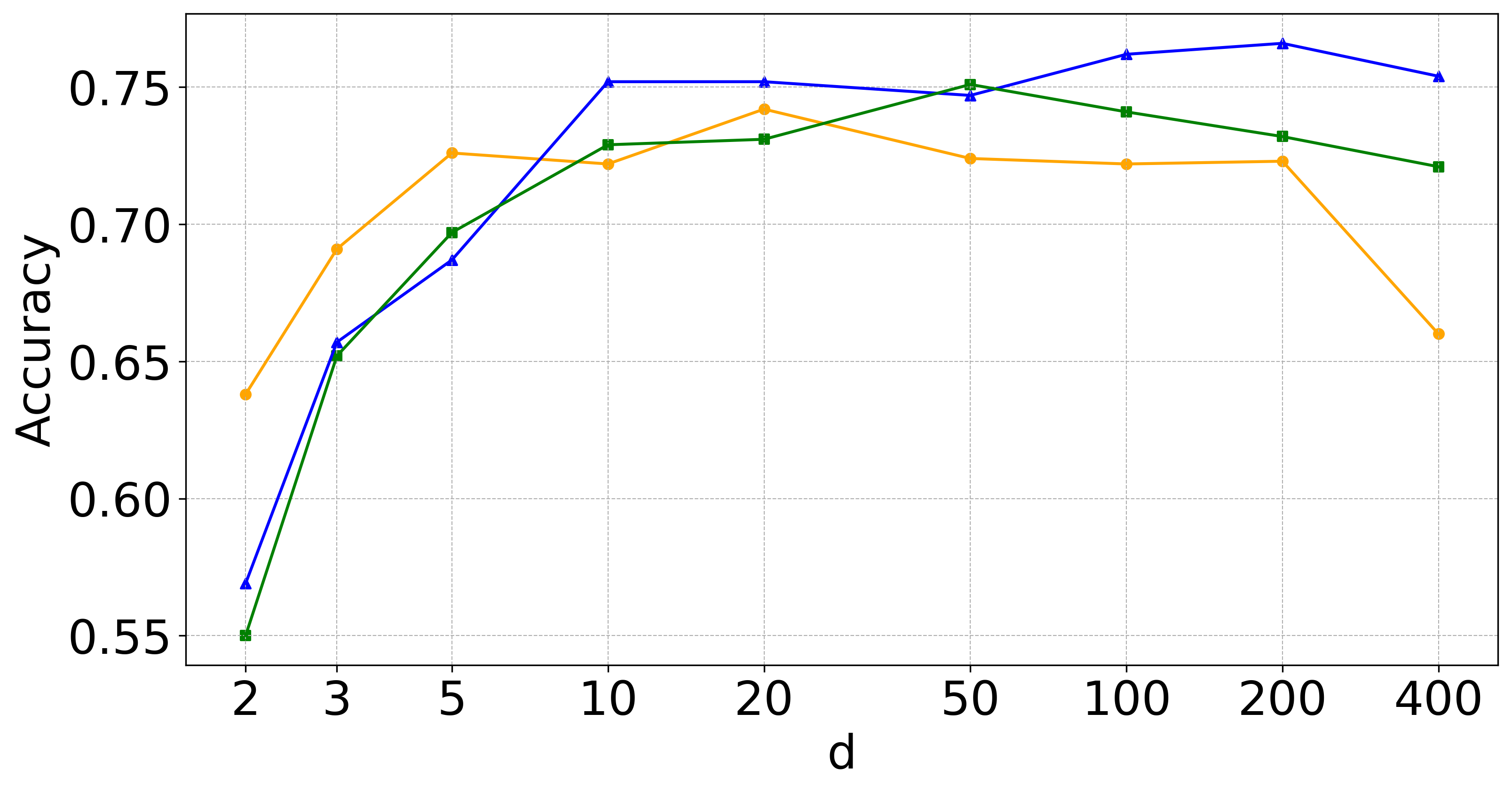}}
    {\includegraphics[width=0.32\textwidth]{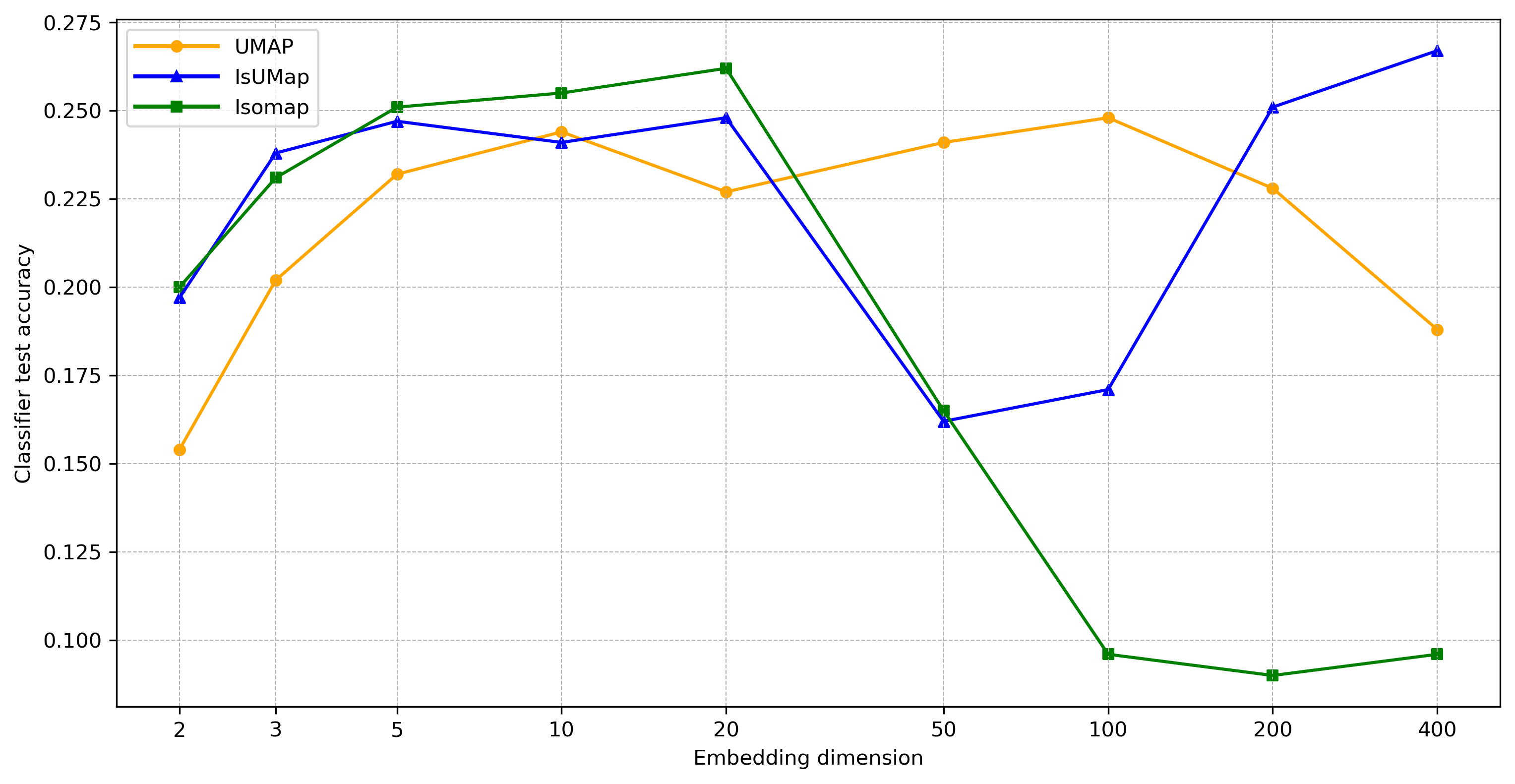}} 
\caption{Test accuracy of a linear classifier against dimension from left to right on MNIST, FashionMNIST, and CIFAR10. }
    \label{fig:classifier}
    \end{figure}

Since this geometric comparison offers a powerful framework for evaluating and comparing different dimensionality reduction or embedding methods, it can be used alongside other (task dependent) evaluation methods to enhance overall performance.  
For instance, one could  evaluate IsUMap, UMAP, Isomap methods as the preprocessing step for linear classification on datasets, as we here do on MNIST, CIFAR-10, and FashionMNIST. More precisely, for each of these methods we first randomly sampled $10000$ datapoints from each dataset, which we then embedded in $d\in\{2,10,50,100,200\}$ dimensional Euclidean spaces using the corresponding method. From the resulting embedding, half of the points were used for training and validation ($80/20$ split) and the remaining as the held-out test set to train a linear classifier with Adam optimizer ($\eta=10^{-3}$), batch size $64$, and for $1000$ epochs. The test accuracy of each method across various dimensions is reported in \cref{fig:classifier}.
 
While UMAP achieves slightly better test accuracy than other methods on the MNIST dataset when the dimension is reduced to $10$ or even as low as $5$, its performance in preserving the intrinsic geometry is less impressive, particularly when compared to Isomap. For MNIST, a combined evaluation using both the curvature profile and test accuracy suggests that up to dimension $10$, all three methods perform similarly in terms of classification accuracy. However, Isomap and IsUMap offer superior geometric fidelity, making them preferable when preserving the underlying structure is a priority. Similarly, one can compare the performance on FMNIST and CIFAR10.
\subsection{Estimating Suitable Dimensions for Embedding}\label{intdim}  
\begin{figure}[ht]
\centering
    { \includegraphics[width=0.32\textwidth]{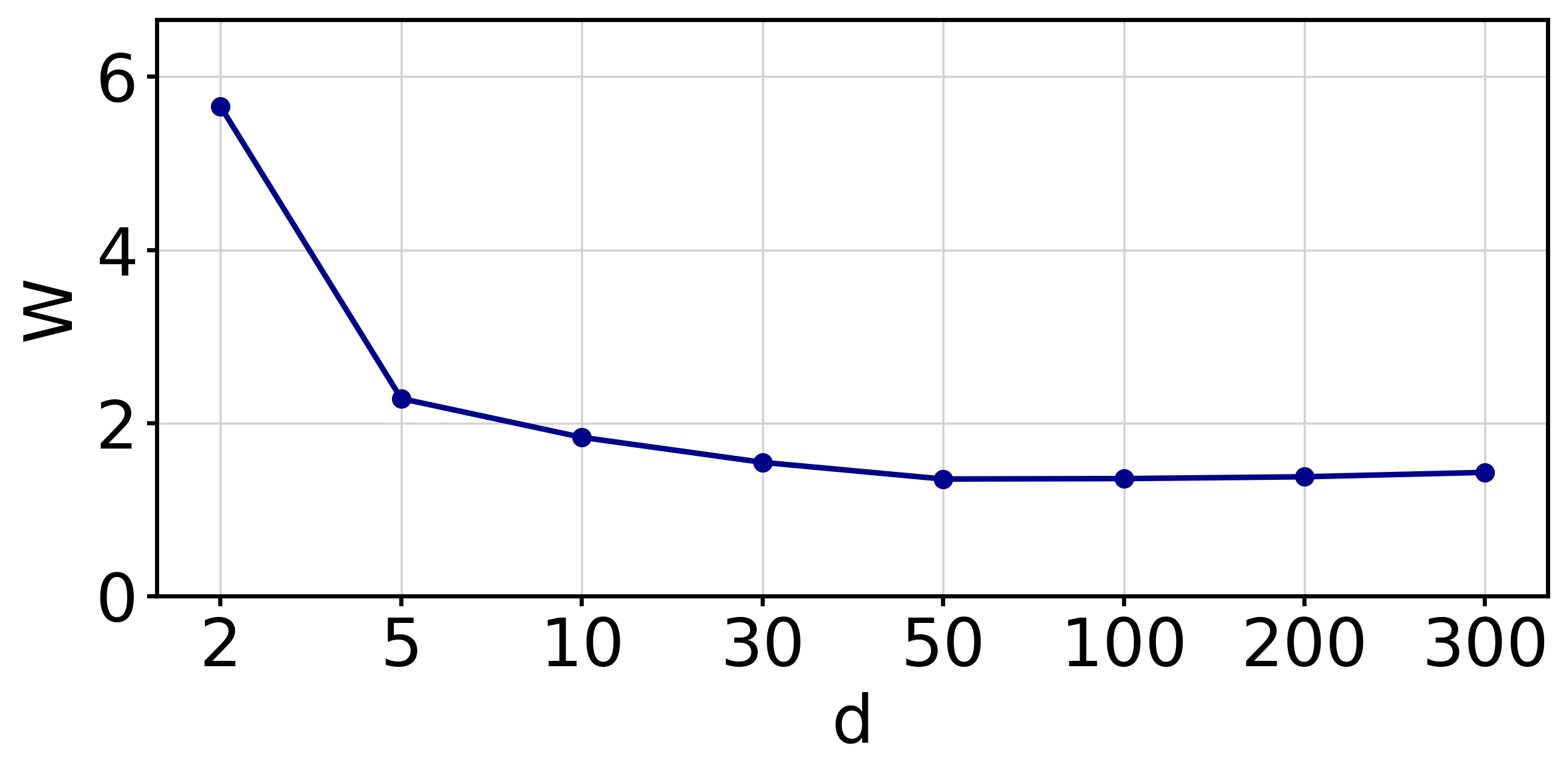}}
    {\includegraphics[width=0.32\textwidth]{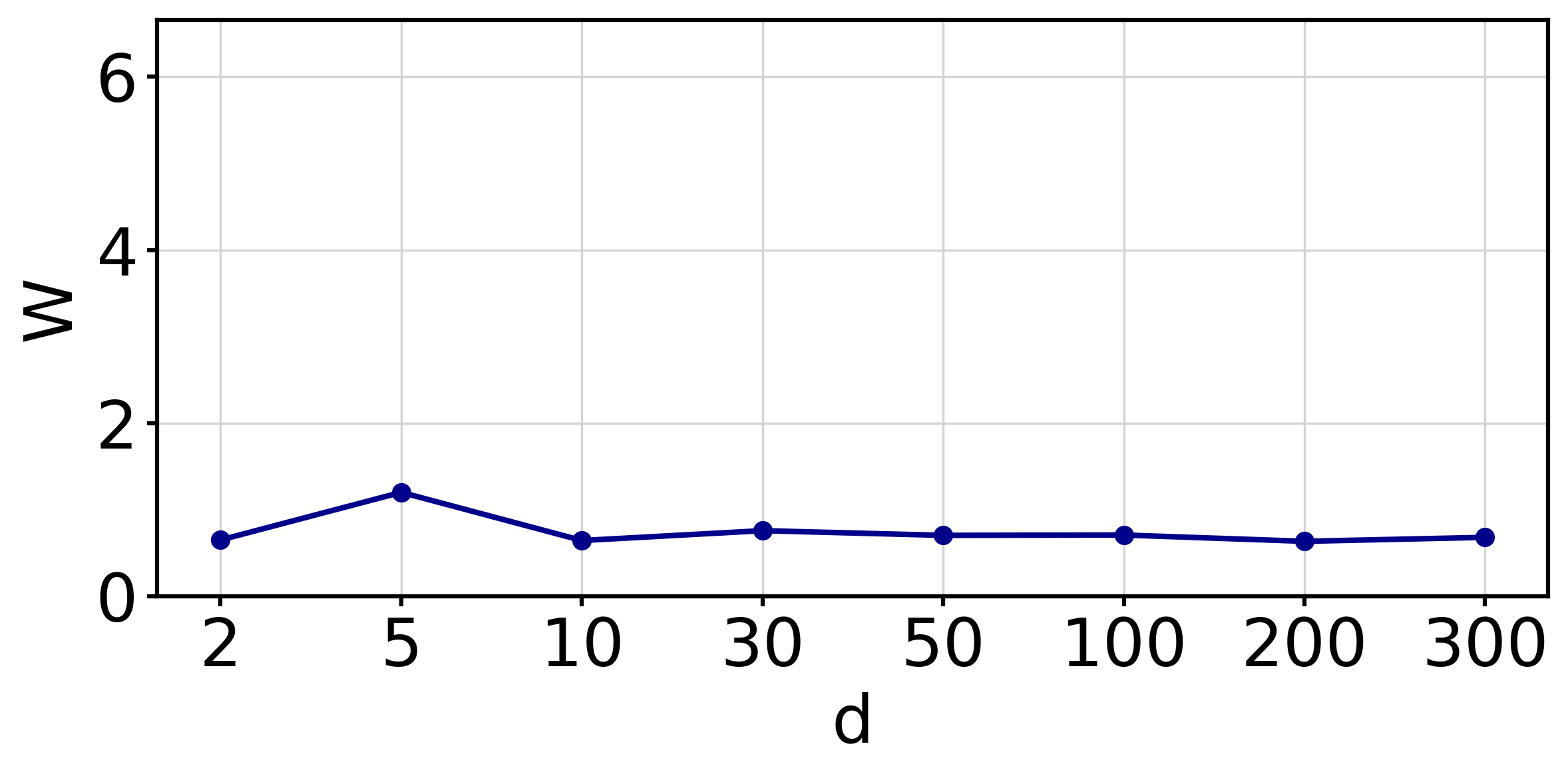}}
    {\includegraphics[width=0.32\textwidth]{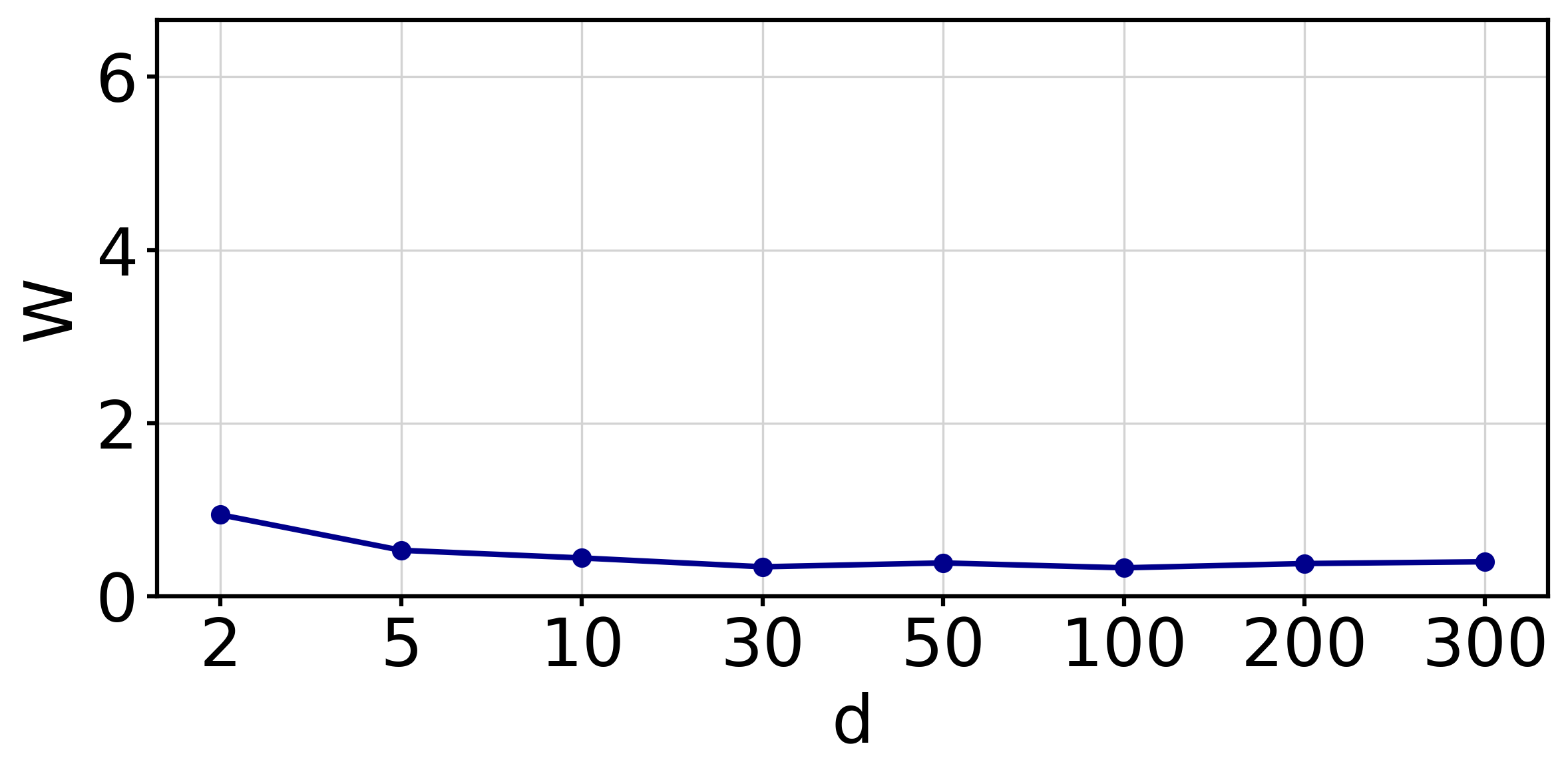}} 
\caption{$W_1$-distance between the $\rho$ distributions between the initial tree graph and its MDS embedding against dimension from left to right for $(k_{\min},k_{\max})= (5,10), (15,20)$ and $(20,30)$. }
    \label{fig:mdstree}
    \end{figure}

The results of comparing the geometry of the initial data representation and its embeddings across different dimensions, as discussed in Section \ref{sec:dimreduction}, suggest that the curvature profile can serve not only an evaluation tool for dimensionality reduction methods, but also as a criterion to determine the appropriate embedding dimension for each method. For example, \cref{tab:Wbench}, together with \cref{fig:classifier}, suggests that an embedding dimension of $10$ for MNIST is optimal (not surprising, as it should distinguish 10 digits), striking a balance between preserving geometric structure and providing a suitable representation for downstream classification tasks. \\
To investigate this observation further, we turn to additional examples.  The first examples is the graph tree from \cref{tab:toyex},  originally  equipped with a distance function (graph distance). Its curvature profile, as reported in \cref{tab:toyex}, emphasizes the structural rigidity characteristic of this dataset. Here, we embed this tree into Euclidean spaces of varying dimension using classical MDS and compare the resulting geometry with the original via embed this tree into Euclidean spaces of varying dimension using classical MDS and compare the resulting geometry with the original via $W_1$-distance between curvature profiles. The results,   shown in \cref{fig:mdstree}, indicates that for higher values of $(k_{\min},k_{\max})$ (i.e., (15, 20) and (20, 30)), the  MDS embedding more effectively preserves the underlying geometry of the tree. However, it is important to notice that as we increase the parameter  $(k_{\min}$ and $k_{\max})$, the analysis shifts toward capturing extrinsic geometry, thereby relying less on short-range distances that more accurately reflect the intrinsic structure of the data.

The next example involves a structure with inherent tree-like characteristics: the artificial tree from \cite{Moon19}, originally constructed in a high-dimensional Euclidean space. Specifically, we employed the gen-dla function, which generates synthetic data with a diffusion-limited aggregation (DLA) tree structure.  
The construction is as follows. In order to generate a tree in dimension $n=km$ with $k$ branches and the length of each branch equal to $l$, one starts with the first branch consisting of $l$ linearly spaced points progressing in the first $m$ dimensions. The second branch remains constant in these $m$ dimensions at the endpoint of the first branch, while the next $m$ dimensions progress linearly. Similarly, the third branch progresses in dimensions $(2m+1) - (3m)$ instead of $(m+1) - (2m)$. The remaining branches follow this pattern with variations in branch length. the tree structure can be visualized as a trunk growing from the root in the first $m$ dimensions and then splitting into branches, each evolving in an $m$ dimensional subspace independent of the others and the trunk. \\
The representation we use from this tree in the pipeline is the coordinates of vertices in the high-dimensional space $\R^n$, $n$ being  a hyperparameter in the generating process as the other parameters, including the number of branches and the number of nodes per branch, are set to $10$ and $300$, respectively. Thus we don't have graph representation as our input, and the graph used for the computation of the curvature profile and consequently the estimation of dimension is the neighborhood graph.   \\
 The curvature profile shows us how well the extrinsic distances in this Euclidean embedding capture the intrinsic geometric properties of the tree for different choices of dimension $n$ in which the tree was generated. In particular, as according to \cref{tab:toyex} the extreme curvature characteristic of a tree is distinctively visible in its profile,  the profile of this embedding in Euclidean space, as presented in \cref{tab:proftree},  reveals how close the embedding is to an isometric one. Afterwards, one can investigate how this geometry will carry out to an embedding in some lower-dimension. In that direction, we also processed the artificial tree in $600$ dimensions through our pipeline, following the same approach used for the benchmark and experimental datasets.  We compute the $W_1$ distance while varying dimension in embedding methods, the corresponding plot presented in \cref{tab:warttree}.  
    
\begin{table}[ht]
  \caption{Curvature profile for artificial trees, with average $\rho$ (top row) non-averaged $\rho$ (bottom row),  with $(k_{\min},k_{\max}) = (10,15)$.}
  \label{tab:proftree}
  \centering
  \begin{tabular}{c c c}
    \toprule
   \multicolumn{1}{c}{\small{n= 200}} & \multicolumn{1}{c}{\small{n=400}} & \multicolumn{1}{c}{\small{n=600}} 
		\\
        \midrule
     \includegraphics[width=0.3\textwidth]{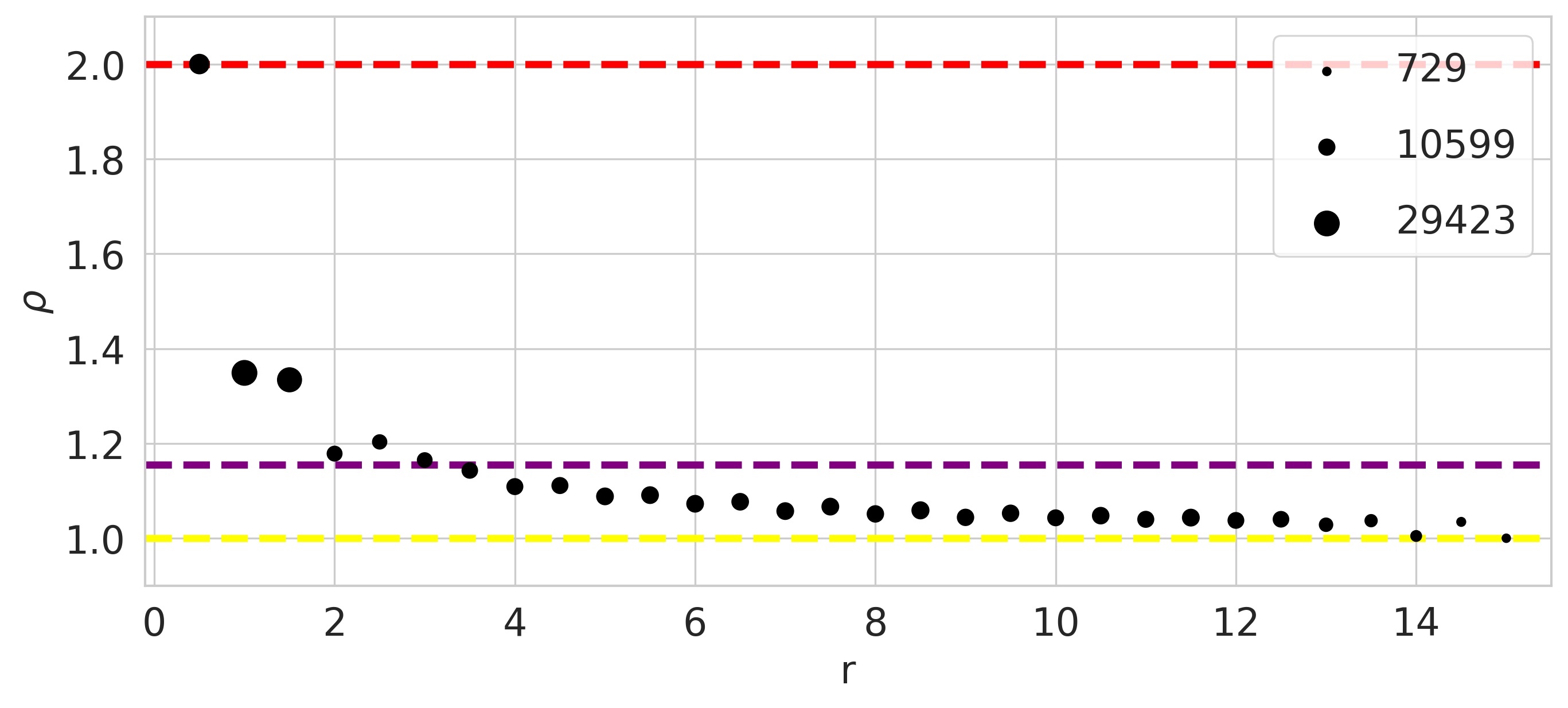} 
  & \includegraphics[width=0.3\textwidth]{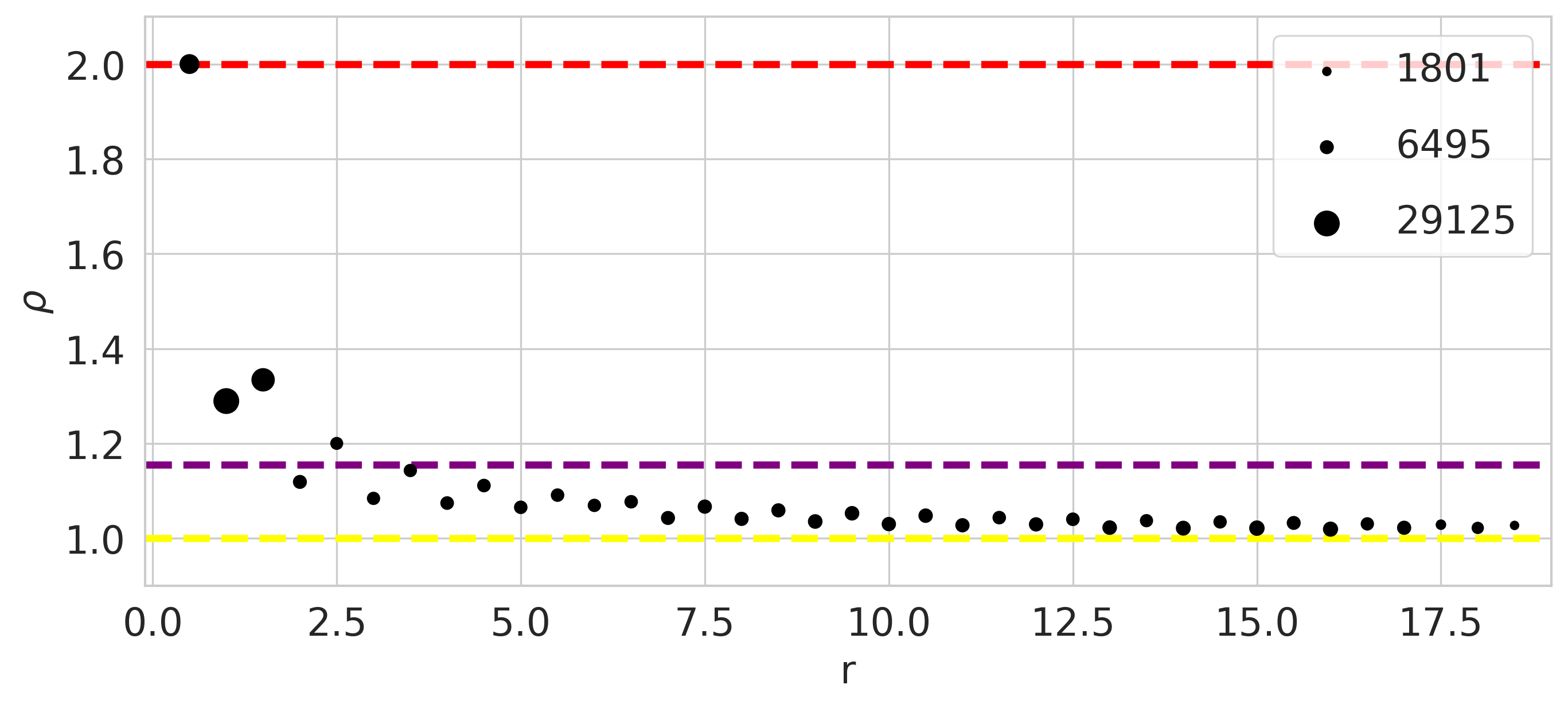} 
  & \includegraphics[width=0.3\textwidth]{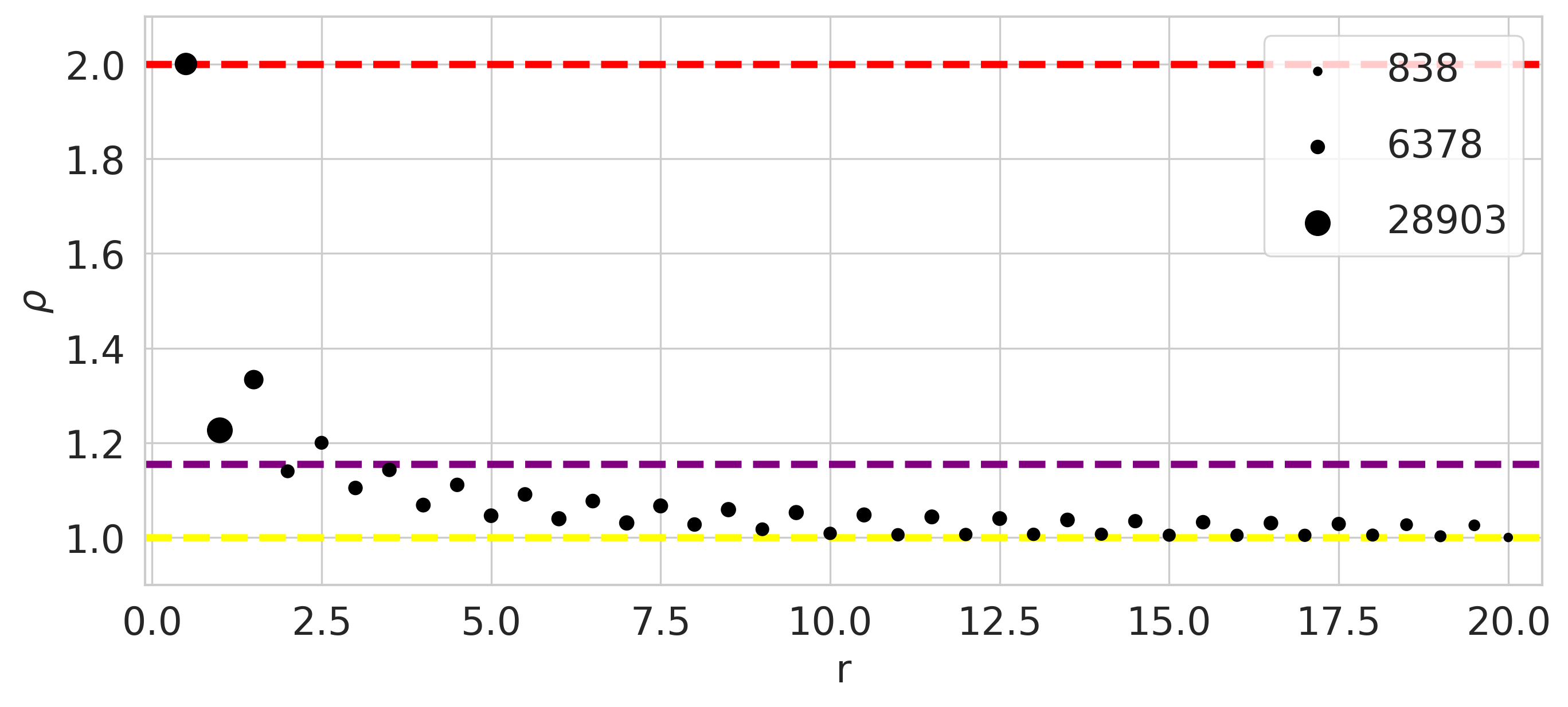}    \\
    \midrule
   \includegraphics[width=0.3\textwidth]{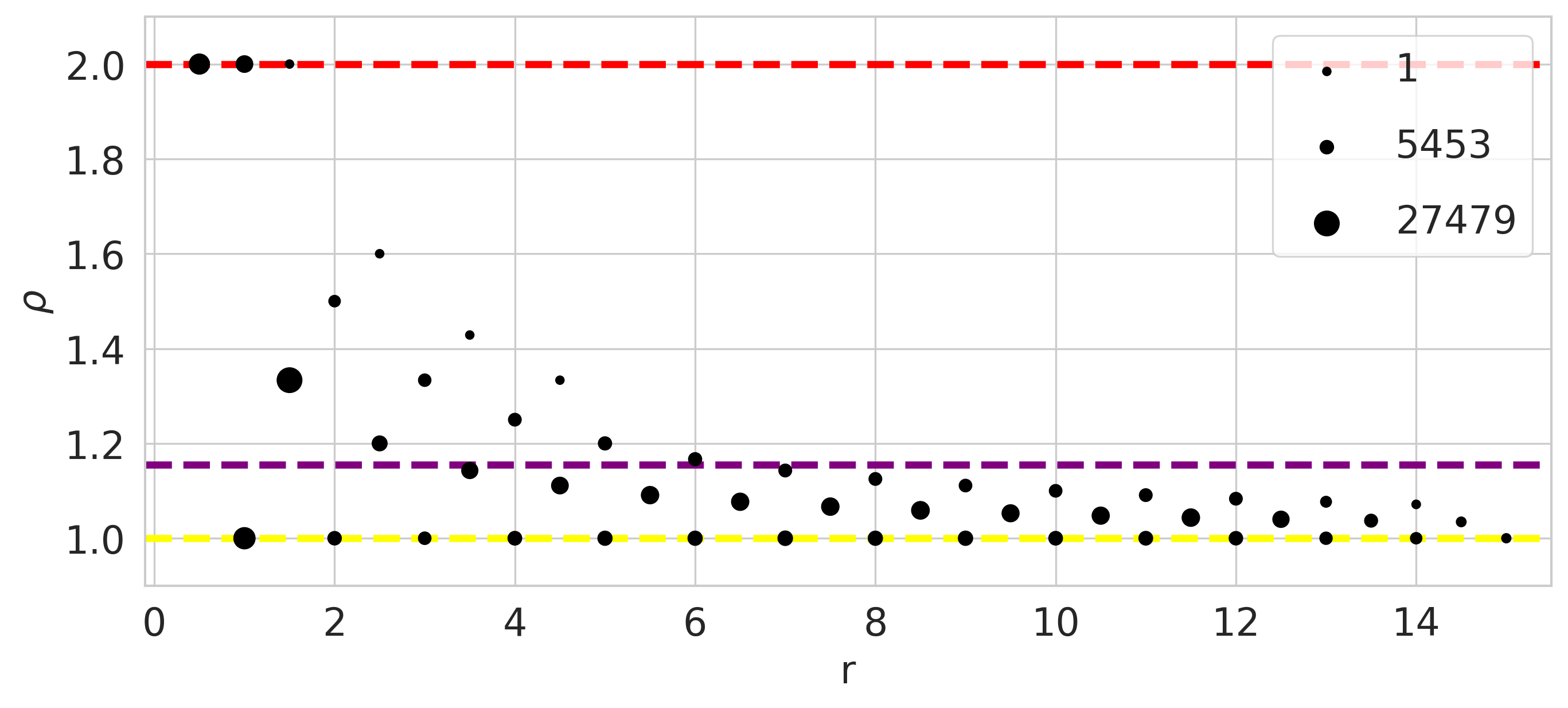} & \includegraphics[width=0.3\textwidth]{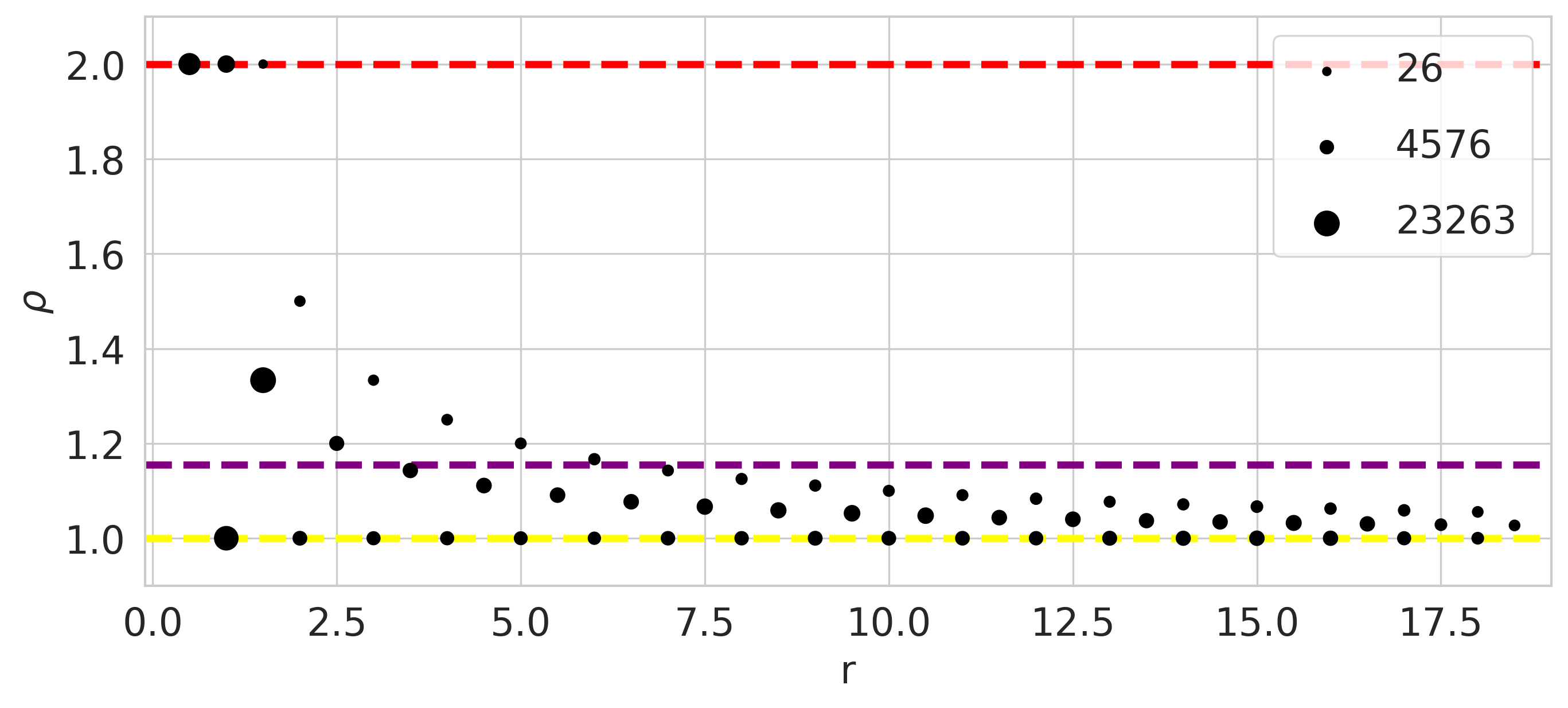} & \includegraphics[width=0.3\textwidth]{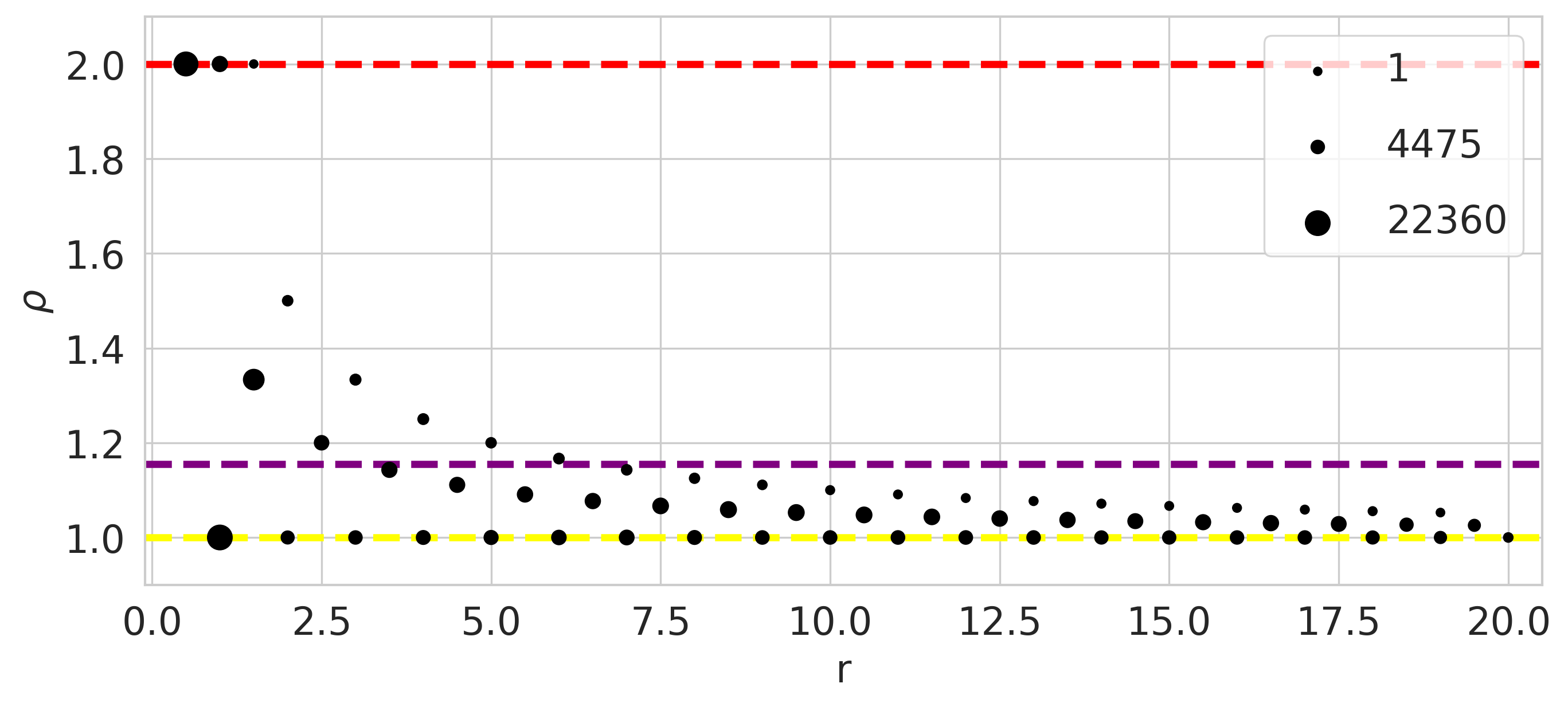}    \\
   \bottomrule
  \end{tabular}
\end{table}    

\begin{figure}[ht]
\centering
    { \includegraphics[width=0.32\textwidth]{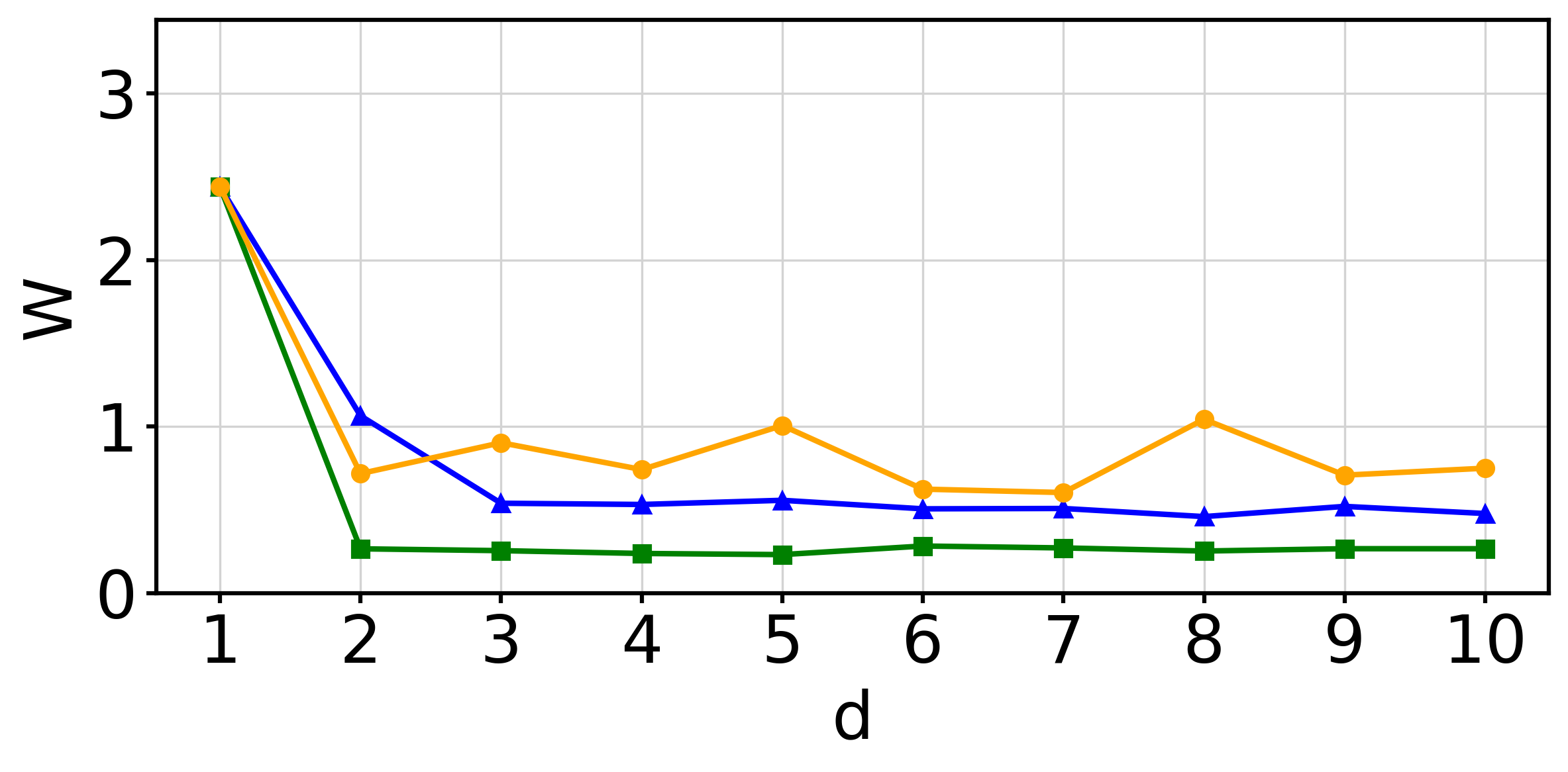}}
 	{\includegraphics[width=0.32\textwidth]{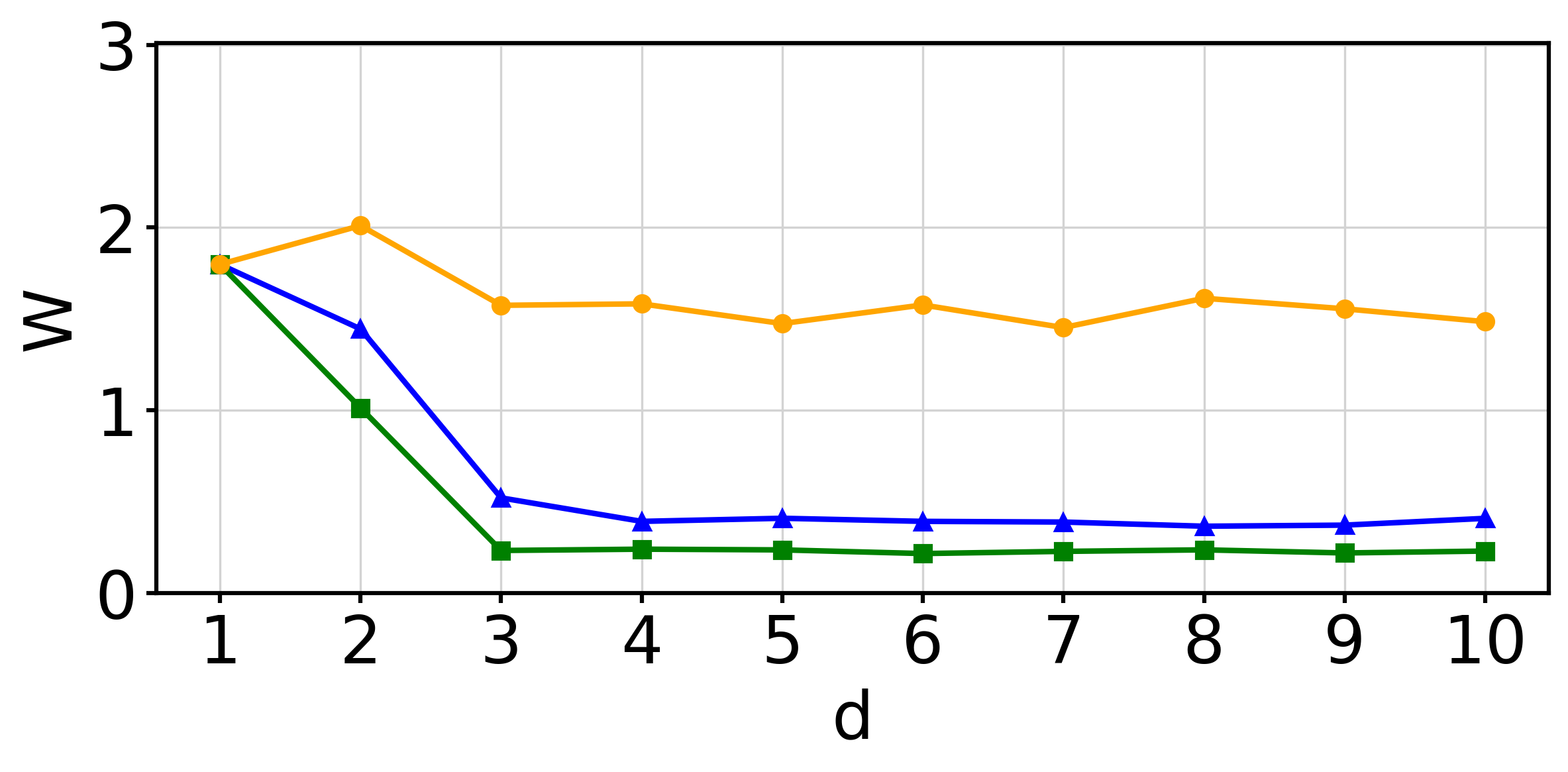}}
 	{\includegraphics[width=0.32\textwidth]{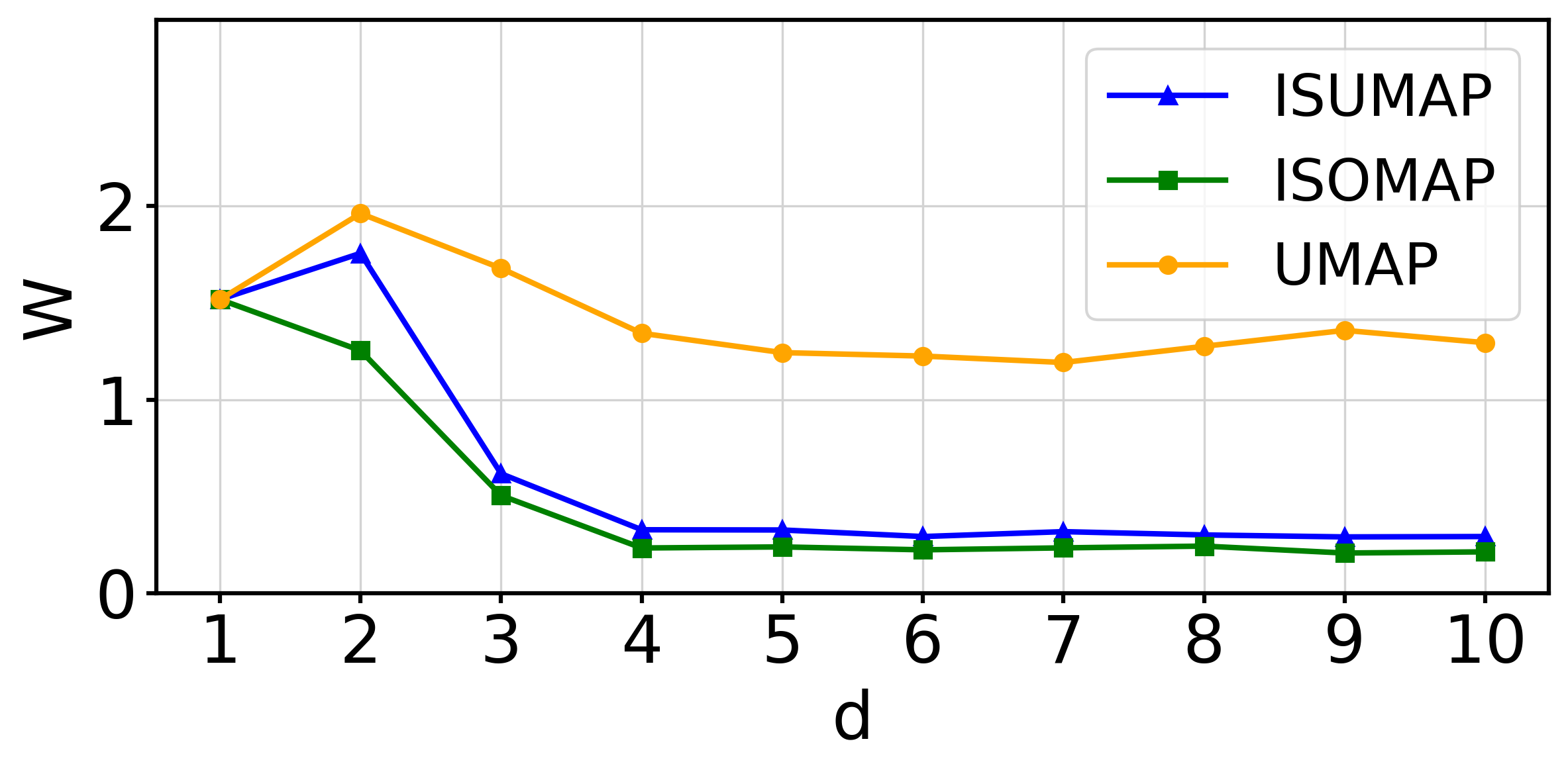}} 
 \caption{$W_1$-distance between the $\rho$ distributions of the original embeddings and their low-dimensional target embedding for synthetic datasets with intrinsic dimensions 2, 3, and 4 (shown left to right). $(k_{\min}, k_{\max}) = (10, 15)$.}
     \label{fig:toydim}
     \end{figure}

\begin{table}[H]
  \caption{The $W_1$ distance between curvature profiles for a range of embedding dimensions and $(k_{\min}, k_{\max}) = (10,15), (15,20),$ and $(20,30)$, across different dimensionality reduction techniques for the artificial tree generated in dimension $n=600$.
}
  \label{tab:warttree}
  \centering
  \begin{tabular}{c c c }
    \toprule
  \multicolumn{1}{c}{\small{10–15}} & \multicolumn{1}{c}{\small{15–20}} & \multicolumn{1}{c}{\small{20–30}} \\
    \midrule
       \includegraphics[width=0.31\linewidth]{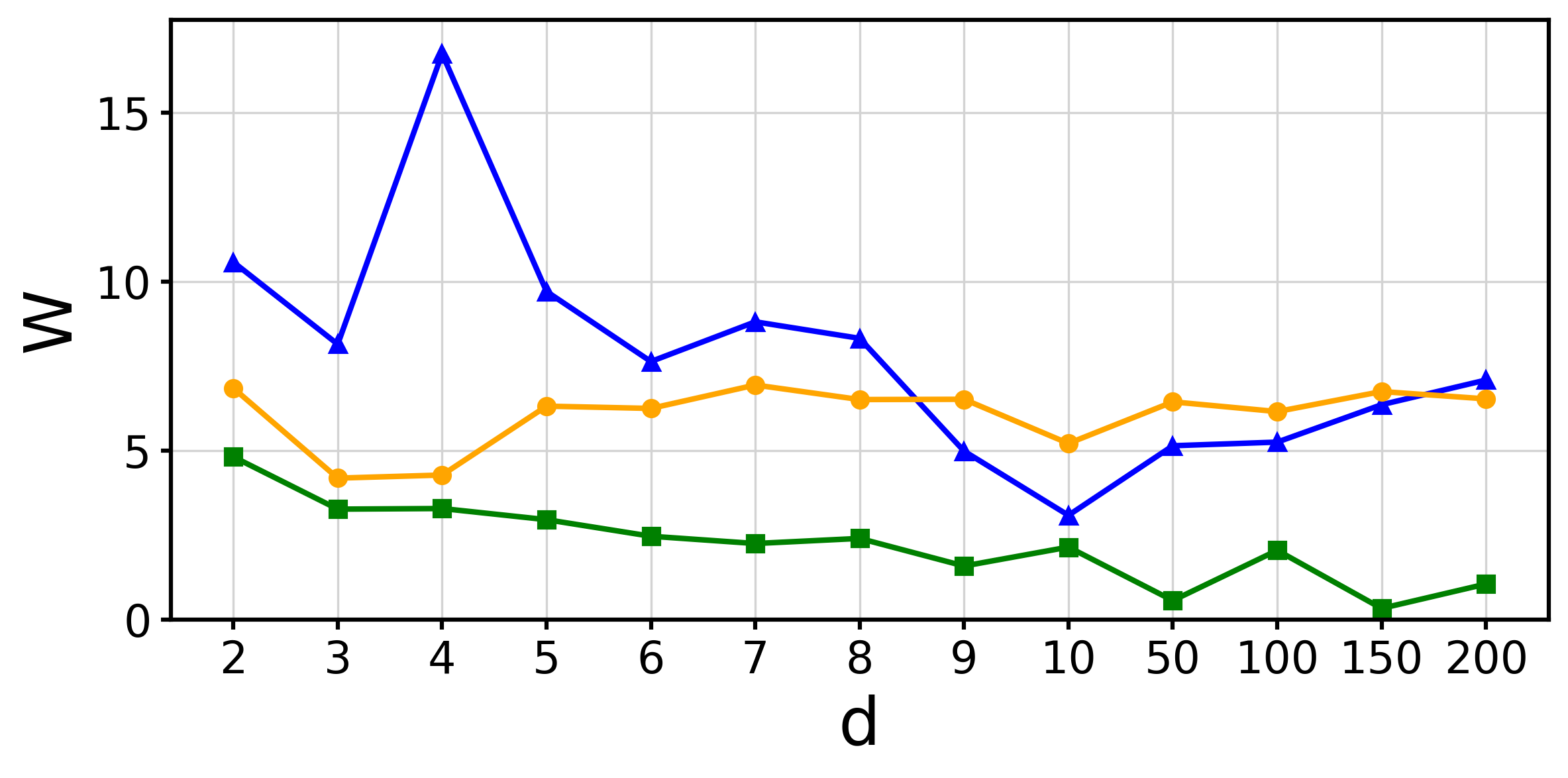} &
    \includegraphics[width=0.31\linewidth]{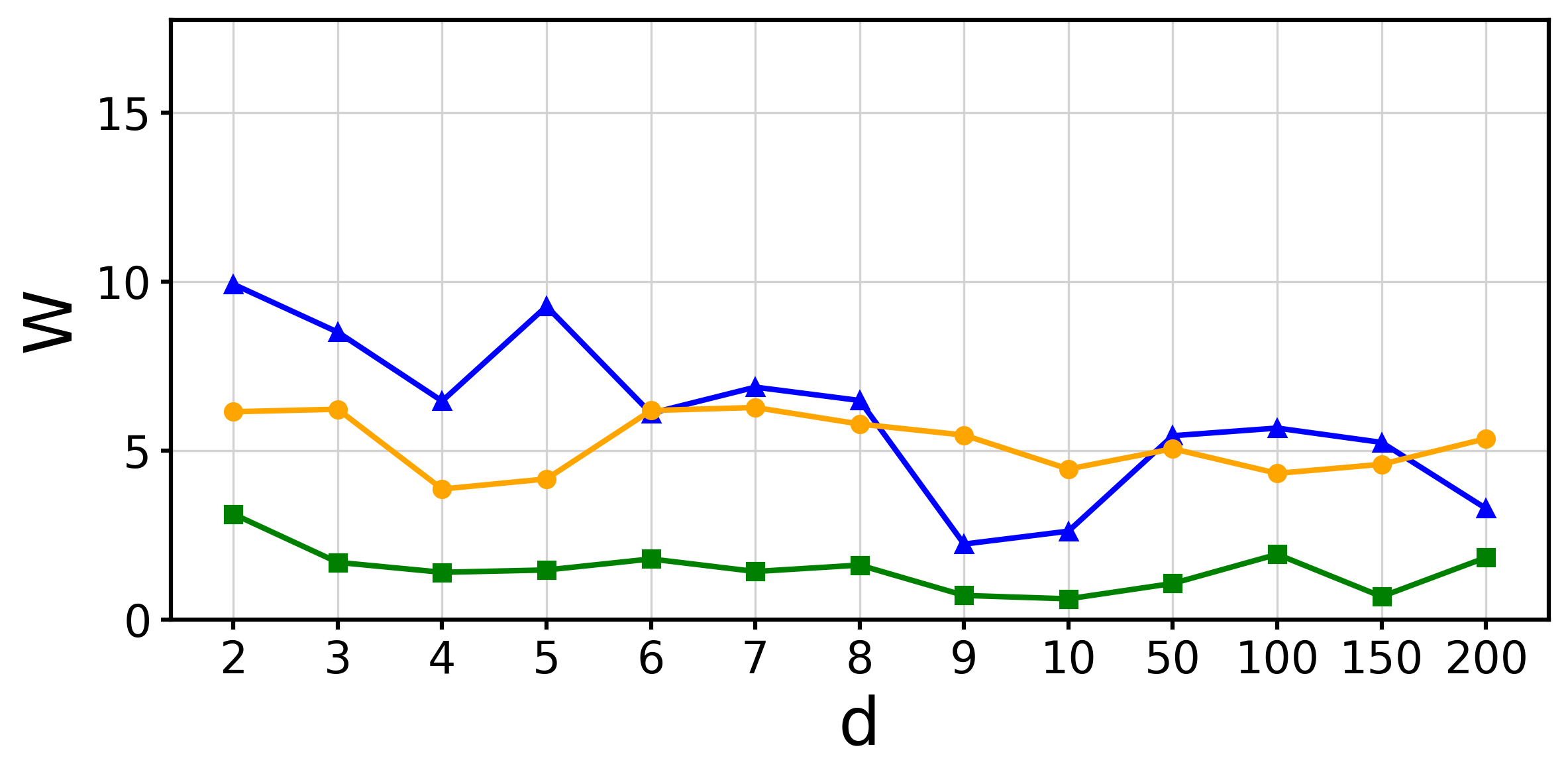} &
    \includegraphics[width=0.31\linewidth]{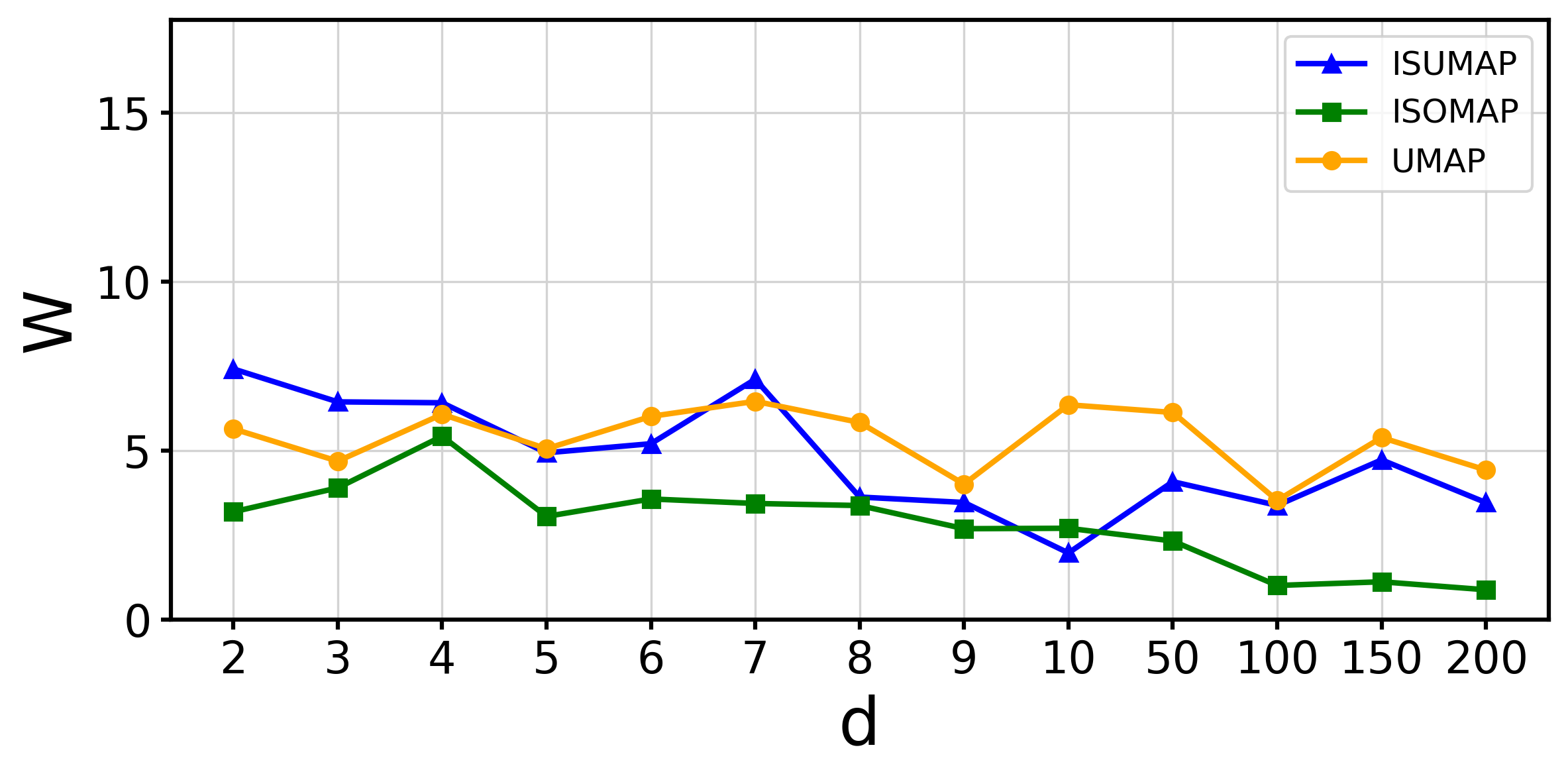} \\
    \bottomrule
  \end{tabular}
\end{table}
Next, we evaluate our method on synthetic datasets sampled from Euclidean spaces $\mathbb{R}^{n}$, with intrinsic dimensions $n \in \{2, 3, 4\}$. Each dataset is embedded into a higher-dimensional space $\mathbb{R}^{D}$, where $D = n + 50$, as follows. 
Let $N \in \mathbb{N}$ be the number of data points and $n$ the initial dimension. We start with  constructing a dataset $X \in \mathbb{R}^{N \times n}$ by sampling each entry independently from the standard normal distribution, i.e. $X_{ij} \sim \mathcal{N}(0, 1)$.
For the chosen dimension $D = n + 50$, we generate a random matrix $A \in \mathbb{R}^{D \times n}$ whose entries are also drawn from $\mathcal{N}(0,1)$. To ensure that the projection preserves distances, we compute the decomposition $A = Q R$ of $A$,  where $Q \in \mathbb{R}^{D \times n}$ has orthonormal columns ($Q^\top Q = I_{n}$) and  $R \in \mathbb{R}^{n \times n}$ is an upper triangular matrix. The matrix $Q$ thus constitutes an orthonormal basis for the column space of $A$ and defines an isometry. The isometric embedding of the dataset $X$ into $\mathbb{R}^{D}$ is given by $Y = X\, Q^\top$,  resulting in $Y \in \mathbb{R}^{N \times D}$. 
This procedure is applied iteratively for intrinsic dimensions $n \in \{1, 2, 3, 4\}$, and the output was fed to our pipeline.  
As shown in \cref{fig:toydim}, Isomap and IsUMap not only outperform UMAP in preserving geometric structure during dimensionality reduction, but also more consistently recover the intrinsic dimension of the original data. 

     \section{Discussion}
Based on a generalized sectional curvature applicable to discrete metric spaces, 
we have proposed a quantitative tool to evaluate the performance of methods that generate representations of metric datasets (e.g., feature embeddings).  Since sectional curvature captures the fundamental invariance in the smooth structures, the resulting curvature profile serves as a robust proxy for the intrinsic structure of both the original data and its embedded representation. This allows for a task-independent assessment of embedding quality.  Moreover, the curvature profile, quantifying a property of triple of points and how they are configured within the whole dataset, can be generalized to higher order correlations. We have demonstrated its ability to accurately distinguish between spaces with distinct geometric characteristics (such as tree-like, spherical, and planar datasets), and we aim to extend this approach further. Moreover, as the next step, our plan is to design a machine learning model capable of classifying datasets based on their underlying geometric properties, utilizing the curvature profile as a foundational feature.\\ 
Our current geometry inference method has some limitations, as it can become prohibitively slow on large datasets (e.g., the sRNA dataset). However, we plan to address this issue  by incorporating techniques from algebraic topology to improve scalability and efficiency. \\
Sectional curvature is not the only curvature notion that can be  used in graph and network analysis. In fact, synthetic Ricci curvatures constitute a well established tool, see e.g. \cite{Samal18}. But while the statistics of Ricci curvatures capture qualitative aspects of the local geometry, sectional curvatures show their strength on global features. It remains to combine these two different curvature based network analysis tool to forge a scheme that captures crucial geometric features at all scales simultaneously. \\ 

\newpage
\bibliographystyle{apalike}    
\bibliography{curvature_based_evaluation}

\end{document}